\documentclass[a4paper, 11pt, twoside]{article}
\usepackage{fullpage}

\usepackage{amsmath,amsfonts,amssymb,amsthm,array}
\usepackage{algorithm}
\usepackage{caption}
\usepackage{subcaption}
\usepackage{algpseudocode}
\usepackage{bm}
\usepackage{color}
\usepackage{graphicx}
\usepackage{enumitem}
\usepackage{nicefrac}

\usepackage{tikz}

\theoremstyle{plain}

\theoremstyle{remark}

\usepackage{mdframed} 
\usepackage{thmtools}

\usepackage{caption}
\usepackage{multirow}
\usepackage{colortbl}
\definecolor{bgcolor}{rgb}{0.8,1,1}
\definecolor{bgcolor2}{rgb}{0.8,1,0.8}
\definecolor{niceblue}{rgb}{0.0,0.19,0.56}

\usepackage{hyperref}
\hypersetup{colorlinks,linkcolor={blue},citecolor={niceblue},urlcolor={blue}}

\usepackage{natbib}
\usepackage{sidecap}

\definecolor{shadecolor}{gray}{0.9}
\declaretheoremstyle[
headfont=\normalfont\bfseries,
notefont=\mdseries, notebraces={(}{)},
bodyfont=\normalfont,
postheadspace=0.5em,
spaceabove=1pt,
mdframed={
  skipabove=8pt,
  skipbelow=8pt,
  hidealllines=true,
  backgroundcolor={shadecolor},
  innerleftmargin=4pt,
  innerrightmargin=4pt}
]{shaded}

\declaretheorem[style=shaded]{theorem}

\declaretheorem[style=shaded]{assumption}
\declaretheorem[style=shaded]{corollary}

\declaretheorem[style=shaded]{lemma}

\usepackage{xspace}

\usepackage[colorinlistoftodos,bordercolor=orange,backgroundcolor=orange!20,linecolor=orange,textsize=scriptsize]{todonotes}

\usepackage{microtype}
\usepackage{graphicx}
\usepackage{booktabs} 

\usepackage{hyperref}

\usepackage{amsmath}
\usepackage{amssymb}
\usepackage{mathtools}
\usepackage{amsthm}
\usepackage{bm}
\usepackage{bbm}
\usepackage{wrapfig}
\usepackage{graphicx}
\usepackage[export]{adjustbox}
\usepackage{arydshln}
\usepackage{subcaption}
\usepackage{array}
\usepackage{caption}
\usepackage{longtable}
\usepackage{nicefrac}  

\usepackage[capitalize,noabbrev]{cleveref}

\usepackage{algorithm}
\usepackage{algpseudocode}

\usepackage{makecell}
\usepackage{caption}
\usepackage{multirow}
\usepackage{colortbl}
\definecolor{bgcolor}{rgb}{0.8,1,1}
\definecolor{bgcolor2}{rgb}{0.8,1,0.8}
\usepackage{threeparttable}

\usepackage{tcolorbox}
\usepackage{pifont}
\definecolor{mydarkgreen}{RGB}{39,130,67}
\definecolor{mydarkred}{RGB}{192,47,25}

\makeatletter
\newcommand{\figcaption}[1]{\def\@captype{figure}\caption{#1}}
\newcommand{\tblcaption}[1]{\def\@captype{table}\caption{#1}}
\makeatother

\def\R{\mathbb{R}}
\newcommand{\E}{{\mathbb E}}

\def\R{\mathbb R}
\def\E{\mathbb E}
\def\EE{\mathbb E}

\allowdisplaybreaks

\title{On Scaled Methods for Saddle Point Problems}

\begin{document}

\title{\textbf{On Scaled Methods for Saddle Point Problems}}

\author{Aleksandr Beznosikov $^{1,2}$
\quad Aibek Alanov $^{3,4}$
\quad Dmitry Kovalev $^{5}$
\\
Martin Tak\'a\v{c} $^{2}$
\quad Alexander Gasnikov $^{1}$
}
\date{$^1$ Moscow Institute of Physics and Technology, Russian Federation\\
$^2$ Mohamed bin Zayed University of Artificial Intelligence, United Arab Emirates\\
$^3$ Artificial Intelligence Research Institute, Russian Federation\\
    $^4$ HSE University, Russian Federation\\
    $^5$ King Abdullah University of Science and Technology, Saudi Arabia
}

\maketitle

\begin{abstract}
Methods with adaptive scaling of different features play a key role in solving saddle point problems, primarily due to Adam's popularity for solving adversarial machine learning problems, including GANS training. 
This paper carries out a theoretical analysis of the following scaling techniques for solving SPPs: the well-known Adam and RmsProp scaling and the newer AdaHessian and OASIS based on Hutchison approximation. We use the Extra Gradient and its improved version with negative momentum as the basic method. Experimental studies on GANs show good applicability not only for Adam, but also for other less popular methods.
\end{abstract}

\section{Introduction}

In this paper, we focus on the saddle point problem (SPP):
\begin{equation}
    \label{eq:minmax}
    \min_{x \in \R^{d_x}} \max_{y \in \R^{d_y}} f(x,y).
\end{equation}
It has been of concern to the applied mathematics community for a long time. For example, SPPs arise in game theory, optimal control and equilibrium search in economics \cite{NeumannGameTheory1944,HarkerVIsurvey1990,VIbook2003}. The problem~\eqref{eq:minmax} is now even more widespread due to the current interest in solving adversarial formulations in machine learning. One of the brightest representatives of such tasks is Generative Adversarial Networks (GANs) \cite{goodfellow2014generative}. The main and most popular method of training GANs is Adam \cite{Kingma2015}.

Since its appearance, Adam has become a cornerstone learning method not only for GANs but also, for example, for NLPs \cite{albert} or models with attention \cite{zhang2020adaptive}. There are two features that distinguish Adam from SGD: a heavy-ball momentum as well as a gradient scaling. If momentum is more or less the standard acceleration approach, then scaling is precisely the key of Adam that allows it to be robust. Scaling is often also referred to as adaptivity, but it is different from classical adaptivity, where we simply recalculate the stepsize using information from previous iterations. Scaling modifies the entire gradient by multiplying each component by a different value. Adam was not the first method to use scaling \cite{Duchi2011,zeiler2012adadelta}, but the Adam update is the most popular amongst the community.

While Adam was gaining its practical relevance, it was also explored from a theoretical point of view for minimization problems \cite{reddi2019convergence, defossez2020simple}. Moreover, Adam and its modifications appear in many theoretical papers on SPPs and GANs training. The most famous of these works follow the same pattern \cite{daskalakis2017training,gidel2018variational,mertikopoulos2018optimistic,chavdarova2019reducing,pmlr-v89-liang19b,peng2020training}. The authors propose studies for methods without scaling, and then without theoretical justifications transfer the new features to Adam and obtain a new more powerful method. For example, \cite{gidel2018variational} adapts the classical and optimal method for SPPs, the Extra Gradient method, and obtains Extra Adam. But while Adam is a good framework into which some theoretical hints and intuitions can be inserted, up to the current time, other methods with scaling are hardly mentioned in the SPPs literature from a theoretical point of view. Recently, papers on scaling algorithms that provide guarantees of convergence have begun to appear \cite{liu2019towards, dou2021one, 9746485}. Despite the recent progress, there are still many disadvantages to these methods that should be addressed.  

\subsection{Our contribution and related works} \label{sec:contr}

\textbf{Eight methods with theory.} In this paper, we present eight methods with scaling for solving \eqref{eq:minmax}. In fact, we give two general methods in which different types of scaling can be substituted. We analyze four popular types of scaling in this paper, although the universality of the analysis allows exploring other scaling approaches. Theoretical convergence guarantees for these methods are provided in the strongly convex--strongly concave, convex--concave, and non-convex--non-concave (Minty) cases of the problem \eqref{eq:minmax}. 

\textbf{Extra Gradient.} Our first method (Algorithm \ref{Scaled_ExtraGrad}) is based on the Extra Gradient method \cite{Korpelevich1976TheEM, Nemirovski2004, juditsky2008solving}, an optimal and frequently used method for SPPs. The key feature of this method is the so-called extrapolation/extra step. 
We add scaling to this method. Unlike most of the works \cite{daskalakis2017training,gidel2018variational,mertikopoulos2018optimistic,chavdarova2019reducing,pmlr-v89-liang19b,peng2020training} which connect Extra Gradient and its modifications with Adam-based methods, we use the same scaling in both Extra Gradient steps and obtain to some extent a method that has not been encountered before. Consequently, the theoretical analysis for this kind of method is done for the first time and does not coincide, for example, with the analysis from \cite{liu2019towards, dou2021one}.

\textbf{Single call with momentum.} Our second algorithm is an improved version of the first one. 

Primarily, we solve the global problem of the Extra Gradient method -- the double calling of gradients at each iteration. We use a single call approach that allows us to compute the gradient once. Such methods without scaling were explored
e.g., in  \cite{popov1980modification, mokhtari2020convergence, hsieh2019convergence}. 

Secondly, we add momentum. Methods with momentum have been partially investigated for SPPs. In paper \cite{gidel2019negative}, the authors show that for SPPs (including
training GANs), one should use negative momentum because classical positive (heavy-ball) momentum can be not only useless but have a negative effect. One can also note that negative momentum was used in works \cite{alacaoglu2021stochastic} and \cite{kovalev2022optimal} to obtain optimal stochastic variance reduced methods for SPPs.

\textbf{Scalings.} As noted above, we consider different types of scaling approaches: two types of Adam-based scaling (Adam \cite{Kingma2015} itself, as well as RMSProp \cite{tieleman2012lecture}) and two types of rather new Hutchinson's approximation \cite{Bekas2007} scalings (AdaHessian \cite{Yao2020} and OASIS \cite{Jahani2021}). None of these methods have been analyzed theory-wise for SPPs before. Moreover, methods based on Hutchinson's approximation are found for the first time in an application to SPPs. Note that AdaHessian \cite{Yao2020} does not provide convergence analysis at all (not even for minimization problems).  
\renewcommand{\arraystretch}{1.5}
\begin{table*}[!h]
    \centering
	\caption{Comparison of scaled methods with theoretical analysis for SPPs.}
    \label{tab:comparison0}
    \scriptsize 
    \footnotesize 
\resizebox{\linewidth}{!}{
  \begin{threeparttable}
    \begin{tabular}{|c|c|c|c|c|c|}
    \hline
    \textbf{Method} & \textbf{Reference} & \textbf{Scaling} & \textbf{Single call?} & \textbf{Momentum} & \textbf{Assumptions}  \\
    \hline
   \texttt{OAdaGrad} & \cite{liu2019towards} & AdaGrad & yes & no & bounded gradients 
    \\ \hline
    \texttt{Extra Gradient AMSGrad} & \cite{dou2021one} & Adam-based & no & positive & bounded gradients
    \\ \hline
    \texttt{AMMO} & \cite{9746485} & Adam-based & no & positive & no theory \tnote{{\color{blue}(1)}} 
    \\ \hline \hline
    \cellcolor{bgcolor2}{\texttt{Extra Adam/RMSProp}} & \cellcolor{bgcolor2}{Ours} & \cellcolor{bgcolor2}{Adam-based}& \cellcolor{bgcolor2}{no}  & \cellcolor{bgcolor2}{no} & \cellcolor{bgcolor2}{bounded gradients}
    \\ \hline
    \cellcolor{bgcolor2}{\texttt{Extra AdaHessian/OASIS}} & \cellcolor{bgcolor2}{Ours} &   \cellcolor{bgcolor2}{Hutchinson}& \cellcolor{bgcolor2}{no}  & \cellcolor{bgcolor2}{no} & \cellcolor{bgcolor2}{}
    \\ \hline
    \cellcolor{bgcolor2}{\texttt{SCM Extra Adam/RMSProp}} & \cellcolor{bgcolor2}{Ours} & \cellcolor{bgcolor2}{Adam-based} & \cellcolor{bgcolor2}{yes}  & \cellcolor{bgcolor2}{negative}  & \cellcolor{bgcolor2}{bounded gradients}
    \\ \hline
    \cellcolor{bgcolor2}{\texttt{SCM Extra AdaHessian/OASIS}} & \cellcolor{bgcolor2}{Ours} & \cellcolor{bgcolor2}{Hutchinson} & \cellcolor{bgcolor2}{yes}  & \cellcolor{bgcolor2}{negative}  & \cellcolor{bgcolor2}{}
    \\ \hline
    \end{tabular}   
     \begin{tablenotes}
     {\scriptsize \footnotesize 
     \item [] \tnote{{\color{blue}(1)}} There is a theory in the paper, but it raises serious concerns. 
     }
 \end{tablenotes}    
    \end{threeparttable}
}
\end{table*}

\textbf{Main competitors.} One can highlight three main papers that also deal with the theoretical analysis of methods with scaling for SPPs. We give a comparison in Table \ref{tab:comparison0}. 
 
The work \cite{liu2019towards} presents an analysis of the Optimistic version of AdaGrad. The Optimistic approach is a single call, making the method more robust and faster in computation. Meanwhile, the authors do not add momentum to this method. The authors also need gradient boundedness as an assumption in the theoretical analysis; we do not need to make such an additional assumption for Hutchinson scaling methods.
 
The authors of \cite{dou2021one} propose an analysis for the Adam-Based (but not Adam) method. Their method is based on the Extra Gradient method with the AMSGrad update \cite{reddi2019convergence}. Unlike some of our methods and the method from paper \cite{liu2019towards}, this method is not single-call, which in practice can slow down the running time by almost a factor of two. 
In this paper,  the boundedness of the gradients is assumed.

Finally, both papers \cite{liu2019towards} and \cite{dou2021one} give convergence only in the non-convex case. We give an analysis for the strongly-convex and convex cases as well. It seems that this is the way to see a more complete picture of how the methods work. This is due to the fact that any guarantees in the non-convex case (not only for SPPs but also for minimization problems) are just convergence to some (most likely the nearest) stationary point, but in practice, for neural network problems, we converge not just to some stationary point, but probably to a good neighborhood of global solution. The practical behavior of the methods lies somewhere between the general non-convex and convex cases.

The paper \cite{9746485} presents several Adam-based methods, but its theoretical foundation is disturbing. We found no proofs in the paper (the authors refer to the arxiv version, which does not exist at the current moment). Moreover, the theorems postulate convergence but not the rate of convergence. 

\textbf{Practical performance.}
In practice, we explore four methods of scaling (RMSProp, Adam, AdaHessian, OASIS) for training GANs. We test how these methods are affected by the extra step technique that arises from the theory. We further investigate the effect of negative momentum on final learning results.


\section{Problem Setup and Assumption}

We write $\langle x,y \rangle = \sum_{i=1}^d x_i y_i$ to denote the standard inner product of vectors $x,y\in\R^d$, where $x_i$ corresponds to the $i$-th component of $x$ in the standard basis in $\R^d$. This induces the standard $\ell_2$-norm in $\R^d$ in the following way: $\|x\| = \sqrt{\langle x, x \rangle}$.
For a given positive definite  matrix $D \in \mathcal{S}^d_{++}$, the weighted Euclidean norm is defined to be $\|x\|_D^2 = \langle x, D x\rangle$, where $x\in \R^d$. We also introduce $\lambda_{\max}(D)$ as the maximum eigenvalue of $D$, $\lambda_{\min}(D)$ as the minimum eigenvalue of $D$. Using this notation, for a given matrix $A \in \R^{d \times d}$, we define the spectral norm as $\| A \| = \sqrt{\lambda_{\max}\left(A^T A \right)}$, and the infinity norm as $\| A\|_{\infty} =  \max_{1\leq i\leq d}\sum _{j=1}^{d}|a_{ij}|$, where $\{a_{ij}\}^{n,n}_{i=1, j = 1}$ are elements of $A$.
The symbol $\odot$ denotes the component-wise product between two vectors, and {\tt diag}$(x)$ denotes the $d\times d$ diagonal matrix whose diagonal entries are the components of the vector $x\in \R^d$.

We consider the problem \eqref{eq:minmax} and assume that for the function $f(x,y)$ we have only access to stochastic oracle $f(x,y, \xi)$. For example, $\xi$ could be the number of the batch for which the loss is computed. Additionally, only stochastic gradients are available $\nabla_x f(x,y,\xi)$, $\nabla_y f(x,y,\xi)$.
Next, we list the main assumptions on \eqref{eq:minmax} and stochastic gradients.

\begin{assumption}[Convexity-concavity] \label{as:conv}
We consider three cases of convexity-concavity of \eqref{eq:minmax}.

\textbf{(SC) Strong convexity-- strong concavity.} The function $f$ is $\mu$-strongly convex--strongly concave, i.e. for all $x_1, x_2 \in \R^{d_x}$ and $y_1, y_2 \in \R^{d_y}$ we have
\begin{align}
    \label{sm1}
    \begin{split}
    &\langle \nabla_x f (x_1,y_1) - \nabla_x f (x_2, y_2), x_1 - x_2 \rangle
    - \langle \nabla_y f (x_1,y_1) - \nabla_y f (x_2, y_2), y_1 - y_2 \rangle \\
    &\hspace{8cm}\geq \mu \left(\|x_1-x_2\|^2 + \|y_1-y_2\|^2\right).
    \end{split}
\end{align}
\textbf{(C) Convexity--concavity.} The function $f$ is convex--concave, i.e. for all $x_1, x_2 \in \R^{d_x}$ and $y_1, y_2 \in \R^{d_y}$ we have
\begin{equation}
    \label{m}
    \begin{split}
    \langle \nabla_x f (x_1,y_1) - \nabla_x f (x_2, y_2), x_1 - x_2 \rangle - \langle \nabla_y f (x_1,y_1) - \nabla_y f (x_2, y_2), y_1 - y_2 \rangle \geq 0.
    \end{split}
\end{equation}
\textbf{(NC) Non-convexity--non-concavity (Minty).} The function $f$ satisfies the Minty condition, if and only if there exists $x^* \in \R^{d_x}$ and $y^* \in \R^{d_y}$ such that for all $x \in \R^{d_x}$ and $y \in \R^{d_y}$ we have
\begin{equation}
    \label{nm}
    \langle \nabla_x f (x,y), x - x^* \rangle - \langle \nabla_y f (x,y), y - y^* \rangle\geq 0.
\end{equation}
\end{assumption}
The last assumption is not a general non-convexity--non-concavity, but it is firmly associated and regarded in the literature as a good relaxation of non-convexity--non-concavity. \eqref{nm} can be found under the names of the Minty or variational stability condition \cite{minty62,liu2019towards, hsieh2020explore, dou2021one, diakonikolas2021efficient}.

\begin{assumption}[$L$-smoothness] \label{as:Lipsh}
For any $\xi$ the function $f(x,y,\xi)$ is $L$-smooth, i.e. for all $x_1, x_2 \in \R^{d_x}$ and $y_1, y_2 \in \R^{d_y}$ we have
\begin{equation}
\label{eq:Lipsh}
\begin{split}
&\|\nabla_x f (x_1,y_1, \xi) - \nabla_x f (x_2, y_2, \xi)\|^2 + \|\nabla_y f (x_1,y_1, \xi) - \nabla_y f (x_2, y_2, \xi) \|^2 \\
&\hspace{8cm}\leq L^2\left(\|x_1-x_2\|^2 + \|y_1-y_2\|^2\right).
\end{split}
\end{equation}
\end{assumption}

\begin{assumption}[Bounded variance] \label{as:var}
Stochastic gradients of $\nabla_{x} f(x,y,\xi)$,
$\nabla_{y} f(x,y,\xi)$
are unbiased
and their variance are bounded, i.e. 
there exists $\sigma^2 \geq 0$
such that
for all $x \in \R^{d_x}$ and $y \in \R^{d_y}$ we have
\begin{equation}
\label{eq:var}
\begin{split}
&\EE \left[\nabla_x f (x,y, \xi)\right] = \nabla_x f (x, y), \quad \EE \left[\nabla_y f (x,y, \xi)\right] = \nabla_y f (x, y), \\
&\EE\big[\|\nabla_x f (x,y, \xi) - \nabla_x f (x, y)\|^2 + \|\nabla_y f (x,y, \xi) - \nabla_y f (x, y) \|^2\big] \leq \frac{\sigma^2}{b},
\end{split}
\end{equation}
where the parameter $b$ stands for the size of the batch in the stochastic gradients.
\end{assumption}

\section{Scaled Methods} \label{sec:scaled}

Unlike SGD-type methods that use a stochastic gradient $g_t$ for an update, scaled methods additionally compute a so-called preconditioning matrix $\hat D_t$ and produce a scaled gradient $\hat g_t = \hat D^{-1}_t g_t$.  
For SPPs, obtaining preconditioning matrices can be written as follows:
\begin{equation}
    \label{eq:precond}
    \begin{split}
    (D^x_{t})^2 =  \beta_{t} (D^x_{t-1})^2 + (1 - \beta_{t}) (H^x_t)^2, \quad 
    (D^y_{t})^2 =  \beta_{t} (D^y_{t-1})^2 + (1 - \beta_{t}) (H^y_t)^2,
    \end{split}
\end{equation}
where $\beta_{t} \in [0;1]$ is a preconditioning momentum parameter (typically close to 1), $(H^x_t)^2$ and $(H^y_t)^2$ are some diagonal matrices. The update \eqref{eq:precond} is satisfied by Adam-based methods with $(H^x_t)^2 = \text{diag}(\nabla_x f (x_t,y_t, \xi^D_t) \odot \nabla_x f (x_t,y_t, \xi^D_t))$ and $(H^y_t)^2 = \text{diag}(\nabla_y f (x_t,y_t, \xi^D_t) \odot \nabla_y f (x_t,y_t, \xi^D_t))$ (here $\xi^D_t$ is an independent random variable specially generated to calculate $H^x_t$ and $H^y_t$), e.g. the original Adam \cite{Kingma2015} ($\beta_t = \frac{\beta - \beta^{t+1}}{1 - \beta^{t+1}}$)
and the earlier method, RMSProp \cite{tieleman2012lecture} ($\beta_t \equiv \beta$). The rule \eqref{eq:precond} also applies to AdaHessian \cite{Yao2020} -- method with Hutchinson scaling, here one should take $\beta_t = \frac{\beta - \beta^{t+1}}{1 - \beta^{t+1}}$, $(H^x_t)^2 = \text{diag}(v^x_t \odot \nabla_{xx} f(x_t,y_t,\xi^D_t) v^x_t )^2$ and $(H^y_t)^2 = \text{diag}(v^y_t \odot \nabla_{yy} f(x_t,y_t,\xi^D_t) v^y_t )^2$, where  $v^x_t, v^y_t$ are from Rademacher distribution.\footnote{All components of vectors $v^x_t, v^y_t$ are independent and equal to $\pm 1$ with probability $1/2$.}
Let us stress that despite AdaHessian includes a stochastic Hessian, it does not need to be calculated explicitly as it needs just Hessian-vector product. Hence, the matrices $H^x_t$ and $H^y_t$ can be computed by performing back-propagation twice: firstly, we compute the stochastic gradients $\nabla_x f(x_t,y_t,\xi^D_t)$ and $\nabla_y f(x_t,y_t,\xi^D_t)$, then compute the gradients of the functions $\langle \nabla_x f(x_t,y_t,\xi^D_t), v^x_t \rangle$ and $\langle \nabla_y f(x_t,y_t,\xi^D_t), v^y_t \rangle$.

The following update can be used as an analogue of \eqref{eq:precond}:
\begin{equation}
\label{eq:precond_add}
\begin{split}
D^x_{t} =  \beta_{t} D^x_{t-1} + (1 - \beta_{t}) H^x_t , \quad 
D^y_{t} =  \beta_{t} D^y_{t-1} + (1 - \beta_{t}) H^y_t.
\end{split}
\end{equation}
For example, such a rule is augmented by OASIS \cite{Jahani2021} with $\beta_t \equiv \beta$ and $H^x_t = \text{diag}(v^x_t \odot \nabla_{xx} f(x_t,y_t,\xi^D_t) v^x_t )$ and $H^y_t = \text{diag}(v^y_t \odot \nabla_{yy} f(x_t,y_t,\xi^D_t) v^y_t )$. It turns out that expression \eqref{eq:precond_add} can easily be calculated by differentiating twice. Firstly, we find $\nabla_{x} f(x_t,y_t,\xi^D_t)$, and then we compute the gradient again, but for $(\nabla_{x} f(x_t,y_t,\xi^D_t))^T v^x_t$. As a result, we have $\nabla_{xx} f(x_t,y_t,\xi^D_t) v^x_t$. Using  the component-wise product between $v^x_t$ and $\nabla_{xx} f(x_t,y_t,\xi^D_t) v^x_t$, we get the diagonal of $D^x_t$.

Moreover, \eqref{eq:precond_add} is a good approximation of the Hessians' diagonals. Indeed, it is easy to see that in the expectation the non-diagonal elements of the matrices $D^x_{t}$, $D^y_{t}$ are equal to $0$ and the diagonal elements are equal the corresponding Hessians' elements.

To make the matrix positive definite, one can use the following transformation that is common in literature:
\begin{equation}
    \label{eq:precond_abs}
    \begin{split}
    (\hat D^x_{t})_{ii} =  \max\left\{ e, |(D^x_{t})_{ii}|\right\},  \quad 
    (\hat D^y_{t})_{ii} =  \max\left\{ e, |(D^y_{t})_{ii}|\right\},
    \end{split}
\end{equation}
where $e$ is a positive parameter.
It was suggested e.g.,  in the original Adam paper \cite{Kingma2015}. Subsequently, the critical necessity of \eqref{eq:precond_abs} was shown in  \cite{reddi2019convergence}. One can alternatively define $(\hat D^x_{t})_{ii} =  |(D^x_{t})_{ii}| + e$ and $(\hat D^y_{t})_{ii} =  |(D^y_{t})_{ii}| + e$. 
It can be seen that all of the above scaling approaches construct diagonal matrices with positive elements. We formulate the properties of such matrices via the following lemma.
\begin{lemma} \label{lem:precond}
Let us assume that $D_0^x$, $D_0^y$ and for all $t$ the  $H^x_t$, $H^y_t$ are diagonal matrices with elements not greater than $\Gamma$ in absolute value. Then for matrices $\hat D_t^x$, $\hat D_t^y$ obtained by rules \eqref{eq:precond} -- \eqref{eq:precond_abs}, the following holds

1) $\hat D_t^x$, $\hat D_t^y$ are diagonal matrices with non-negative elements and $eI \preccurlyeq \hat D^x_{t} \preccurlyeq \Gamma I$,~$eI \preccurlyeq \hat D^y_{t} \preccurlyeq \Gamma I$;

2) $\hat D^x_{t+1} \preccurlyeq \left(1 + \frac{(1-\beta_{t+1})\Gamma^2 }{2e^2}\right) \hat D^x_{t}$ and  $\hat D^y_{t+1} \preccurlyeq \left(1 + \frac{(1-\beta_{t+1}) \Gamma^2 }{2e^2}\right) \hat D^y_{t}$ for \eqref{eq:precond};

3) $\hat D^x_{t+1} \preccurlyeq \left(1 + \frac{2(1-\beta_{t+1})\Gamma }{e}\right) \hat D^x_{t}$ and $\hat D^y_{t+1} \preccurlyeq \left(1 + \frac{2(1-\beta_{t+1}) \Gamma }{e}\right) \hat D^y_{t}$ for  \eqref{eq:precond_add}.
\end{lemma}
 Let us formally prove that all scalings: Adam, RMSProp, AdaHessian, and OASIS, satisfy the conditions of Lemma~\ref{lem:precond}.
\begin{lemma} \label{lem:pract_prec}
Let Assumption \ref{as:Lipsh} be satisfied, then the AdaHessian and OASIS preconditioners satisfy the conditions of the Lemma \ref{lem:precond} with $\Gamma = \sqrt{d_x+ d_y} L$. If $\|\nabla_x f(x, y, \xi)\| \leq M$ and $\|\nabla_y f(x, y, \xi)\| \leq M$ for all $x,y,\xi$, then the Adam and RMSProp preconditioners satisfy the conditions of the Lemma \ref{lem:precond} with $\Gamma = M$.
\end{lemma}

\begin{algorithm*}
\caption{{\tt Scaled Stochastic Extra Gradient}}
\label{Scaled_ExtraGrad}
	\begin{algorithmic}[1]
		\State \textbf{Input:} initial point $x_0, y_0$
		\For{$t = 0,1,2,\ldots T-1$}
		\State Generate independently $\xi^D_{t}$ for $\hat{D}^x_{t}, \hat{D}^y_{t}$ \label{lin:2}
		\State Compute the preconditioners $\hat{D}^x_{t}$ and $\hat{D}^y_{t}$ according   to \eqref{eq:precond}--\eqref{eq:precond_abs} \label{lin:3}
		\State Generate independently $\xi_{t}$ for $\nabla_x f(x_t, y_t, \xi_t), \nabla_y f(x_t, y_t, \xi_t)$ \label{lin:4}
		\State $x_{t+1/2} = x_t - \gamma (\hat D^x_t)^{-1} \nabla_x f(x_t, y_t, \xi_t)$ \label{lin:5}
		\State $y_{t+1/2} = y_t + \gamma (\hat D^y_t)^{-1} \nabla_y f(x_t, y_t, \xi_t)$ \label{lin:6}
		\State Generate independently $\xi_{t+1/2}$ for \text{\small{$\nabla_x f(x_{t+1/2}, y_{t+1/2}, \xi_{t+1/2})$,  $\nabla_y f(x_{t+1/2}, y_{t+1/2}, \xi_{t+1/2})$}}
		\State $x_{t+1} = x_t - \gamma (\hat D^x_t)^{-1} \nabla_x f(x_{t+1/2}, y_{t+1/2}, \xi_{t+1/2})$ \label{lin:8}
		\State $y_{t+1} = y_t + \gamma (\hat D^y_t)^{-1} \nabla_y f(x_{t+1/2}, y_{t+1/2}, \xi_{t+1/2})$ \label{lin:9}
		\EndFor
	\end{algorithmic}
\end{algorithm*}

The rule chains \eqref{eq:precond} + \eqref{eq:precond_abs} or \eqref{eq:precond_add} + \eqref{eq:precond_abs} allow to construct preconditioning matrices. Next, we present Algorithm \ref{Scaled_ExtraGrad}. As mentioned in Section \ref{sec:contr}, we use the stochastic Extra Gradient as the base algorithm. Such an algorithm is 
a modification of SGD for SPPs. Its main difference is the presence of an extrapolation step (lines \ref{lin:5} and \ref{lin:6}) before the main update (lines \ref{lin:8} and \ref{lin:9}). This complicates the algorithm, because it is necessary to double count the gradient at each iteration, but meanwhile the original SGD is not optimal for SPPs. Moreover, it can perform worse and diverge even on the simplest problems (see Sections 7.2 and 8.2 of \cite{goodfellow2016nips}). 
We add the same preconditioners to both steps of the Extra Gradient. These matrices are computed at the beginning of each iteration (lines \ref{lin:2} and \ref{lin:3}).

The following theorem gives a convergence estimate for Algorithm \ref{Scaled_ExtraGrad}. 
\begin{theorem} \label{th:main0}
Consider the problem \eqref{eq:minmax} under Assumptions~\ref{as:Lipsh} and \ref{as:var} and conditions of Lemma~\ref{lem:precond}. Let  $\{x_t, y_t\}$ and $\{x_{t+1/2}, y_{t+1/2}\}$ be the sequences generated by Algorithm \ref{Scaled_ExtraGrad} after $T$ iterations. Also assume that the sequence $\{\beta_t\}$ is not random and $\gamma \leq \frac{e}{4 L}$. Then, 

$\bullet$ under Assumption \ref{as:conv}(SC) it holds that 
\begin{align}
\label{eq:th_sc}
\begin{split}
    \E\left[ R^2_{t+1} \right]  \leq&  \alpha_t \E\left[R^2_{t}\right] +\theta_t,
\end{split}
\end{align}
with $\alpha_t= \left( 1 - \frac{\gamma \mu}{\Gamma} + (1-\beta_{t+1})C \right)$ and $\theta_t= \frac{6\sigma^2 \left( 1 + (1-\beta_{t+1})C \right) \gamma^2}{be}$;

$\bullet$ under Assumption \ref{as:conv}(C) with $\| x_t \| \leq \Omega$, $\| y_t \| \leq \Omega$ for all $t \geq 0$ it holds that 
 \begin{align}
 \label{eq:th_c}
 \begin{split}
 \E\big[\text{gap}(x_{T}^{avg}, y_{T}^{avg})\big] \leq  \frac{3\Gamma\Omega^2}{\gamma T} + 
\frac{2 C\Gamma \Omega^2}{\gamma T} \sum\limits_{t=1}^{T}  (1-\beta_{t}) + \frac{4\gamma \sigma^2}{e b};    
 \end{split}
 \end{align}

$\bullet$ under Assumption \ref{as:conv}(NC) with $\| x_t \| \leq \Omega$, $\| y_t \| \leq \Omega$ for all $t \geq 0$ it holds that 
\begin{align}
 \label{eq:th_nc}
 \begin{split}
\E\big[ \| \nabla_x f(\bar x_T, \bar y_T) \|^2 + \| \nabla_y f(\bar x_T, \bar y_T) \|^2\big] \leq   \frac{12\Gamma^2 \Omega^2}{\gamma^2 T} + \frac{21 \Gamma \sigma^2}{e b} +  \frac{12 C\Gamma^2 \Omega^2}{ \gamma^2 T } \sum\limits_{t=1}^{T} (1-\beta_{t}).
 \end{split}
\end{align}
Here we use $R^2_{t} = \|x_{t} - x^* \|^2_{\hat D^x_{t}} + \|y_{t} - y^* \|^2_{\hat D^y_{t}}$, $\text{gap}(x, y) = \max\limits_{y' \in \mathcal{Y}} f\left(x, y'\right) - \min\limits_{x' \in \mathcal{X}} f\left(x', y\right) $,  $x_{T}^{avg} = \frac{1}{T}\sum_{t=0}^{T-1} x_{t+1/2}$, $y_{T}^{avg} = \frac{1}{T}\sum_{t=0}^{T-1} y_{t+1/2}$, $(\bar x_T, \bar y_T)$ is picked uniformly from all $(x_t, y_t)$, 
$C = \frac{\Gamma^2}{2e^2}$ for \eqref{eq:precond} or $C = \frac{2\Gamma}{e}$ for \eqref{eq:precond_add}.
\end{theorem}
The estimate \eqref{eq:th_c} uses a so-called gap function on some bounded set $\mathcal{X} \times \mathcal{Y}$. Eventhough we solve the problem \eqref{eq:minmax} on the whole set $\mathbb{R}^{d_x} \times \mathbb{R}^{d_y}$, this is permissible, since such a version of the criterion is valid if the solution $x^{*}, y^*$ lies in $\mathcal{X} \times \mathcal{Y}$; for details see \cite{nesterov2007dual}.

Depending on the choice of $\beta_t$, we can get convergence for the four methods that are given before. In particular we can estimate the number of iterations to find a solution with $\varepsilon$ precision (in the strongly convex--strongly concave case this means that $\|x_{t} - x^* \|^2+ \|y_{t} - y^* \|^2 \sim \varepsilon$, in the convex--concave that $\text{gap}(x, y) \sim \varepsilon$, and in the non-convex--non-concave that $\| \nabla_x f(x, y) \|^2 + \| \nabla_y f(x, y) \|^2 \sim \varepsilon^2$).
\begin{corollary}[RMSProp and OASIS] \label{cor:main_oasis}
Under the assumptions of Theorem \ref{th:main0} let additionally take $\beta_t \equiv \beta$. Then, the number of iterations to achieve $\varepsilon$-solution (in expectation) 

$\bullet$ in the strongly convex--strongly concave case with $\beta \geq 1 - \frac{e\gamma \mu}{ 2 \Gamma C}$ is $\mathcal{\tilde O} \left( \frac{\Gamma L}{e \mu} \log \frac{1}{\varepsilon}  + \frac{\Gamma^3 \sigma^2}{e^2 b \mu^2  \varepsilon}\right)$;

$\bullet$ in the convex--concave case with $\beta \geq 1 - \frac{1}{C T}$  is $\mathcal{O}\left(\frac{\Gamma L\Omega^2}{e \varepsilon} + 
 \frac{\Gamma\sigma^2 \Omega^2}{e b\varepsilon^2} \right)$;

$\bullet$ in the non-convex--non-concave case with $\beta \geq 1 - \frac{1}{C T}$ and $b \sim \frac{\Gamma \sigma^2}{e \varepsilon^2}$ is $\mathcal{O}\left( \frac{\Gamma^2 L^2 \Omega^2}{e^2 \varepsilon^2} \right)$.
\end{corollary}

\begin{corollary}[Adam and AdaHessian] \label{cor:main_adam}
Under the assumptions of Theorem \ref{th:main0}, let additionally take $\beta_t = \frac{\beta - \beta^{t+1}}{1 - \beta^{t+1}}$. 
Then, the complexities to achieve $\varepsilon$-solution (in expectation)

$\bullet$ in the str. convex--str. concave case with $\beta = 1 - \left(\frac{\gamma \mu}{ 2 \Gamma C}\right)^2$ is $\mathcal{O} \left(\frac{C^2\Gamma^3 L^2 \Omega^2 }{e^3 \mu^2 \varepsilon} + \frac{C^4 \Gamma^5 \sigma^4 \Omega^2}{e^5 b^2 \mu^4 \varepsilon^3}\right)$;

$\bullet$ in the convex--concave case with $\beta = 1 - \frac{1}{T}$  is $\mathcal{O}\left(\frac{C^2\Gamma^2 L^2\Omega^4}{e^2 \varepsilon^2 } + \frac{C^2\Gamma^2 \sigma^4 \Omega^4}{e^2 b^2 \varepsilon^4} \right)$;

$\bullet$ in the non-convex--non-concave case with $\beta = 1 - \frac{1}{T}$ and $b \sim \frac{\Gamma \sigma^2}{e \varepsilon^2}$ is $\mathcal{O}\left(\frac{C^2\Gamma^4 L^4 \Omega^4 }{ e^4 \varepsilon^4}\right)$.
\end{corollary}
Corollaries \ref{cor:main_oasis} and \ref{cor:main_adam} together with of Lemma \ref{lem:pract_prec} give a complete picture for the 4 methods based on Algorithm \ref{Scaled_ExtraGrad} with Adam, RMSProp, AdaHessian and OASIS scalings.

\textbf{Discussion.} The following is a brief observation of our results.

$\bullet$ In our theoretical estimates, the convergence of methods with time-variable $\beta_t$ is worse than for methods with constant $\beta$. This is observed in other papers as well \cite{reddi2019convergence,defossez2020simple,dou2021one}. Typically, methods with time-variable $\beta_t$ perform their best in practical problems. But in our experiments (Appendix \ref{sec:exp}), the performance is the same; RMSProp outperforms Adam, and OASIS wins over AdaHessian.

$\bullet$ For all methods, we need to get the $\beta$ reasonably close to 1. The same is observed in Adam's analysis for minimization problems \cite{defossez2020simple}. Moreover, such a choice is confirmed in the experiments, both in ours and in the classical settings of Adam-based methods \cite{Kingma2015} (recommend $\beta = 0.999$).

$\bullet$ It is interesting to see how our estimates depend on the dimension $d=d_x + d_y$. This is relevant especially for large size problems. The dependence on $d$ appears in the constants $\Gamma$ and $C$ (Lemmas~\ref{lem:precond},\ref{lem:pract_prec}). For OASIS and AdaHessian, $\Gamma \sim \sqrt{d}$, but for Adam and RMSProp there is no dependence on $d$. This differs from the results of earlier papers \cite{defossez2020simple, liu2019towards, dou2021one}, where the convergence estimate for the Adam-based methods depends on $d$ more often linearly.

$\bullet$ As noted earlier, our analysis is universal for any scaled method for which Lemma \ref{lem:precond} holds. 

$\bullet$ Among the problems of analysis, let us highlight the use of $\| x_t \| \leq \Omega$, $\| y_t \| \leq \Omega$. Only in the strongly convex -- strongly concave case is it avoided. At the moment this assumption can be  found in the literature on saddle point problems, moreover, all the papers \cite{liu2019towards, dou2021one} on methods with scaling also use it.


\section{Make Scaled Methods More Robust}

\begin{algorithm*}[h!]
\caption{{\tt Scaled Stochastic Single Call Extra Gradient with Momentum}}
\label{Scaled_ExtraGrad_mom}
	\begin{algorithmic}[1]
		\State \textbf{Input:} initial point $x_0 = w_0^x, y_0 = w_0^y$, learning rate $\gamma$, probability $p \in [0;1]$
		\For{$t = 0,1,2,\ldots T-1$}
		\State Compute the preconditioners $\hat{D}^x_{t-1/2}$ and $\hat{D}^y_{t-1/2}$ 
		according to \eqref{eq:precond}--\eqref{eq:precond_abs}
		\State $x_{t+1/2} = x_t - \gamma (\hat D^x_{t-1/2})^{-1} \nabla_x f(x_{t-1/2}, y_{t-1/2}, \xi_{t-1/2})$ \label{lin14}
		\State $y_{t+1/2} = y_t + \gamma (\hat D^y_{t-1/2})^{-1} \nabla_y f(x_{t-1/2}, y_{t-1/2}, \xi_{t-1/2})$ \label{lin15}
		\State Generate $\xi_{t+1/2}$ for \text{\small{$\nabla_x f(x_{t+1/2}, y_{t+1/2}, \xi_{t+1/2})$, $\nabla_y f(x_{t+1/2}, y_{t+1/2}, \xi_{t+1/2})$ and $\hat{D}^x_{t+1/2}, \hat{D}^y_{t+1/2}$}}
		\State $x_{t+1} = x_t + \eta (\hat D^x_{t-1/2})^{-1}(w^x_t - x_t) - \gamma (\hat D^x_{t-1/2})^{-1} \nabla_x f(x_{t+1/2}, y_{t+1/2}, \xi_{t+1/2})$ \label{lin17}
		\State $y_{t+1} = y_t + \eta (\hat D^y_{t-1/2})^{-1} (w^y_t - y_t) + \gamma (\hat D^y_{t-1/2})^{-1} \nabla_y f(x_{t+1/2}, y_{t+1/2}, \xi_{t+1/2})$ \label{lin18}
		\State $w^x_{t+1}, w^y_{t+1}  = \begin{cases}
			x_{t}, y_{t},& \text{with probability }p\\
			w^x_{t}, w^y_{t},& \text{with probability }1-p
			\end{cases}$ \label{lin19}
		\EndFor
	\end{algorithmic}
\end{algorithm*}

Algorithm~\ref{Scaled_ExtraGrad} as well as Extra Gradient has an important disadvantage, two gradient calculations per iteration. The next method (Algorithm~\ref{Scaled_ExtraGrad_mom}) uses the so-called Optimistic approach \cite{popov1980modification}, which solves the problem of Extra Gradient mentioned above. The difference from Algorithm~\ref{Scaled_ExtraGrad} is that in the extrapolation step (lines \ref{lin14} and \ref{lin15}) we use the gradients from the previous main update (lines \ref{lin17} and \ref{lin18}) instead of the new gradients at point $x_t, y_t$. 
Also Algorithm~\ref{Scaled_ExtraGrad_mom} adds momentum as a useful tool (lines \ref{lin17} and \ref{lin18}). We use negative momentum with a random update (line \ref{lin19}). The same momentum is investigated in \cite{alacaoglu2021stochastic, kovalev2022optimal}. The usefulness of negative momentum for GANs training is discussed by \cite{gidel2019negative}. Both modifications of Algorithm \ref{Scaled_ExtraGrad_mom} are of a practical nature. From the theoretical point of view, they do not give any improvement in terms of convergence estimates, including the fact that saddle problems cannot be accelerated in terms of $\varepsilon$ as minimization problems \cite{ouyang2019lower}.

Similar to Algorithm \ref{Scaled_ExtraGrad}, we give a convergence theorem and its corollaries for Algorithm \ref{Scaled_ExtraGrad_mom}.
\begin{theorem} \label{th:main1}
Consider the problem \eqref{eq:minmax} under Assumptions~\ref{as:Lipsh} and \ref{as:var} and conditions of Lemma~\ref{lem:precond}. Let  $\{x_t, y_t\}$ and $\{x_{t+1/2}, y_{t+1/2}\}$ be the sequences generated by Algorithm \ref{Scaled_ExtraGrad_mom} after $T$ iterations. Then, with $\gamma \leq \frac{e}{10 L}$, $\eta \leq e p$ and $p \leq \frac{1}{4}$

$\bullet$ under Assumption \ref{as:conv}(SC) it holds that 
 \begin{equation}
 \label{eq:th1_sc}
 \begin{split}
     \E\left[\Psi_{t+1}\right] \leq&   \alpha_t \E\left[\Psi_{t}\right] + \frac{12\sigma^2 \left( 1 + (1-\beta_{t+1})C \right) \gamma^2}{e b},
 \end{split}
 \end{equation}
with $\alpha_t= \max \left\{\left(1 - \frac{\gamma \mu}{2 \Gamma} + (1-\beta_{t+1}) C \right); \left(1- \frac{\gamma \mu p}{2\eta + \gamma \mu}\right)\right\}$ and $\theta_t= \frac{12\sigma^2 \left( 1 + (1-\beta_{t+1})C \right) \gamma^2}{e b}$;

$\bullet$ under Assumption \ref{as:conv}(C) with $\| x_t \| \leq \Omega$ and $\| y_t \| \leq \Omega$ for all $t \geq 0$  it holds that
\begin{align}
\label{eq:th1_c}
\begin{split}
\E\big[\text{gap}&(x_{T}^{avg}, y_{T}^{avg})\big] 
\leq \frac{50\Gamma \Omega^2}{\gamma T}+ \frac{18C \Gamma \Omega^2}{\gamma T}\sum\limits_{t=1}^{T} (1-\beta_{t}) + \frac{55\gamma \sigma^2}{e b};
\end{split}
\end{align}

$\bullet$ under Assumption \ref{as:conv}(NC) with $\| x_t \| \leq \Omega$ and $\| y_t \| \leq \Omega$ for all $t \geq 0$ it holds that 
 \begin{equation}
 \begin{split}
\label{eq:th1_nc}
&\E[\| \nabla_x f(\bar x_{T-1/2}, \bar y_{T-1/2}) \|^2 
+ \| \nabla_y f(\bar x_{T-1/2}, \bar y_{T-1/2}) \|^2] \\
&\hspace{2cm}\leq \frac{96\Gamma^2 \Omega^2}{\gamma^2 T}  + \frac{48 \Gamma L^2 \Omega^2}{e T} + \frac{32 \Gamma^2 C \Omega^2}{\gamma^2 T} \sum\limits_{t=1}^{T} (1-\beta_{t}) +\frac{105 \Gamma \sigma^2}{e b}.
 \end{split}
 \end{equation}
 Here we use $\Psi_{t+1}$ defined in \eqref{eq:psi}, $(\bar x_{T-1/2}, \bar y_{T-1/2})$ is picked uniformly from all $(x_{t-1/2}, y_{t-1/2})$. Other notations are similar to Theorem \ref{th:main0}.
\end{theorem}
\begin{corollary}[RMSProp and OASIS] \label{cor:main1_oasis}
Under the assumptions of Theorem \ref{th:main1} let additionally take $\beta_t \equiv \beta$. Then, the number of iterations to achieve $\varepsilon$-solution (in expectation)

$\bullet$ in the strongly convex--strongly concave case with $\beta \geq 1 - \frac{\gamma \mu}{ 4 \Gamma C}$ is $\mathcal{\tilde O} \left(\left[ p + \frac{\Gamma L}{e \mu}\right] \log \frac{1}{\varepsilon}  + \frac{\Gamma^2 \sigma^2}{e b \mu^2  \varepsilon}\right)$;

$\bullet$ in the convex--concave case with $\beta \geq 1 - \frac{1}{C T}$ is $\mathcal{O}\left(\frac{\Gamma L\Omega^2}{e \varepsilon} + 
 \frac{\Gamma\sigma^2 \Omega^2}{e b\varepsilon^2} \right)$;
 
$\bullet$ in the non-convex--non-concave case with $\beta \geq 1 - \frac{1}{C T}$ and $b \sim \frac{\Gamma \sigma^2}{e \varepsilon^2}$ is $\mathcal{O}\left( \frac{\Gamma^2 L^2 \Omega^2}{e^2 \varepsilon^2} \right)$.
\end{corollary}

\begin{corollary}[Adam and AdaHessian] \label{cor:main1_adam}
Under the assumptions of Theorem \ref{th:main1}, let additionally take $\beta_t = \frac{\beta - \beta^{t+1}}{1 - \beta^{t+1}}$. Then, the complexities to achieve $\varepsilon$-solution (in expectation)

$\bullet$ in the str. convex--str. concave case with $\beta = 1 - \left(\frac{\gamma \mu}{ 4 \Gamma C}\right)^2$ is $\mathcal{O}(\frac{C^2\Gamma^3 L^2 \Omega^2 }{e^3 \mu^2 \varepsilon} + \frac{ C^2\Gamma^2 L^2 \Omega^2 }{e^2 \mu^2 \varepsilon} + \frac{C^2\Gamma^2  L \Omega^2 }{e^2 p \mu \varepsilon} + \frac{C^4 \Gamma^5 \sigma^4 \Omega^2}{e^5 \mu^4  b^2 \varepsilon^3}  + \frac{ C^4 \Gamma^2 \sigma^4 \Omega^2}{e^2 \mu^4 b^2 \varepsilon^3}  + (\frac{C^5 \Gamma^4 \sigma^2 \Omega^4}{e^4 \mu^2 p^3 b \varepsilon^3})^{1/2} )$;

$\bullet$ in the convex--concave case with $\beta = 1 - \frac{1}{T}$  is $\mathcal{O}\left(\frac{C^2\Gamma^2 L^2\Omega^4}{e^2 \varepsilon^2 } + \frac{C^2\Gamma^2 \sigma^4 \Omega^4}{e^2 b^2 \varepsilon^4} \right)$;

$\bullet$ in the non-convex--non-concave case with $\beta = 1 - \frac{1}{T}$ and $b \sim \frac{\Gamma \sigma^2}{e \varepsilon^2}$ is $\mathcal{O}\left(\frac{C^2 \Gamma^4 L^4 \Omega^4}{e^4 \varepsilon^4}\right)$.
\end{corollary}
The estimates obtained are very similar to Section \ref{sec:scaled}. The same observations can be made about these results as those given in Section \ref{sec:scaled} after Corollary \ref{cor:main_adam}.

\section{Experiments} \label{sec:exp}
This section provides extensive experiments and compares the considered optimization methods for training GANs.

\subsection{Setup}

\textbf{Datasets.} We use the standard dataset for training GANs,  CIFAR-10 \cite{krizhevsky2009learning} that has 50k of 32x32 resolution images in the training set and 10k  images in the test.

\textbf{Metrics.} We use Inception score (IS) \cite{salimans2016improved} and Fre\'chet Inception distance (FID) \cite{heusel2017gans} as the most common metrics for assessing image synthesis models. 

\textbf{GAN architectures.} We implemented ResNet architectures for the generator and the discriminator from \cite{miyato2018spectral}. See details in Appendix \ref{sec:exp_app}. 

\textbf{Optimization methods.} We employ RMSProp, Adam, AdaHessian, and OASIS, its extragradient versions Extra RMSProp, Extra Adam, Extra AdaHessian, and Extra OASIS. Additionally, we implement OASIS with momentum and Extra OASIS with momentum. We compute the uniform average and exponential moving average (EMA) of weights as well-known techniques to yield faster convergence. The hyperparameter settings of the methods are given in Appendices \ref{sec:exp_app}.

\subsection{Results}

The goal of the first experiment is to investigate the four methods for which we give a theory. These include the popular and one of the best-known methods, Adam (the benchmark for the other competitors), the known, but less popular RMSProp, and the very new and never used for GANs training methods with scaling (OASIS and AdaHessian).

\begin{figure*}[h!]
\centering
  \includegraphics[width=0.8\linewidth]{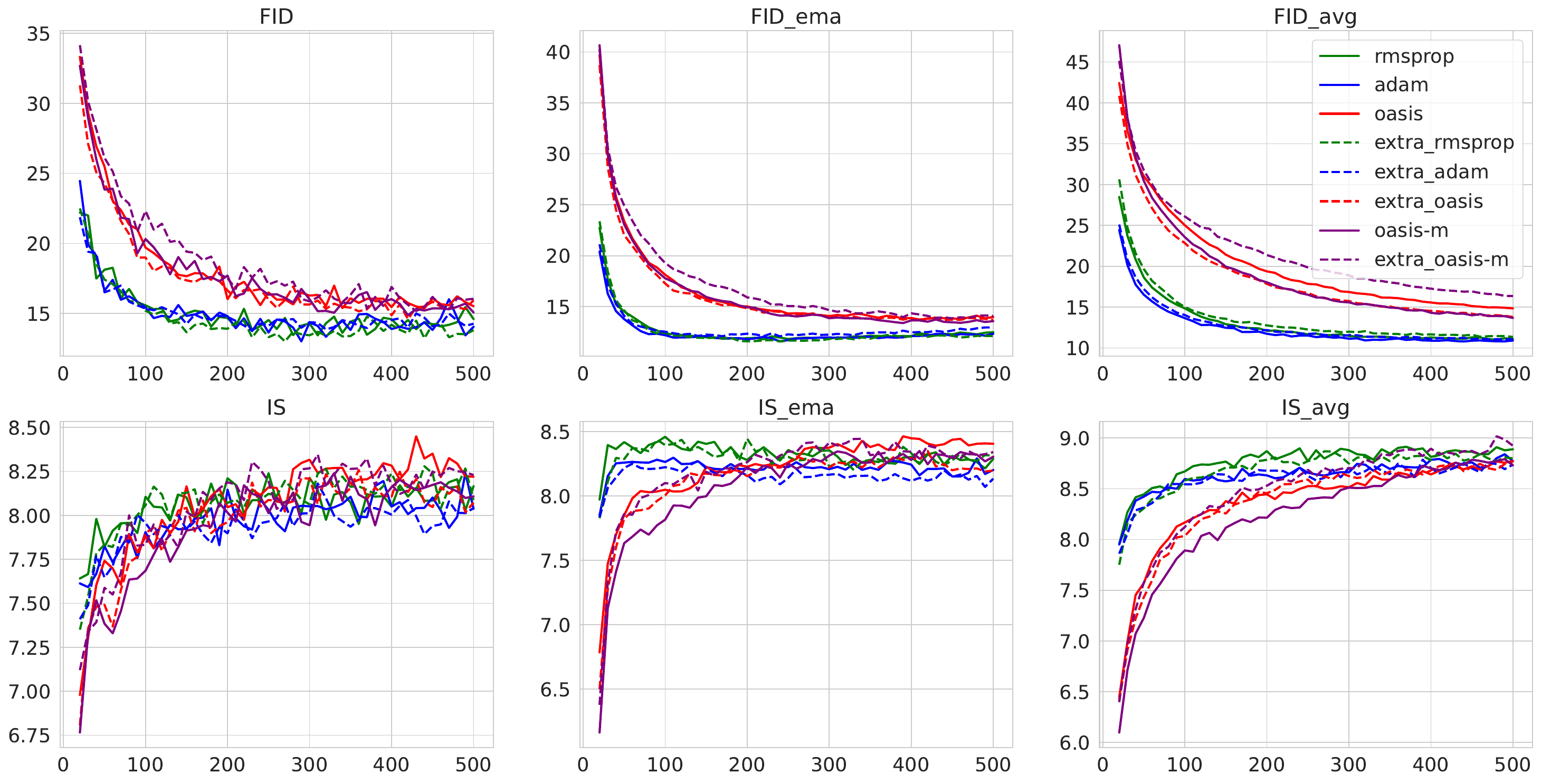}
  \caption{Averaged dynamics of FID and IS change by epoch number for different optimizers from Tables \ref{tab:res1} and \ref{tab:res2}.}
  \label{fig:metrics}
\end{figure*}

For each of the optimizers, we have implemented two versions: classical and with extra step (both Algorithm \ref{Scaled_ExtraGrad} and Algorithm \ref{Scaled_ExtraGrad_mom} have extra steps). We compare how the qualities of the four different scaling approaches relate to each other and also verify the popular idea in the community that using extra step is a necessary trick to increase the learning quality \cite{mertikopoulos2018optimistic, gidel2018variational, chavdarova2019reducing, liu2019towards, dou2021one}. See results in Table \ref{tab:res1} and Figure \ref{fig:metrics}.

We see that three optimizers out of four performed well. The not very popular scaling techniques OASIS and RMSProp perform better than state-of-the-art Adam. The use of the extra step technique can provide improvement but is not an important element for the optimizer.

In the second group of experiments, we decided to investigate the OASIS scaling technique in more detail, as it is new to GANs in comparison to Adam and RMSProp, and also showed good results in the first group of experiments. In particular, we explore how OASIS is affected by negative momentum and the combination of extra step and negative momentum (Algorithm \ref{Scaled_ExtraGrad_mom}). See results in Table \ref{tab:res2} and Figure~\ref{fig:metrics}.

We can see that negative momentum can indeed increase the quality of learning, sometimes its combination with extra step is also useful. 

\begin{table*}[h!] 
\centering
\caption{Comparisons of the considered optimization methods and their extragradient versions using FID and IS, scores obtained at the end of training. EMA and uniform avg denote as exponential moving average and uniform averaging, respectively. Results are averaged over 5 runs. We run each experiment for 500k iterations. For each column the best score is in bold along with any score within its standard deviation reach.}
\label{tab:res1}
\resizebox{0.8\linewidth}{!}
{
  \begin{tabular}{ccccccc}
      \toprule
      & \multicolumn{3}{c}{Fr\'echet Inception distance $\downarrow$}  & \multicolumn{3}{c}{Inception score $\uparrow$}                \\
    \cmidrule(r){2-4} \cmidrule(r){5-7}
      \textbf{CIFAR-10}     & no avg   & EMA &  uniform avg  & no avg   & EMA &  uniform avg \\
      \midrule\midrule
      RMSProp & 14.54 $\pm$ 0.55 & 12.42 $\pm$ 0.08 & 11.14 $\pm$ 0.10 & 8.15 $\pm$ 0.07 & 8.27 $\pm$ 0.03 & \textbf{8.87 $\pm$ 0.03} \\
      Adam & 14.57 $\pm$ 0.96 & 12.27 $\pm$ 0.07 & \textbf{10.87 $\pm$ 0.05} & 8.05 $\pm$ 0.12 & 8.20 $\pm$ 0.05 & 8.77 $\pm$ 0.06 \\
      AdaHessian & 39.54 $\pm$ 4.43 & 37.71 $\pm$ 0.20 & 53.96 $\pm$ 0.19 & 6.92 $\pm$ 0.06 & 6.85 $\pm$ 0.06 & 6.52 $\pm$ 0.05 \\
      OASIS & 15.76 $\pm$ 0.28 & 13.93 $\pm$ 0.13 & 14.96 $\pm$ 0.08 & \textbf{8.26 $\pm$ 0.05} & \textbf{8.41 $\pm$ 0.02} & 8.74 $\pm$ 0.03\\
      \midrule
      Extra RMSProp & \textbf{13.68 $\pm$ 0.37} & \textbf{12.08 $\pm$ 0.07} & 11.42 $\pm$ 0.05 & 8.14 $\pm$ 0.09 & 8.30 $\pm$ 0.02 & 8.80 $\pm$ 0.02 \\
      Extra Adam & 14.41 $\pm$ 0.36 & 12.87 $\pm$ 0.10 & 11.09 $\pm$ 0.03 & 8.02 $\pm$ 0.05 & 8.13 $\pm$ 0.04 & 8.72 $\pm$ 0.04 \\
      Extra AdaHessian & 40.11 $\pm$ 0.50 & 40.51 $\pm$ 0.43 & 55.46 $\pm$ 0.38 & 6.72 $\pm$ 0.06 & 6.66 $\pm$ 0.04 & 5.98 $\pm$ 0.03 \\
      Extra OASIS & 15.51 $\pm$ 0.25 & 13.79 $\pm$ 0.09 & 13.95 $\pm$ 0.09 & 8.11 $\pm$ 0.06 & 8.20 $\pm$ 0.01 & 8.73 $\pm$ 0.03 \\
      \bottomrule
    \end{tabular}
    }
\centering
\vskip8pt
\caption{Comparisons of the OASIS and OASIS with momentum (OASIS-M) optimization methods and their extragradient versions using FID and IS, scores obtained at the end of training. EMA and uniform avg denote as exponential moving average and uniform averaging, respectively. Results are averaged over 5 runs. We run each experiment for 500k iterations. For each column the best score is in bold along with any score within its standard deviation reach.}
\vskip5pt
\label{tab:res2}
\resizebox{0.8\linewidth}{!}{
  \begin{tabular}{ccccccc}
      \toprule
      & \multicolumn{3}{c}{Fr\'echet Inception distance $\downarrow$}  & \multicolumn{3}{c}{Inception score $\uparrow$}                \\
    \cmidrule(r){2-4} \cmidrule(r){5-7}
      \textbf{CIFAR-10}     & no avg   & EMA &  uniform avg  & no avg   & EMA &  uniform avg \\
      \midrule\midrule
      OASIS & 15.76 $\pm$ 0.28 & 13.93 $\pm$ 0.13 & 14.96 $\pm$ 0.08 & \textbf{8.26 $\pm$ 0.05} & \textbf{8.41 $\pm$ 0.02} & 8.74 $\pm$ 0.03\\
      OASIS-M & \textbf{15.38 $\pm$ 0.23} & \textbf{13.54 $\pm$ 0.10} & 1\textbf{3.91 $\pm$ 0.17} & 8.14 $\pm$ 0.04 & 8.31 $\pm$ 0.03 & 8.77 $\pm$ 0.02 \\
      Extra OASIS & \textbf{15.51 $\pm$ 0.25} & 13.79 $\pm$ 0.09 & \textbf{13.95 $\pm$ 0.09} & 8.11 $\pm$ 0.06 & 8.20 $\pm$ 0.01 & 8.73 $\pm$ 0.03 \\
      Extra OASIS-M & 16.10 $\pm$ 0.70 & 14.32 $\pm$ 0.22 & 17.70 $\pm$ 0.27 & \textbf{8.26 $\pm$ 0.06} & 8.35 $\pm$ 0.04 & \textbf{8.86 $\pm$ 0.03} \\
      \bottomrule
    \end{tabular}
    }
\end{table*}

\subsection*{Acknowledgments}

This research was supported in part through computational resources of HPC facilities at HSE University.

\bibliography{refs}
\newpage
\appendix
\onecolumn

\section{EXPERIMENTS} \label{sec:exp_app}
\subsection{Implementation Details}
In our implementation of experiments we use the PyTorch\footnote{\hyperlink{https://pytorch.org}{https://pytorch.org}} framework. For the computation of the FID and IS metrics we utilize the the package \emph{pytorch\_gan\_metrics}\footnote{\hyperlink{https://pypi.org/project/pytorch-gan-metrics/}{https://pypi.org/project/pytorch-gan-metrics/}} that reproduces the original Tensorflow implementation in PyTorch. We provide all source code of our experiments as a part of supplementary material. We also attach configuration files for each model to run its training with specified hyperparameters. 

\subsection{Architectures}
\begin{figure}[ht]
	\begin{tabular}{cc}
    	\begin{minipage}{.2\textwidth}
            \centering
             \includegraphics[width=1.0\textwidth]{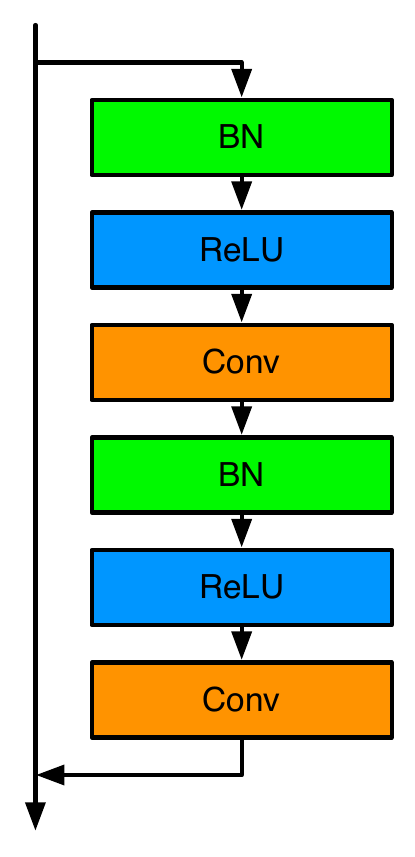}
            \figcaption{\label{fig:resblock} ResBlock architecture.
             For the discriminator we removed BN layers in ResBlock.}
        \end{minipage}
        \begin{minipage}{.75\textwidth}
          \tblcaption{\label{tab:resnets_cifar10}ResNet architectures for CIFAR10 dataset. We use similar architectures to the ones used in~\cite{miyato2018spectral}. }
          \centering
          \begin{subtable}{.375\textwidth}
              \centering
              {\begin{tabular}{c}
                  \toprule
                  \midrule
                  $z\in \R^{128} \sim \mathcal{N}(0, I)$ \\
                  \midrule
                  dense, $4 \times 4 \times 256$ \\
                  \midrule
                  ResBlock up 256\\
                  \midrule
                  ResBlock up 256\\
                  \midrule
                  ResBlock up 256\\
                  \midrule
                  BN, ReLU, 3$\times$3 conv, 3 Tanh\\
                  \midrule
                  \bottomrule
              \end{tabular}}
              \caption{Generator}
          \end{subtable}\hspace{15mm}
          \begin{subtable}{.375\textwidth}
              \centering
              {\begin{tabular}{c}
                  \toprule
                  \midrule
                  RGB image $x\in \R^{32\times 32 \times 3}$ \\
                  \midrule
                  ResBlock down 128\\
                  \midrule
                  ResBlock down 128\\
                  \midrule
                  ResBlock 128\\
                  \midrule
                  ResBlock 128\\
                  \midrule
                  ReLU\\
                  \midrule
                  Global sum pooling\\
                  \midrule
                  dense $\rightarrow$ 1\\
                  \midrule
                  \bottomrule
              \end{tabular}}
              \caption{Discriminator}
          \end{subtable}
        \end{minipage}
    \end{tabular}
\end{figure}

We replicate the experimental setup as in \cite{miyato2018spectral, chavdarova2019reducing}. We use ResNet architectures for the generator $G$ and the discriminator $D$ as described in Table \ref{tab:resnets_cifar10}. 

\subsection{Training Losses}
For training we use the hinge version of the non-saturating adversarial loss as in \cite{miyato2018spectral}:

\begin{align}
	 \mathcal{L}_D &= \E_{\bm x\sim q_{\rm data}(\bm x)}\left[{\rm min}\left(0, -1+D(\bm x)\right)\right] + \E_{\bm z\sim p(\bm z)} \left[{\rm min}\left(0, -1-D\left(G(\bm z)\right)\right)\right]\\
     \mathcal{L}_G &= -\E_{\bm z\sim p(\bm z)}\left[D\left(G(\bm z)\right)\right], \label{eq:hinge}
\end{align}

\subsection{Hyperparameters}
Table \ref{tab:hypers} lists the hyperparameters we used to obtain the reported results on CIFAR-10 dataset. These values were tuned for each method independently. $D\_steps$ means the number of the discriminator steps per one generator step. $eps$ denotes the threshold in OASIS algorithm that is used to clip the diagonal elements of the Hessian. $hess\_upd\_each$ means how often we update the diagonal elements of the Hessian. 

\begin{table}[h]
\centering
\caption{Hyperparameters we used for CIFAR-10 dataset.}
\label{tab:hypers}
\resizebox{\linewidth}{!}{
  \begin{tabular}{ccccccccc}
      \toprule
      & \multicolumn{8}{c}{\textbf{Hyperparameters}}\\
    \cmidrule(r){2-9}
      \textbf{Method}     & $lr_G$   & $lr_D$ &  $\beta_1$  & $\beta_2$   & $D\_steps$ &  batch size & $eps$ & $hess\_upd\_each$ \\
      \midrule\midrule
      RMSProp & 0.0002 & 0.0002 & -- & 0.999 & 5 & 64 & -- & -- \\
      Adam & 0.0002& 0.0002 & 0.0 & 0.999 & 5 & 64 & -- & -- \\
      AdaHessian & 0.0002& 0.0002 & 0.0 & 0.999 & 5 & 64 & -- & 20 \\
      OASIS & 0.0002& 0.0002 & 0.0 & 0.999 & 5 & 64 & 0.01 & 20 \\
      \midrule
      Extra RMSProp & 0.0002 & 0.0002 & -- & 0.999 & 5 & 64 & -- & -- \\
      Extra Adam & 0.0002& 0.0002 & 0.0 & 0.999 & 5 & 64 & -- & -- \\
      Extra AdaHessian & 0.0002& 0.0002 & 0.0 & 0.999 & 5 & 64 & -- & 20 \\
      Extra OASIS & 0.0002& 0.0002 & 0.0 & 0.999 & 5 & 64 & 0.01 & 20 \\
      \midrule
      OASIS-M & 0.0002& 0.0002 & 0.1 & 0.999 & 5 & 64 & 0.01 & 20 \\
      Extra OASIS-M & 0.0002& 0.0002 & 0.1 & 0.999 & 5 & 64 & 0.01 & 20 \\
      \bottomrule
    \end{tabular}
    }
\end{table}

\subsection{Total Amount of Compute Resources}
We run our experiments on Tesla V100 GPUs. We used approximately 2500 GPU hours to obtain the reported results. 

\subsection{Samples of the Trained Generators}
\begin{figure}[h]
\centering
  \includegraphics[width=0.3\linewidth]{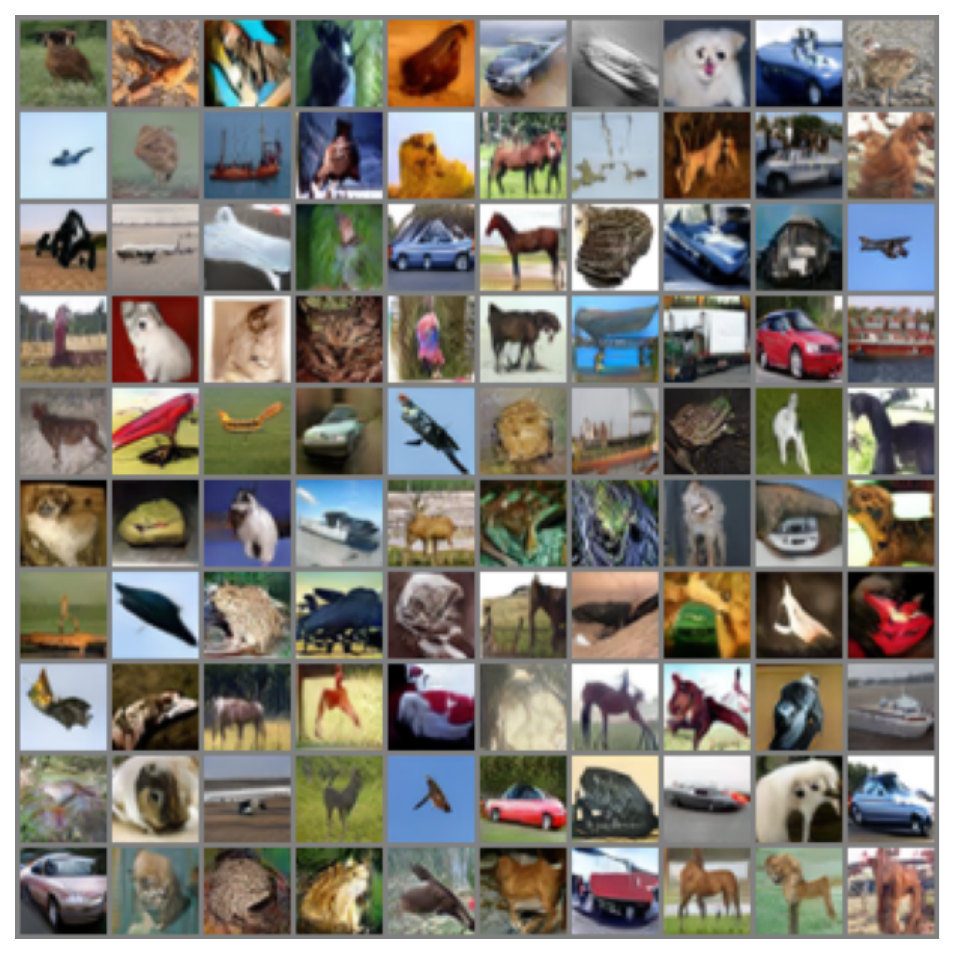}
  \caption{Samples produced by the generator trained by the RMSProp algorithm using uniform average on the weights.}
  \label{fig:rmsprop_avg}
\end{figure}

\begin{figure}[h]
\centering
  \includegraphics[width=0.3\linewidth]{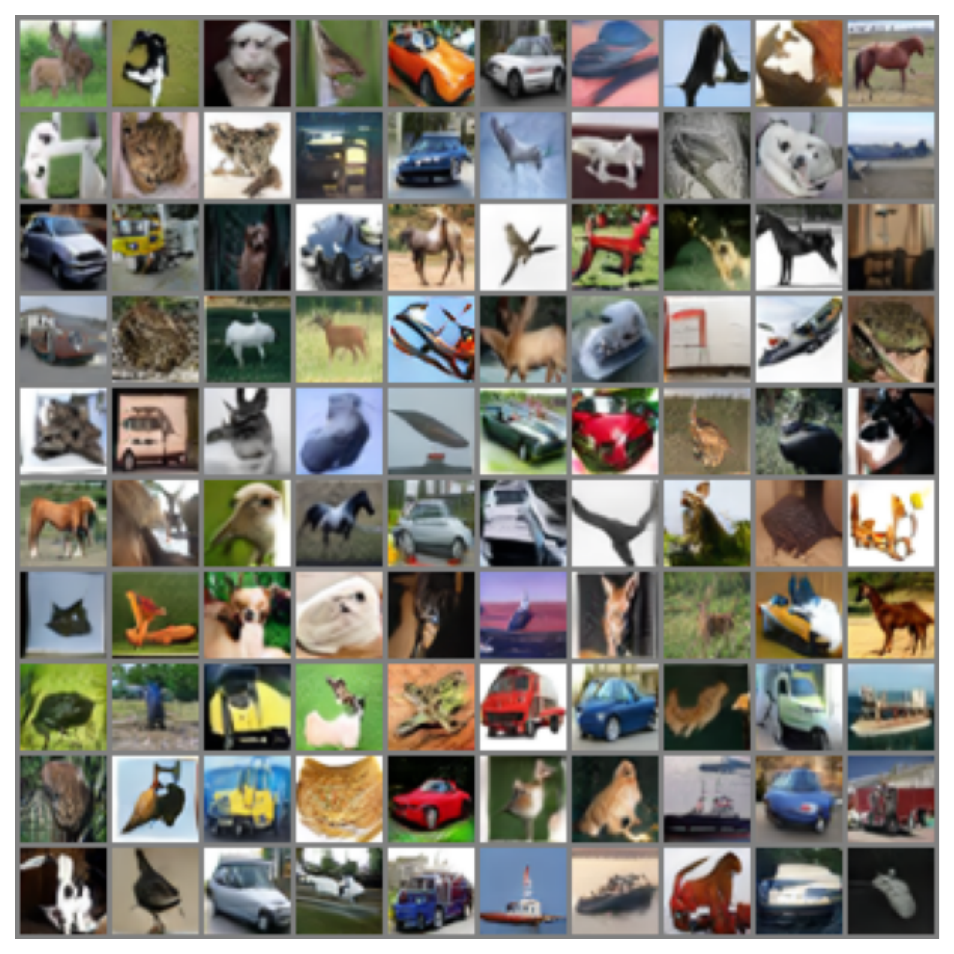}
  \caption{Samples produced by the generator trained by the Adam algorithm using uniform average on the weights.}
  \label{fig:adam_avg}
\end{figure}

\begin{figure}[h]
\centering
  \includegraphics[width=0.3\linewidth]{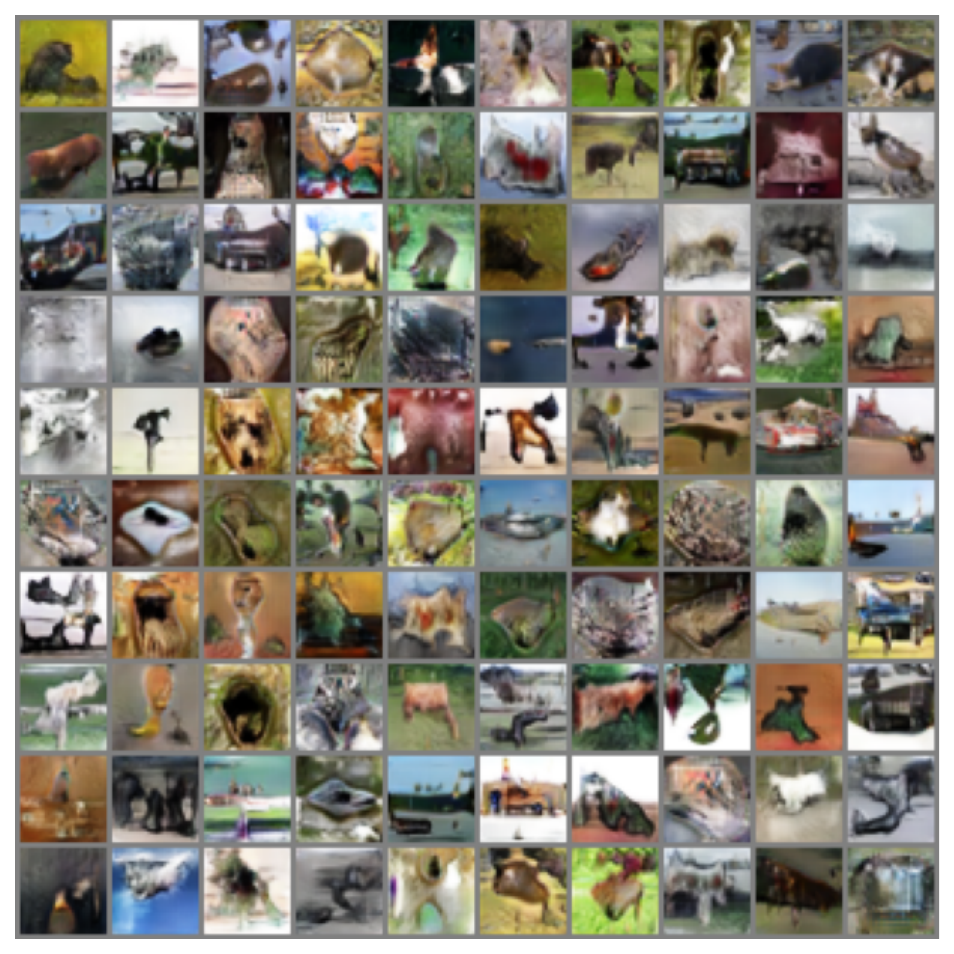}
  \caption{Samples produced by the generator trained by the AdaHessian algorithm using exponential moving average on the weights.}
  \label{fig:adahess_ema}
\end{figure}

\begin{figure}[h]
\centering
  \includegraphics[width=0.3\linewidth]{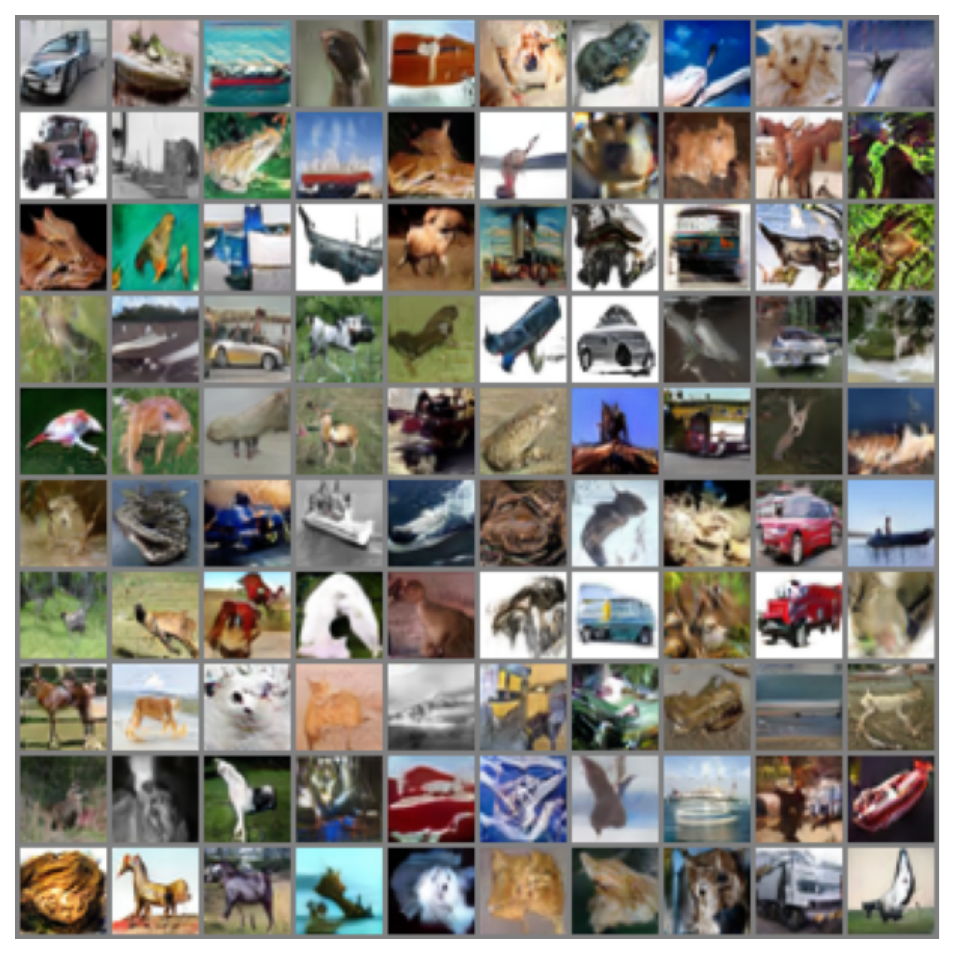}
  \caption{Samples produced by the generator trained by the OASIS algorithm using exponential moving average on the weights.}
  \label{fig:oasis_ema}
\end{figure}

\begin{figure}[h]
\centering
  \includegraphics[width=0.3\linewidth]{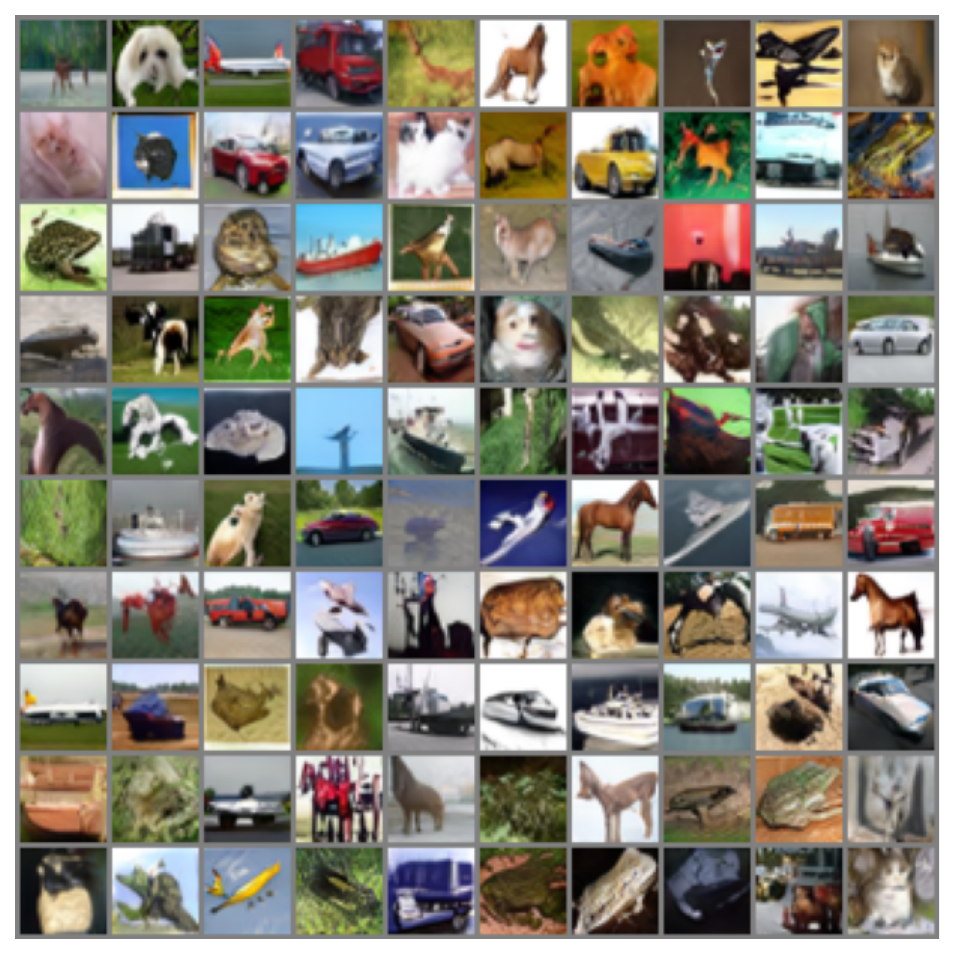}
  \caption{Samples produced by the generator trained by the Extra RMSProp algorithm using uniform average on the weights.}
  \label{fig:extra_rmsprop_avg}
\end{figure}

\begin{figure}[h]
\centering
  \includegraphics[width=0.3\linewidth]{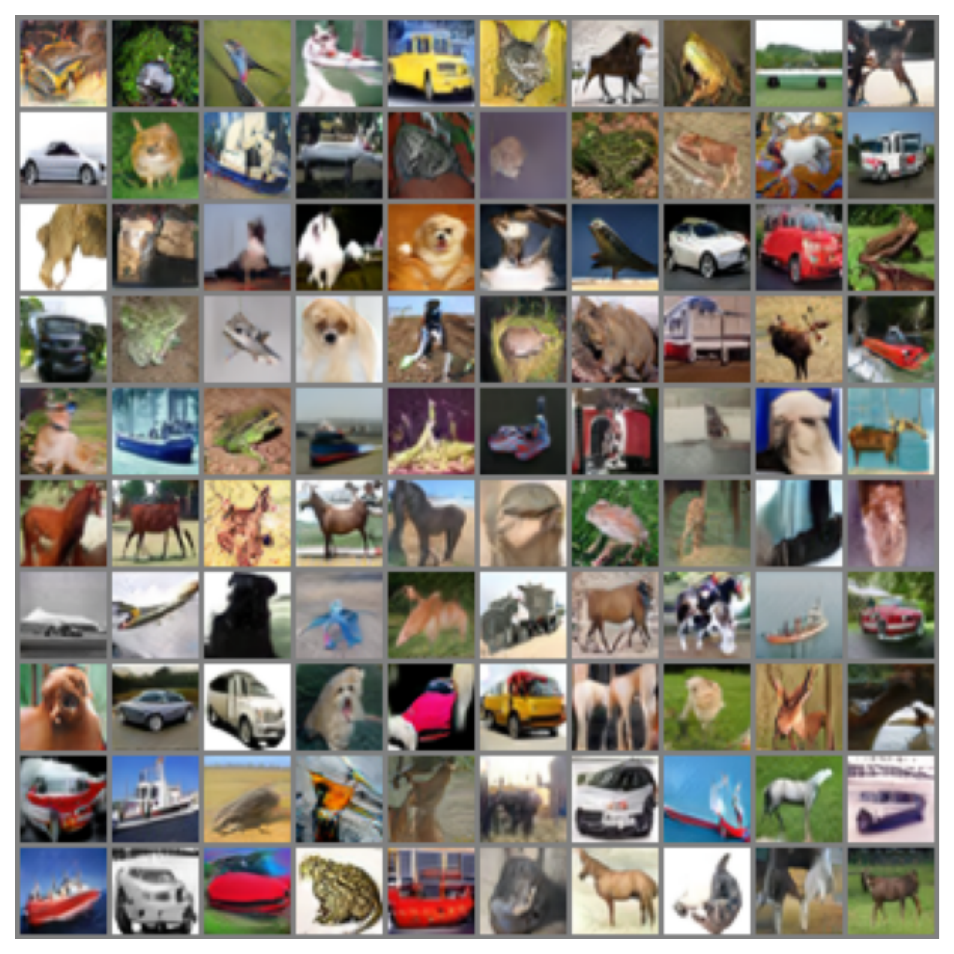}
  \caption{Samples produced by the generator trained by the Extra Adam algorithm using uniform average on the weights.}
  \label{fig:extra_adam_avg}
\end{figure}

\begin{figure}[h]
\centering
  \includegraphics[width=0.3\linewidth]{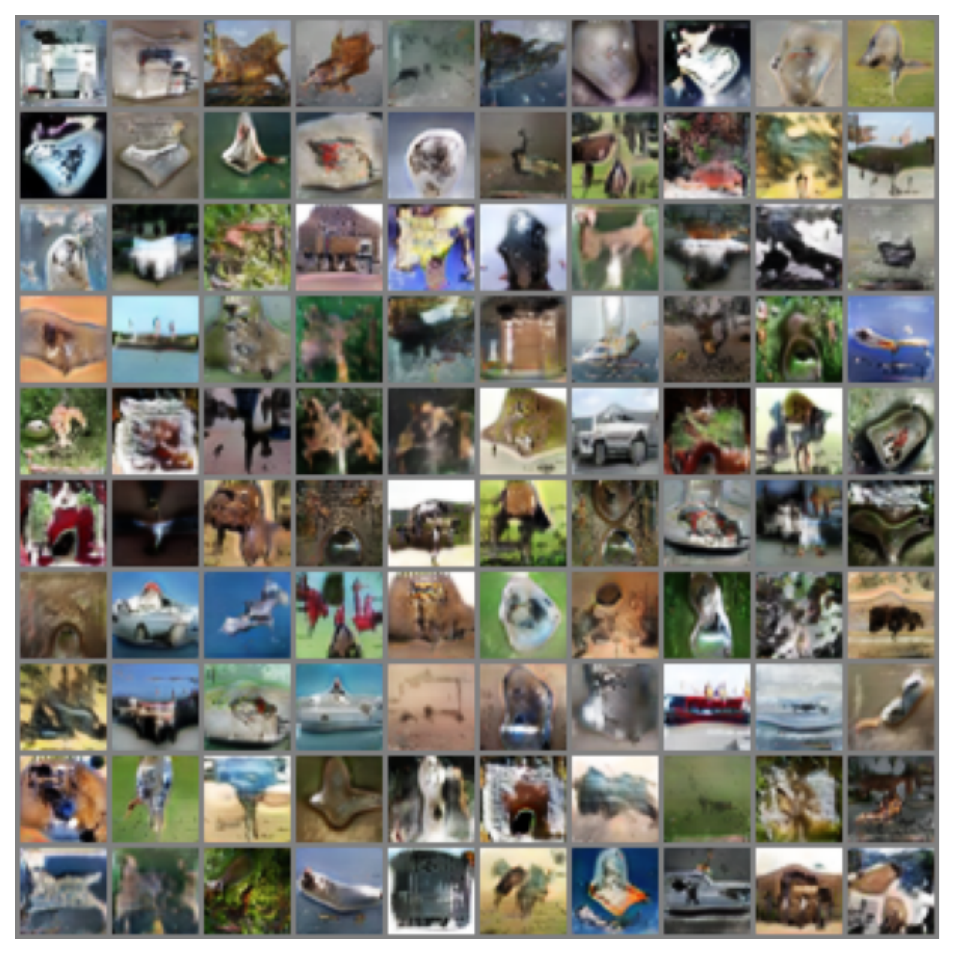}
  \caption{Samples produced by the generator trained by the Extra AdaHessian algorithm using exponential moving average on the weights.}
  \label{fig:extra_adahess_ema}
\end{figure}

\begin{figure}[h]
\centering
  \includegraphics[width=0.3\linewidth]{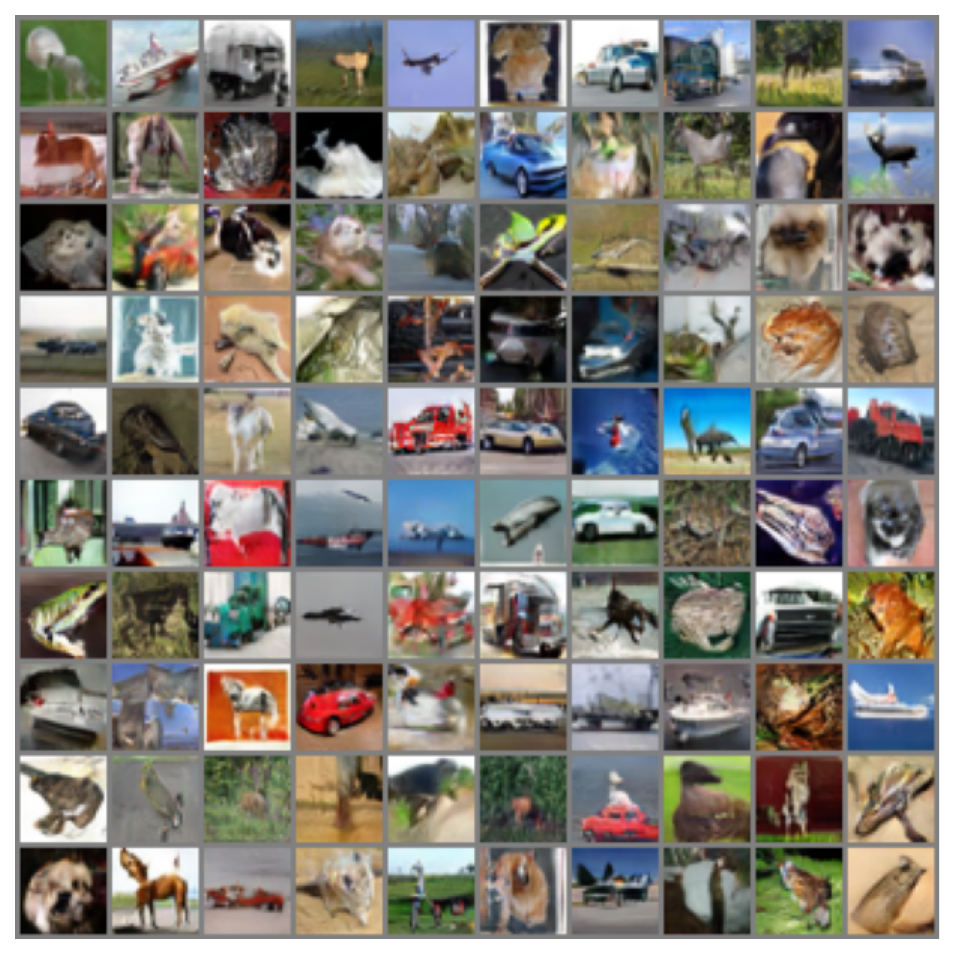}
  \caption{Samples produced by the generator trained by the Extra OASIS algorithm using exponential moving average on the weights.}
  \label{fig:extra_oasis_ema}
\end{figure}

\begin{figure}[h]
\centering
  \includegraphics[width=0.3\linewidth]{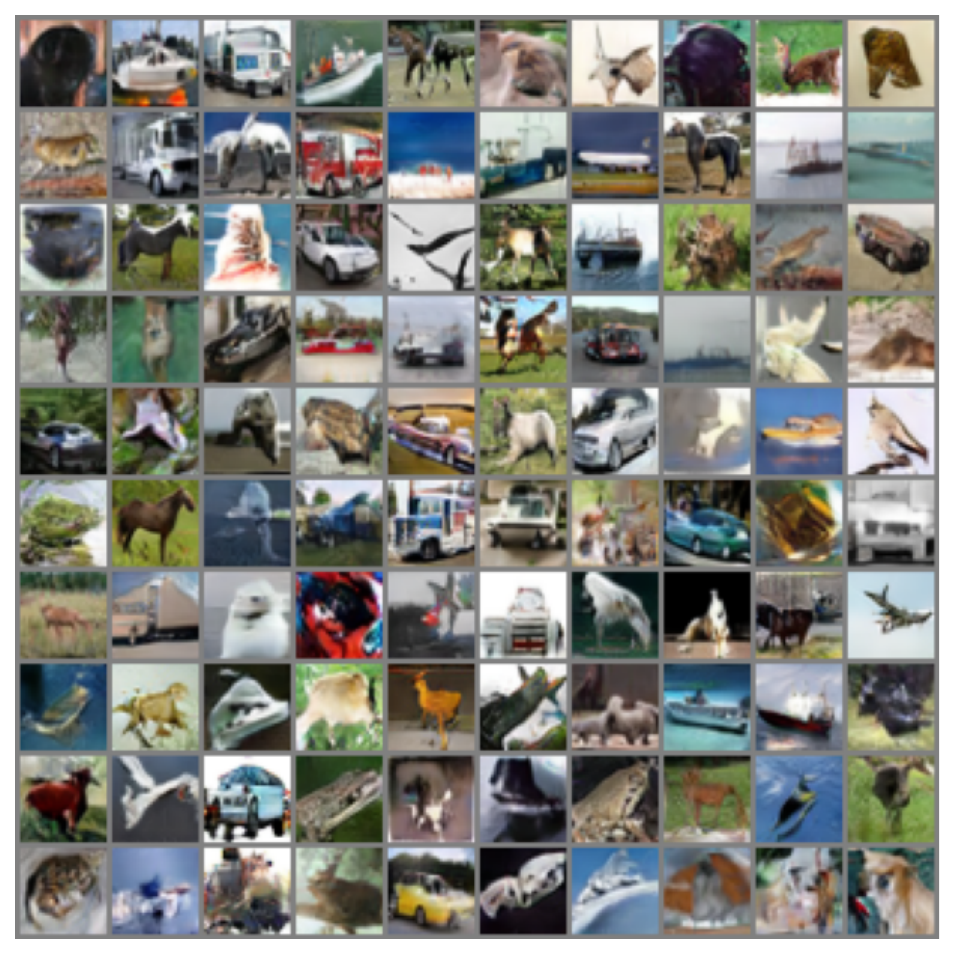}
  \caption{Samples produced by the generator trained by the OASIS-M algorithm using exponential moving average on the weights.}
  \label{fig:oasis_m_ema}
\end{figure}

\begin{figure}[h]
\centering
  \includegraphics[width=0.3\linewidth]{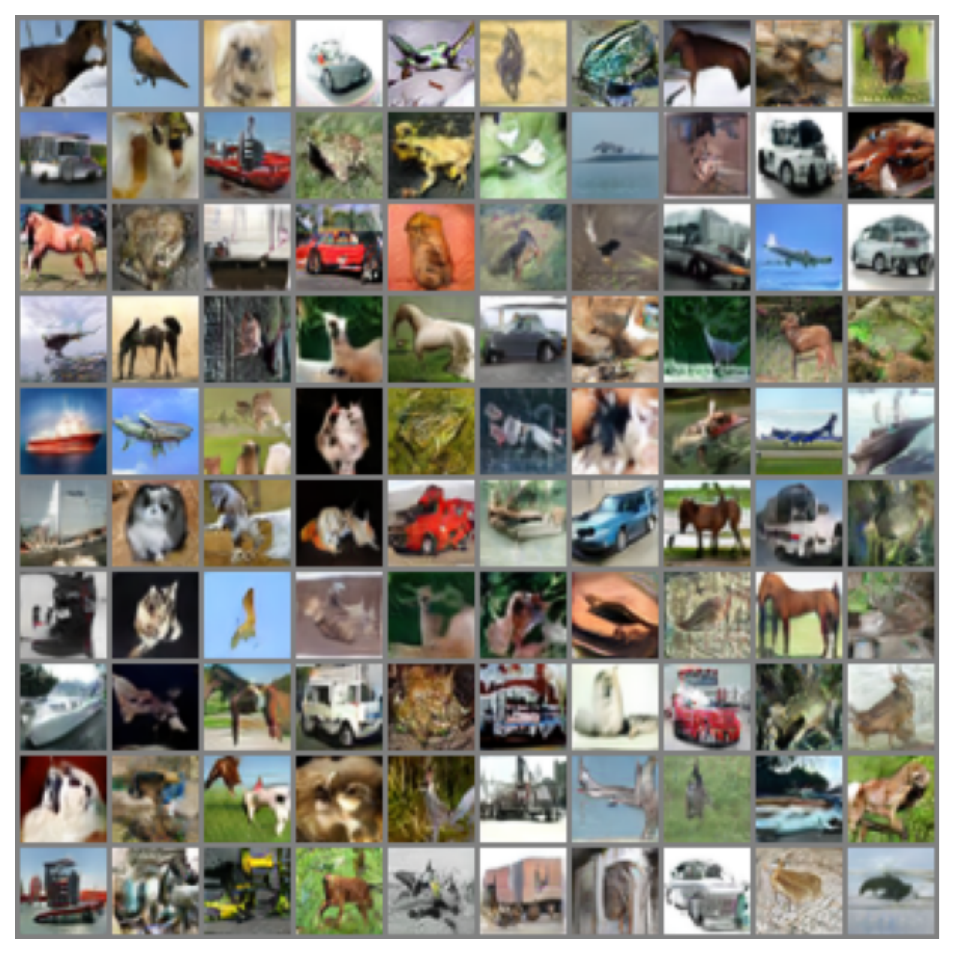}
  \caption{Samples produced by the generator trained by the Extra OASIS-M algorithm using exponential moving average on the weights.}
  \label{fig:extra_oasis_m_ema}
\end{figure}

We show random samples of the trained generators for each method. We chose the best averaging strategy for each algorithm based on the visual quality. The provided samples are in Figures \ref{fig:rmsprop_avg}-\ref{fig:extra_oasis_m_ema}. 

\clearpage

\section{TECHNICAL LEMMA}

\begin{lemma}
For any $T \in \mathbb{N}$, we get
\begin{equation}
    \label{eq:tech_lem}
    \left( 1 - \frac{1}{T}\right)^{\sqrt{T}} \leq 1 - \frac{1}{2 \sqrt{T}}.
\end{equation}
\end{lemma}
\textbf{Proof:}
We prove it by \href{https://www.wolfram.com/mathematica/}{Wolfram Mathematica}:
\begin{figure}[!h]
\centering
\includegraphics[width=0.68\textwidth]{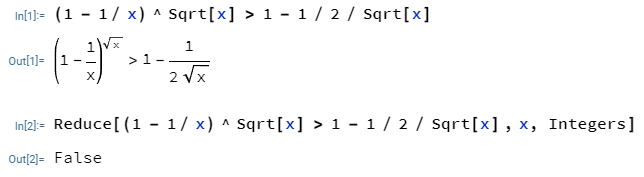}
\end{figure}

$\square$

\section{MISSING PROOFS}

\subsection{Proof of Lemma \ref{lem:precond}}

1) It is easy to see that due to the fact that all matrices $H^x_t$ and $D^x_0$ are diagonal, then based on rules \eqref{eq:precond} and \eqref{eq:precond_add}, we can conclude that all matrices $D^x_t$ are also diagonal. Again from rules \eqref{eq:precond}, \eqref{eq:precond_add} and $\beta_t \in [0;1]$, we get that all elements of $D^x_t$ matrices are not greater than $\Gamma$ in absolute value. \eqref{eq:precond_abs} gives that $\hat D^x_t$ matrices are diagonal and their elements from $e$ to $\Gamma$. This completes the proof. Similar proof for the variable $y$. $\square$




2) 
We start from simple steps:
\begin{align*}
  e I \preccurlyeq \hat D^x_{t+1}  = \hat D^x_{t} + \hat D^x_{t+1} -  \hat D^x_{t} = \hat D^x_{t} + ( \hat D^x_{t+1} -  \hat D^x_{t})(\hat D^x_{t+1} +  \hat D^x_{t})(\hat D^x_{t+1} +  \hat D^x_{t})^{-1}.
\end{align*}
Since matrices $\hat D^x_{t+1}$ and $\hat D^x_{t}$ are diagonal, we have that $\hat D^x_{t+1} \hat D^x_{t} =\hat D^x_{t} \hat D^x_{t+1}$ and then
\begin{align*}
  e I &\preccurlyeq \hat D^x_{t+1}  = \hat D^x_{t} + ( (\hat D^x_{t+1})^2 -  (\hat D^x_{t})^2)(\hat D^x_{t+1} +  \hat D^x_{t})^{-1} \\
  &\preccurlyeq \hat D^x_{t} + \| (\hat D^x_{t+1})^2 -  (\hat D^x_{t})^2 \|_{\infty} \|(\hat D^x_{t+1} +  \hat D^x_{t})^{-1}\|_{\infty} I.
\end{align*}
All diagonal elements of  $\hat D^x_{t+1}$ and $\hat D^x_{t}$ are positive from $e$ to $\Gamma$, it means that
$\|(\hat D^x_{t+1} +  \hat D^x_{t})^{-1}\|_{\infty} \leq \frac{1}{2e}$. Hence,
\begin{align*}
  e I &\preccurlyeq \hat D^x_{t+1} \preccurlyeq \hat D^x_{t} + \frac{1}{2e} \| (\hat D^x_{t+1})^2 -  (\hat D^x_{t})^2 \|_{\infty} I.
\end{align*}
One can note the following:

$\bullet$ if $((D^x_{t+1})^2)_{ii} \geq e^2$ and $((D^x_{t})^2)_{ii} \geq e^2$ then $|((\hat D^x_{t+1})^2)_{ii} - ((\hat D^x_{t})^2)_{ii}| = |(( D^x_{t+1})^2)_{ii} - ((D^x_{t})^2)_{ii}|$;

$\bullet$ if $((D^x_{t+1})^2)_{ii} \geq e^2$ and $((D^x_{t})^2)_{ii} < e^2$ then $|((\hat D^x_{t+1})^2)_{ii} - ((\hat D^x_{t})^2)_{ii}| \leq |(( D^x_{t+1})^2)_{ii} - ((D^x_{t})^2)_{ii}|$;

$\bullet$ if $((D^x_{t+1})^2)_{ii} < e^2$ and $((D^x_{t})^2)_{ii} \geq e^2$ then $|((\hat D^x_{t+1})^2)_{ii} - ((\hat D^x_{t})^2)_{ii}| \leq |(( D^x_{t+1})^2)_{ii} - ((D^x_{t})^2)_{ii}|$;

$\bullet$ if $((D^x_{t+1})^2)_{ii} < e^2$ and $((D^x_{t})^2)_{ii} < e^2$ then $|((
\hat D^x_{t+1})^2)_{ii} - ((\hat D^x_{t})^2)_{ii}| = 0 \leq |(( D^x_{t+1})^2)_{ii} - ((D^x_{t})^2)_{ii}|$.

Hence,
\begin{align*}
  e I &\preccurlyeq \hat D^x_{t+1} \preccurlyeq \hat D^x_{t} + \frac{1}{2e} \| (D^x_{t+1})^2 -  (D^x_{t})^2 \|_{\infty} I \\
  &= \hat D^x_{t} + \frac{1-\beta_{t+1}}{2e} \| (H^x_t)^2 -  (D^x_{t})^2 \|_{\infty} I
\end{align*}
All diagonal elements of $(H^x_t)^2$ and $(D^x_{t})^2$ are from $0$ to $\Gamma^2$, then
\begin{align*}
  e I &\preccurlyeq \hat D^x_{t+1} \preccurlyeq \hat D^x_{t} + \frac{(1-\beta_{t+1}) \Gamma^2 }{2e} I.
\end{align*}
Finally, using that $ I\preccurlyeq \frac{1}{e}  $ and get
\begin{align*}
  e I &\preccurlyeq \hat D^x_{t+1} \preccurlyeq \left(1 + \frac{(1-\beta_{t+1}) \Gamma^2 }{2e^2}\right) \hat{D}^x_t .
\end{align*}
Similar proof for the variable $y$. $\square$

3) 
We start from simple steps:
\begin{align*}
  e I \preccurlyeq \hat D^x_{t+1}  = \hat D^x_{t} + \hat D^x_{t+1} -  \hat D^x_{t}.
\end{align*}
Then,
\begin{align*}
  e I &\preccurlyeq \hat D^x_{t+1} \preccurlyeq \hat D^x_{t} + \| \hat D^x_{t+1} -  \hat D^x_{t} \|_{\infty} I.
\end{align*}
One can note the following:

$\bullet$ if $(D^x_{t+1})_{ii} \geq e$ and $(D^x_{t})_{ii} \geq e$ then $|(\hat D^x_{t+1})_{ii} - (\hat D^x_{t})_{ii}| = |(D^x_{t+1})_{ii} - (D^x_{t})_{ii}|$;

$\bullet$ if $(D^x_{t+1})_{ii} \geq e$ and $(D^x_{t})_{ii} < e$ then $|(\hat D^x_{t+1})_{ii} - (\hat D^x_{t})_{ii}| \leq |(D^x_{t+1})_{ii} - (D^x_{t})_{ii}|$;

$\bullet$ if $(D^x_{t+1})_{ii} < e$ and $(D^x_{t})_{ii} \geq e$ then $|(\hat D^x_{t+1})_{ii} - (\hat D^x_{t})_{ii}| \leq |(D^x_{t+1})_{ii} - (D^x_{t})_{ii}|$;

$\bullet$ if $(D^x_{t+1})_{ii} < e$ and $(D^x_{t})_{ii} < e$ then $|(
\hat D^x_{t+1})_{ii} - (\hat D^x_{t})_{ii}| = 0 \leq |(D^x_{t+1})_{ii} - (D^x_{t})_{ii}|$.

Hence,
\begin{align*}
  e I &\preccurlyeq \hat D^x_{t+1} \preccurlyeq \hat D^x_{t} + \| D^x_{t+1} -  D^x_{t} \|_{\infty} I \\
  &= \hat D^x_{t} + (1-\beta_{t+1}) \| H^x_t -  D^x_{t} \|_{\infty} I
\end{align*}
All diagonal elements of $H^x_t$ and $D^x_{t}$ are from $-\Gamma$ to $\Gamma$, then
\begin{align*}
  e I &\preccurlyeq \hat D^x_{t+1} \preccurlyeq \hat D^x_{t} + 2(1-\beta_{t+1}) \Gamma I.
\end{align*}
Finally, using that $ I\preccurlyeq \frac{1}{e}  $ and get
\begin{align*}
  e I &\preccurlyeq \hat D^x_{t+1} \preccurlyeq \left(1 + \frac{2(1-\beta_{t+1}) \Gamma }{e}\right) \hat{D}^x_t .
\end{align*}
Similar proof for the variable $y$. $\square$

\subsection{Proof of Lemma \ref{lem:pract_prec}}

1) \textbf{OASIS:} $H^x = \text{diag}(q^x)$, where $q^x = (v^x \odot \nabla^2_{xx} f(x,y, \xi)  v^x)$ with Rademacher vector $v$. $H^x$ is diagonal from the definition. Moreover,
\begin{align*}
  \| H^x \|^2_{\infty} &= \| q^x \|^2_{\infty} \leq \left(\max_i \left[ \sum\limits_{j=1}^{d_x} |(\nabla^2_{xx} f(x,y, \xi))_{ij}|\right]\right)^2 \\
  &\leq \max_i \left( \sum\limits_{j=1}^{d_x}|(\nabla^2_{xx} f(x,y, \xi))_{ij}|\right)^2 \\
  &\leq \max_i \left[{d_x} \sum\limits_{j=1}^{d_x} (\nabla^2_{xx} f(x,y, \xi))_{ij}^2\right] \\
  &\leq {d_x} \sum\limits_{i=1}^{d_x} \sum\limits_{j=1}^{d_x} (\nabla^2_{xx} f(x,y, \xi))_{ij}^2 \\
  & = {d_x} \| \nabla^2_{xx} f(x,y, \xi) \|^2_2 \leq {d_x} L^2.
\end{align*}
Finally, we have $\| H^x \|_{\infty} \leq \sqrt{{d_x}} L = \Gamma$. Similar proof for the variable $y$.

2) \textbf{AdaHessian:} $(H^x)^2 = \text{diag}(v^x \odot v^x)$. The proof can be given in a similar way to the previous point.

3)  \textbf{Adam:} $(H^x)^2 = \text{diag}(\nabla_x f(x, y, \xi) \odot \nabla_x f(x, y, \xi))$. $H^x$ is diagonal from the definition. Moreover, using diagonal structure of $H^x$ and $(H^x)^2$,
\begin{align*}
  \| H^x \|^2_{\infty} &= \| (H^x)^2 \|_{\infty} = \| \text{diag}(\nabla_x f(x, y, \xi) \odot \nabla_x f(x, y, \xi)) \|_{\infty} \\
  &=  \| \nabla_x f(x, y, \xi) \odot \nabla_x f(x, y, \xi) \|_{\infty} \\
  &=  \| \nabla_x f(x, y, \xi) \|^2_{\infty} \\
  &\leq  \| \nabla_x f(x, y, \xi) \|^2 \leq M^2.
\end{align*}
Finally, we have $\| H^x \|_{\infty} \leq M = \Gamma$. Similar proof for the variable $y$.

4) \textbf{RMSProp:} $(H^x)^2 = \text{diag}(\nabla_x f(x, y, \xi) \odot \nabla_x f(x, y, \xi))$. The proof can be given in a similar way to the previous point. $\square$

\subsection{Proof of Theorem \ref{th:main0}}

We start with additional notation:
\begin{align*}
    F(z_t) = F(x_t, y_t) = \binom{\nabla_x f(x_t,y_t)}{-\nabla_y f(x_t,y_t)}, \quad g_t = \binom{\nabla_x f(x_t,y_t, \xi_t)}{-\nabla_y f(x_t,y_t, \xi_t)},
\end{align*}
\begin{align*}
\hat V_t =
    \begin{pmatrix}
    \hat D^x_t & 0 \\
    0 & \hat D^y_t
    \end{pmatrix}.
\end{align*}
For the matrices $\{\hat V_t\}$ we can prove an analogue of Lemma \ref{lem:precond}. With new notation Algorithm \ref{Scaled_ExtraGrad} can be rewritten as follows:
\begin{align*}
    z_{t+1/2} &= z_t - \gamma_t (\hat V_t)^{-1} g_t \\
    z_{t+1} &= z_{t} - \gamma_t (\hat V_t)^{-1} g_{t+1/2}.
\end{align*}
This is the notation we will use in the proof. Before prove Theorem, we consider the following lemma:
\begin{lemma} 
\label{l1}
Let $z,y \in \mathbb{R}^n$ and $D \in \mathcal{S}^d_{++}$. We set $z^+ = z - y$, then for all $u \in \mathbb{R}^n$:
$$\|z^+ - u\|^2_D = \|z - u \|^2_D - 2 \langle y, z^+ - u\rangle_D - \|z^+ - z \|^2_D.$$
\end{lemma}
\textbf{Proof of Lemma:}
Let us consider the following chain:
\begin{align*}
 \|z^+ - u \|^2_D &= \|z^+ - z + z - u \|^2_D \\
&= \|z-u \|^2_D + 2 \langle z^+ - z, z - u \rangle_D + \|z^+ -z \|^2_D \\
&= \|z-u \|^2_D + 2 \langle z^+ - z, z^+ - u \rangle_D - \|z^+ - z \|^2_D \\
&= \|z-u \|^2_D + 2 \langle z^+ - (z - y), z^+ - u \rangle_D - 2 \langle y, z^+ - u \rangle_D - \|z^+ - z \|^2_D \\
&= \|z - u\|^2_D - 2 \langle y, z^+ - u \rangle_D - \|z^+ - z \|^2_D.
 \end{align*}
 $\square$

\textbf{Proof of Theorem \ref{th:main0}:} Applying the previous Lemma with $z^+ = z_{t+1}$,  $z=z_{t}$, $u = z$, $y = \gamma \hat V^{-1}_t g_{t+1/2}$ and $D = \hat V_t$, we get
\begin{align*}
 \|z_{t+1} - z \|^2_{\hat V_t} &= \| z_t - z \|^2_{\hat V_t} - 2 \gamma \langle \hat V^{-1}_t g_{t+1/2},  z_{t+1} - z \rangle_{\hat V_t} - \| z_{t+1} -  z_t \|^2_{\hat V_t} \\
 &= \| z_t - z \|^2_{\hat V_t} - 2 \gamma \langle g_{t+1/2},  z_{t+1} - z \rangle - \| z_{t+1} -  z_t \|^2_{\hat V_t},
 \end{align*}
and with $z^+ =  z_{t+1/2}$,  $z= z_{t}$, $u = z_{t+1}$, $y = \gamma V^{-1}_t g_{t}$, $D = \hat V_t$:
\begin{align*}
 \| z_{t+1/2} -  z_{t+1} \|^2_{\hat V_t} &= \| z_t - z_{t+1} \|^2_{\hat V_t} - 2 \gamma \langle \hat V^{-1}_t g_{t},  z_{t+1/2} -  z_{t+1} \rangle_{\hat V_t} - \| z_{t+1/2} -  z_t \|^2_{\hat V_t} \\
 &= \| z_t - z_{t+1} \|^2_{\hat V_t} - 2 \gamma \langle g_{t},  z_{t+1/2} -  z_{t+1} \rangle - \| z_{t+1/2} -  z_t \|^2_{\hat V_t}.
 \end{align*}
Next, we sum up the two previous equalities
 \begin{align*}
 \|z_{t+1} - z \|^2_{\hat V_t} + \| z_{t+1/2} -  z_{t+1} \|^2_{\hat V_t} =& \| z_t - z \|^2_{\hat V_t} - \| z_{t+1/2} -  z_t \|^2_{\hat V_t} \\
  &- 2 \gamma \langle g_{t+1/2},  z_{t+1} - z \rangle - 2 \gamma \langle g_{t},  z_{t+1/2} -  z_{t+1} \rangle.
 \end{align*}
A small rearrangement gives
\begin{align*}
\|z_{t+1} - z \|^2_{\hat V_t} +& \| z_{t+1/2} -  z_{t+1} \|^2_{\hat V_t} \nonumber\\ 
=& \| z_t - z \|^2_{\hat V_t} - \| z_{t+1/2} -  z_t \|^2_{\hat V_t} \nonumber\\ 
 &- 2 \gamma \langle g_{t+1/2},  z_{t+1/2} - z \rangle - 2 \gamma \langle g_{t} - g_{t+1/2},  z_{t+1/2} -  z_{t+1} \rangle \nonumber\\
 \leq& \| z_t - z \|^2_{\hat V_t} - \| z_{t+1/2} -  z_t \|^2_{\hat V_t}  - 2 \gamma \langle  g^{t+1/2},  z_{t+1/2} - z \rangle \nonumber\\ 
 &  + \gamma^2 \| g_{t+1/2} -  g_{t}\|^2_{\hat V^{-1}_t} + \|  z_{t+1/2} -  z_{t+1}\|^2_{\hat V_t}.
 \end{align*}
And then
\begin{align*}
\|z_{t+1} - z \|^2_{\hat V_t} \leq& \| z_t - z \|^2_{\hat V_t} - \| z_{t+1/2} -  z_t \|^2_{\hat V_t}  - 2 \gamma \langle  g^{t+1/2},  z_{t+1/2} - z \rangle \nonumber\\ 
 &  + \gamma^2 \| g_{t+1/2} -  g_{t}\|^2_{\hat V^{-1}_t}.
\end{align*}
Next, we use that $\hat{V}^{-1}_t \preccurlyeq  \frac{1}{e} I $ and get
\begin{align*}
\|z_{t+1} - z \|^2_{\hat V_t} \leq& \| z_t - z \|^2_{\hat V_t} - \| z_{t+1/2} -  z_t \|^2_{\hat V_t}  - 2 \gamma \langle  g^{t+1/2},  z_{t+1/2} - z \rangle + \frac{\gamma^2}{e} \| g_{t+1/2} -  g_{t}\|^2 \nonumber\\
 \leq& \| z_t - z \|^2_{\hat V_t} - \| z_{t+1/2} -  z_t \|^2_{\hat V_t}  - 2 \gamma \langle  g^{t+1/2},  z_{t+1/2} - z \rangle  \nonumber\\ 
 &+ \frac{3\gamma^2}{e} \| F (z_{t+1/2}) -  F (z_{t})\|^2 + \frac{3\gamma^2}{e} \| g_{t+1/2} -   F (z_{t+1/2})\|^2 \nonumber\\ 
 & + \frac{3\gamma^2}{e} \| F (z_{t}) -  g_{t}\|^2.
\end{align*}
Smoothness of function $f$ (Assumption \ref{as:Lipsh}) gives
\begin{align}
\label{t1}
\|z_{t+1} - z \|^2_{\hat V_t} \leq& \| z_t - z \|^2_{\hat V_t} - \| z_{t+1/2} -  z_t \|^2_{\hat V_t}  - 2 \gamma \langle  g^{t+1/2},  z_{t+1/2} - z \rangle \nonumber\\ 
 &  + \frac{3\gamma^2 L^2}{e} \| z_{t+1/2} -  z_{t}\|^2 + \frac{3\gamma^2}{e} \| g_{t+1/2} -   F (z_{t+1/2})\|^2 \nonumber\\ 
 & + \frac{3\gamma^2}{e} \| F (z_{t}) -  g_{t}\|^2 \nonumber\\ 
=& \| z_t - z \|^2_{\hat V_t} - \| z_{t+1/2} -  z_t \|^2_{\hat V_t}  - 2 \gamma \langle  F(z_{t+1/2}),  z_{t+1/2} - z \rangle \nonumber\\ 
 &  + \frac{3\gamma^2 L^2}{e} \| z_{t+1/2} -  z_{t}\|^2 + \frac{3\gamma^2}{e} \| g_{t+1/2} -   F (z_{t+1/2})\|^2 \nonumber\\ 
 & + \frac{3\gamma^2}{e} \| F (z_{t}) -  g_{t}\|^2 + 2 \gamma \langle  F(z_{t+1/2}) - g_{t+1/2},  z_{t+1/2} - z \rangle .
\end{align}
\textbf{Strongly convex--strongly concave case.} We substitute $z = z^*$  and take the total expectation of both sides of the equation
\begin{align*}
\E\left[\|z_{t+1} - z^* \|^2_{\hat V_t} \right] \leq&  \E\left[\| z_t - z^* \|^2_{\hat V_t}\right] - \E\left[\| z_{t+1/2} -  z_t \|^2_{\hat V_t}\right]  - 2 \gamma \E\left[\langle  F(z_{t+1/2}),  z_{t+1/2} - z^* \rangle\right] \nonumber\\ 
 &  + \frac{3\gamma^2 L^2}{e} \E\left[\| z_{t+1/2} -  z_{t}\|^2\right] + \frac{3\gamma^2}{e} \E\left[\| g_{t+1/2} -   F (z_{t+1/2})\|^2\right] \nonumber\\ 
 & + \frac{3\gamma^2}{e} \E\left[\| F (z_{t}) -  g_{t}\|^2\right] + 2 \gamma \E\left[\langle  F(z_{t+1/2}) - g_{t+1/2},  z_{t+1/2} - z^* \rangle \right].
\end{align*}
Next, using the property of the solution $z^*$: $\langle F(z^*),  z_{t+1/2} - z^*\rangle = 0$, and get
\begin{align*}
\E\left[\|z_{t+1} - z^* \|^2_{\hat V_t} \right] \leq&  \E\left[\| z_t - z^* \|^2_{\hat V_t}\right] - \E\left[\| z_{t+1/2} -  z_t \|^2_{\hat V_t}\right]  \nonumber\\ 
&- 2 \gamma \E\left[\langle  F(z_{t+1/2}) - F(z^*),  z_{t+1/2} - z^* \rangle\right] \nonumber\\ 
 &  + \frac{3\gamma^2 L^2}{e} \E\left[\| z_{t+1/2} -  z_{t}\|^2\right] + \frac{3\gamma^2}{e} \E\left[\| g_{t+1/2} -   F (z_{t+1/2})\|^2\right] \nonumber\\ 
 & + \frac{3\gamma^2}{e} \E\left[\| F (z_{t}) -  g_{t}\|^2\right] + 2 \gamma \E\left[\langle  F(z_{t+1/2}) - g_{t+1/2},  z_{t+1/2} - z^* \rangle \right] \nonumber\\ 
=&  \E\left[\| z_t - z^* \|^2_{\hat V_t}\right] - \E\left[\| z_{t+1/2} -  z_t \|^2_{\hat V_t}\right]  \nonumber\\ 
&- 2 \gamma \E\left[\langle  F(z_{t+1/2}) - F(z^*),  z_{t+1/2} - z^* \rangle\right] + \frac{3\gamma^2 L^2}{e} \E\left[\| z_{t+1/2} -  z_{t}\|^2\right] \nonumber\\ 
 & + \frac{3\gamma^2}{e} \E\left[\E_{t+1/2}\left[\| g_{t+1/2} -   F (z_{t+1/2})\|^2\right]\right] + \frac{3\gamma^2}{e} \E\left[\E_{t}\left[\| F (z_{t}) -  g_{t}\|^2\right]\right] \nonumber\\ 
 &  + 2 \gamma \E\left[\langle  \E_{t+1/2}\left[F(z_{t+1/2}) - g_{t+1/2}\right],  z_{t+1/2} - z^* \rangle \right] .
\end{align*}
Assumptions \ref{as:conv} and \ref{as:var} on strongly convexity - strongly concavity and on stochastisity give
\begin{align*}
\E\left[\|z_{t+1} - z^* \|^2_{\hat V_t} \right] \leq&  \E\left[\| z_t - z^* \|^2_{\hat V_t}\right] - \E\left[\| z_{t+1/2} -  z_t \|^2_{\hat V_t}\right]  \nonumber\\ 
&- 2 \gamma \mu \E\left[\| z_{t+1/2} -  z^*\|^2\right] + \frac{3\gamma^2 L^2}{e} \E\left[\| z_{t+1/2} -  z_{t}\|^2\right] \nonumber\\ 
 & + \frac{6\gamma^2 \sigma^2}{eb} \nonumber\\
 \leq&  \E\left[\| z_t - z^* \|^2_{\hat V_t}\right] - \E\left[\| z_{t+1/2} -  z_t \|^2_{\hat V_t}\right] - \gamma \mu \E\left[\| z_{t} -  z^*\|^2\right]  \nonumber\\ 
& +  \left(2 \gamma \mu + \frac{3\gamma^2 L^2}{e}\right)\E\left[\| z_{t+1/2} -  z_t\|^2\right]  + \frac{6\gamma^2 \sigma^2}{eb}.
\end{align*}
With $\frac{1}{\Gamma}\hat{V}_t \preccurlyeq  I\preccurlyeq \frac{1}{e}\hat{V}_t$, we get
\begin{align*}
\E\left[\|z_{t+1} - z^* \|^2_{\hat V_t} \right] 
 \leq&  \left( 1 - \frac{\gamma \mu}{\Gamma}\right)\E\left[\| z_t - z^* \|^2_{\hat V_t}\right] \nonumber\\ 
& -  \left(1 - \frac{2 \gamma \mu}{e} - \frac{3\gamma^2 L^2}{e^2}\right)\E\left[\| z_{t+1/2} -  z_t\|^2_{\hat V_t}\right]  + \frac{6\gamma^2 \sigma^2}{eb}.
\end{align*}
The choice of $\gamma \leq \min\left\{\frac{e}{4 \mu} ; \frac{e}{3 L} \right\}$ move us to
\begin{align*}
\E\left[\|z_{t+1} - z^* \|^2_{\hat V_t} \right] 
 \leq&  \left( 1 - \frac{\gamma \mu}{\Gamma}\right)\E\left[\| z_t - z^* \|^2_{\hat V_t}\right] + \frac{6\gamma^2 \sigma^2}{eb}.
\end{align*}
Lemma \ref{lem:precond} with $C = \frac{\Gamma^2}{2e^2}$ for \eqref{eq:precond} and $C = \frac{2\Gamma}{e}$ for \eqref{eq:precond_add} gives
\begin{align*}
\E\left[\|z_{t+1} - z^* \|^2_{\hat V_{t+1}} \right] 
 \leq&  \left( 1 + (1-\beta_{t+1})C \right) \E\left[\|z_{t+1} - z^* \|^2_{\hat V_{t}} \right] .
\end{align*}
Then we have
\begin{align*}
\E\left[\|z_{t+1} - z^* \|^2_{\hat V_{t+1}} \right] 
 \leq&  \left( 1 - \frac{\gamma \mu}{\Gamma}\right)\left( 1 + (1-\beta_{t+1})C \right)\E\left[\| z_t - z^* \|^2_{\hat V_t}\right] \nonumber\\ 
 &+ \frac{6\gamma^2 \sigma^2}{eb} \left( 1 + (1-\beta_{t+1})C \right).
\end{align*}
This completes the proof of the strongly convex-strongly concave case. $\square$

\textbf{Convex--concave case.} We start from \eqref{t1} with small rearrangement
\begin{align*}
2 \gamma \langle  F(z_{t+1/2}),  z_{t+1/2} - z \rangle \leq& \| z_t - z \|^2_{\hat V_t} - \|z_{t+1} - z \|^2_{\hat V_t} - \| z_{t+1/2} -  z_t \|^2_{\hat V_t}  \nonumber\\ 
 &  + \frac{3\gamma^2 L^2}{e} \| z_{t+1/2} -  z_{t}\|^2 + \frac{3\gamma^2}{e} \| g_{t+1/2} -   F (z_{t+1/2})\|^2 \nonumber\\ 
 & + \frac{3\gamma^2}{e} \| F (z_{t}) -  g_{t}\|^2 + 2 \gamma \langle  F(z_{t+1/2}) - g_{t+1/2},  z_{t+1/2} - z \rangle .
\end{align*}
With $ I \preccurlyeq  \frac{1}{e}\hat{V}_t$ and Lemma \ref{lem:precond} with $C = \frac{\Gamma^2}{2e^2}$ for \eqref{eq:precond} and $C = \frac{2\Gamma}{e}$ for \eqref{eq:precond_add}, we get
\begin{align*}
2 \gamma \langle  F(z_{t+1/2}),  z_{t+1/2} - z \rangle \leq& \| z_t - z \|^2_{\hat V_t} - 
\left( 1 + (1-\beta_{t+1})C \right)^{-1}\|z_{t+1} - z \|^2_{\hat V_{t+1}} \nonumber\\ 
 &  -\left(1 -  \frac{3\gamma^2 L^2}{e^2}\right) \| z_{t+1/2} -  z_{t}\|^2_{\hat V_t} + \frac{3\gamma^2}{e} \| g_{t+1/2} -   F (z_{t+1/2})\|^2 \nonumber\\ 
 & + \frac{3\gamma^2}{e} \| F (z_{t}) -  g_{t}\|^2 + 2 \gamma \langle  F(z_{t+1/2}) - g_{t+1/2},  z_{t+1/2} - z \rangle .
\end{align*}
 Next, we sum over all $t$ from $0$ to $T-1$ and take maximum on $\mathcal{Z}$
\begin{align}
\label{temp7}
2 \gamma \max\limits_{z \in \mathcal{Z}} \frac{1}{T}&\sum\limits_{t=0}^{T-1}  \langle F(z_{t+1/2}),  z_{t+1/2} - z \rangle \nonumber\\ 
\leq& \max\limits_{z \in \mathcal{Z}} \frac{1}{T}\sum\limits_{t=0}^{T-1} \left[\| z_t - z \|^2_{\hat V_t} - 
\left( 1 + (1-\beta_{t+1})C \right)^{-1}\|z_{t+1} - z \|^2_{\hat V_{t+1}} \right]\nonumber\\ 
 &  -\left(1 -  \frac{3\gamma^2 L^2}{e^2}\right) \frac{1}{T}\sum\limits_{t=0}^{T-1} \left[\| z_{t+1/2} -  z_{t}\|^2_{\hat V_t} \right]+ \frac{3\gamma^2}{e} \frac{1}{T}\sum\limits_{t=0}^{T-1} \left[\| g_{t+1/2} -   F (z_{t+1/2})\|^2 \right]\nonumber\\ 
 & + \frac{3\gamma^2}{e} \frac{1}{T}\sum\limits_{t=0}^{T-1} \left[\| F (z_{t}) -  g_{t}\|^2 \right]+  \max\limits_{z \in \mathcal{Z}} \frac{1}{T}\sum\limits_{t=0}^{T-1} \left[ 2 \gamma \langle  F(z_{t+1/2}) - g_{t+1/2},  z_{t+1/2} - z \rangle \right] .
 \end{align}
 Then, by $x_{T}^{avg} = \frac{1}{T}\sum_{t=0}^{T-1} x_{t+1/2}$ and $y_{T}^{avg} = \frac{1}{T}\sum_{t=0}^{T-1} y_{t+1/2}$, Jensen's inequality and convexity-concavity of $f$:
\begin{align*}
    \text{gap}(z_{T}^{avg})
    &\leq \max\limits_{y' \in \mathcal{Y}} f\left(\frac{1}{T} \left(\sum^{T-1}_{t=0} x_{t+1/2} \right), y'\right) - \min\limits_{x' \in \mathcal{X}} f\left(x', \frac{1}{T} \left(\sum^{T-1}_{t=0} y_{t+1/2} \right)\right) 
    \nonumber \\
    &\leq \max\limits_{y' \in \mathcal{Y}} \frac{1}{T} \sum^{T-1}_{t=0} f(x_{t+1/2}, y')  - \min\limits_{x' \in \mathcal{X}} \frac{1}{T} \sum^{T-1}_{t=0} f(x', y_{t+1/2}).
\end{align*}
Given the fact of linear independence of $x'$ and $y'$:
\begin{align*}
    \text{gap}(z_{T}^{avg}) &\leq \max\limits_{(x', y') \in \mathcal{Z}}\frac{1}{T} \sum^{T-1}_{t=0}  \left(f(x_{t+1/2}, y')  - f(x', y_{t+1/2}) \right).
\end{align*}
Using convexity and concavity of the function $f$:
\begin{align}
\label{temp8}
    &\text{gap}(z_{T}^{avg}) \leq  \max\limits_{(x', y') \in \mathcal{Z}}\frac{1}{T} \sum^{T-1}_{t=0}  \left(f(x_{t+1/2}, y')  - f(x', y_{t+1/2}) \right)  \nonumber \\
    &=  \max\limits_{(x', y') \in \mathcal{Z}} \frac{1}{T} \sum^{T-1}_{t=0} \Big(f(x_{t+1/2}, y') -f(x_{t+1/2}, y_{t+1/2})  + f(x_{t+1/2}, y_{t+1/2}) - f(x', y_{t+1/2}) \Big) \nonumber \\
    &\leq \max\limits_{(x', y') \in \mathcal{Z}} \frac{1}{T} \sum^{T-1}_{t=0} \Big(\langle \nabla_y f (x_{t+1/2}, y_{t+1/2}), y'-y_{t+1/2} \rangle + \langle \nabla_x f (x_{t+1/2}, y_{t+1/2}), x_{t+1/2}-x' \rangle \Big) \nonumber \\
    &\leq \max\limits_{z \in \mathcal{Z}} \frac{1}{T} \sum^{T-1}_{t=0} \langle  F(z_{t+1/2}), z_{t+1/2} - z\rangle.
\end{align}
Combining \eqref{temp7} and \eqref{temp8}, we get
\begin{align*}
2\gamma \text{gap}(z_{T}^{avg})
\leq& \max\limits_{z \in \mathcal{Z}} \frac{1}{T}\sum\limits_{t=0}^{T-1} \left[\| z_t - z \|^2_{\hat V_t} - 
\left( 1 + (1-\beta_{t+1}) C \right)^{-1}\|z_{t+1} - z \|^2_{\hat V_{t+1}} \right]\nonumber\\ 
 &  -\left(1 -  \frac{3\gamma^2 L^2}{e^2}\right) \frac{1}{T}\sum\limits_{t=0}^{T-1} \left[\| z_{t+1/2} -  z_{t}\|^2_{\hat V_t} \right] + \frac{3\gamma^2}{e} \frac{1}{T}\sum\limits_{t=0}^{T-1} \left[\| g_{t+1/2} -   F (z_{t+1/2})\|^2\right] \nonumber\\ 
 & + \frac{3\gamma^2}{e} \frac{1}{T}\sum\limits_{t=0}^{T-1} \left[\| F (z_{t}) -  g_{t}\|^2 \right] +  \max\limits_{z \in \mathcal{Z}} \frac{1}{T}\sum\limits_{t=0}^{T-1} \left[ 2 \gamma \langle  F(z_{t+1/2}) - g_{t+1/2},  z_{t+1/2} - z \rangle \right] .
 \end{align*}
The choice of $\gamma \leq \frac{e}{2 L}$ gives
\begin{align*}
2\gamma \cdot \text{gap}(z_{T}^{avg})
\leq& \max\limits_{z \in \mathcal{Z}} \frac{1}{T}\sum\limits_{t=0}^{T-1} \left[\| z_t - z \|^2_{\hat V_t} - 
\left( 1 + (1-\beta_{t+1}) C \right)^{-1}\|z_{t+1} - z \|^2_{\hat V_{t+1}} \right]\nonumber\\ 
 & + \frac{3\gamma^2}{e} \frac{1}{T}\sum\limits_{t=0}^{T-1} \left[\| g_{t+1/2} -   F (z_{t+1/2})\|^2\right] + \frac{3\gamma^2}{e} \frac{1}{T}\sum\limits_{t=0}^{T-1} \left[\| F (z_{t}) -  g_{t}\|^2 \right]\nonumber\\
 &  +  \max\limits_{z \in \mathcal{Z}} \frac{1}{T}\sum\limits_{t=0}^{T-1} \left[ 2 \gamma \langle  F(z_{t+1/2}) - g_{t+1/2},  z_{t+1/2} - z \rangle \right] \nonumber\\
 \leq& \max\limits_{z \in \mathcal{Z}} \left[\frac{\| z_0 - z \|^2_{\hat V_0}}{T}+ 
\left( 1 + (1-\beta_{t+1}) C \right)^{-1} (1-\beta_{t+1})C \cdot \frac{1}{T}\sum\limits_{t=1}^{T-1} \|z_{t} - z \|^2_{\hat V_{t}} \right]\nonumber\\ 
 & + \frac{3\gamma^2}{e} \frac{1}{T}\sum\limits_{t=0}^{T-1} \left[\| g_{t+1/2} -   F (z_{t+1/2})\|^2 \right]+ \frac{3\gamma^2}{e} \frac{1}{T}\sum\limits_{t=0}^{T-1} \left[\| F (z_{t}) -  g_{t}\|^2 \right] \nonumber\\ 
 &  +  \frac{1}{T}\sum\limits_{t=0}^{T-1} \left[ 2 \gamma \langle  F(z_{t+1/2}) - g_{t+1/2},  z_{t+1/2} \rangle \right]
 \nonumber\\ 
 &  +  \max\limits_{z \in \mathcal{Z}} \left[ 2 \gamma \langle  \frac{1}{T}\sum\limits_{t=0}^{T-1} [F(z_{t+1/2}) - g_{t+1/2}],  - z \rangle \right].
 \end{align*}
One can note that $\left( 1 + (1-\beta_{t+1}) C \right) \geq 1$, and get 
\begin{align*}
2\gamma \cdot \text{gap}(z_{T}^{avg})
 \leq& \max\limits_{z \in \mathcal{Z}} \left[\frac{\| z_0 - z \|^2_{\hat V_0}}{T}+ 
\frac{1}{T}\sum\limits_{t=1}^{T-1} (1-\beta_{t+1}) C \|z_{t} - z \|^2_{\hat V_{t}} \right]\nonumber\\ 
 & + \frac{3\gamma^2}{e} \frac{1}{T}\sum\limits_{t=0}^{T-1} \left[\| g_{t+1/2} -   F (z_{t+1/2})\|^2 \right] + \frac{3\gamma^2}{e} \frac{1}{T}\sum\limits_{t=0}^{T-1} \left[\| F (z_{t}) -  g_{t}\|^2\right] \nonumber\\
 &  +  \frac{1}{T}\sum\limits_{t=0}^{T-1} \left[ 2 \gamma \langle  F(z_{t+1/2}) - g_{t+1/2},  z_{t+1/2} \rangle \right]
 \nonumber\\ 
 &  +  \max\limits_{z \in \mathcal{Z}} \left[ 2 \gamma \langle  \frac{1}{T}\sum\limits_{t=0}^{T-1} [F(z_{t+1/2}) - g_{t+1/2}],  - z \rangle \right].
 \end{align*}
Next, we use $\hat{V}_t \preccurlyeq \Gamma I$ and the fact that iterations and the set $\mathcal{Z}$ are bounded: $\|z_{t}\|, \|z\| \leq \Omega$
\begin{align*}
2\gamma \cdot \text{gap}(z_{T}^{avg})
  \leq& \max\limits_{z \in \mathcal{Z}} \left[\frac{\Gamma\| z_0 - z \|^2}{T}+ 
\frac{1}{T}\sum\limits_{t=1}^{T-1} (1-\beta_{t+1}) C \Gamma \|z_{t} - z \|^2 \right]\nonumber\\ 
 & + \frac{3\gamma^2}{e} \frac{1}{T}\sum\limits_{t=0}^{T-1} \left[\| g_{t+1/2} -   F (z_{t+1/2})\|^2 \right] + \frac{3\gamma^2}{e} \frac{1}{T}\sum\limits_{t=0}^{T-1} \left[\| F (z_{t}) -  g_{t}\|^2 \right] \nonumber\\ 
 &  +  \frac{1}{T}\sum\limits_{t=0}^{T-1} \left[ 2 \gamma \langle  F(z_{t+1/2}) - g_{t+1/2},  z_{t+1/2} \rangle \right]
 \nonumber\\ 
 &  +  \max\limits_{z \in \mathcal{Z}} \left[ 2 \gamma \langle  \frac{1}{T}\sum\limits_{t=0}^{T-1} [F(z_{t+1/2}) - g_{t+1/2}],  - z \rangle \right]  \nonumber\\ 
 \leq& \frac{4 \Gamma \Omega^2}{T} + 
\frac{1}{T}\sum\limits_{t=1}^{T-1} 4(1-\beta_{t+1}) C \Gamma \Omega^2\nonumber\\ 
 & + \frac{3\gamma^2}{e} \frac{1}{T}\sum\limits_{t=0}^{T-1} \left[\| g_{t+1/2} -   F (z_{t+1/2})\|^2\right] + \frac{3\gamma^2}{e} \frac{1}{T}\sum\limits_{t=0}^{T-1} \left[\| F (z_{t}) -  g_{t}\|^2\right] \nonumber\\
 &  +  \frac{1}{T}\sum\limits_{t=0}^{T-1} \left[ 2 \gamma \langle  F(z_{t+1/2}) - g_{t+1/2},  z_{t+1/2} \rangle \right]
 \nonumber\\ 
 &  +  \max\limits_{z \in \mathcal{Z}} \left[ 2 \gamma \langle  \frac{1}{T}\sum\limits_{t=0}^{T-1} [F(z_{t+1/2}) - g_{t+1/2}],  - z \rangle \right].
 \end{align*}
Next, we take a full expectation and use Assumption \ref{as:var}
 \begin{align}
 \label{eq:temp404}
2\gamma \cdot \E\left[\text{gap}(z_{T}^{avg})\right]
\leq& \frac{4 \Gamma \Omega^2}{T} + \frac{1}{T}\sum\limits_{t=1}^{T-1}
4(1-\beta_{t+1}) C \Gamma \Omega^2\nonumber\\ 
 & + \frac{3\gamma^2}{e} \frac{1}{T}\sum\limits_{t=0}^{T-1} \left[\E\left[\| g_{t+1/2} -   F (z_{t+1/2})\|^2\right]\right] \nonumber\\ 
 & + \frac{3\gamma^2}{e} \frac{1}{T}\sum\limits_{t=0}^{T-1} \left[\E\left[\| F (z_{t}) -  g_{t}\|^2 \right] \right]\nonumber\\ 
 &  +  \frac{1}{T}\sum\limits_{t=0}^{T-1} \E\left[ 2 \gamma \langle  F(z_{t+1/2}) - g_{t+1/2},  z_{t+1/2} \rangle \right]
 \nonumber\\ 
 &  +  \E\left[\max\limits_{z \in \mathcal{Z}} \left[ 2 \gamma \langle  \frac{1}{T}\sum\limits_{t=0}^{T-1} [F(z_{t+1/2}) - g_{t+1/2}],  - z \rangle \right]\right] \nonumber\\ 
 \leq& \frac{4\Gamma\Omega^2}{T} + 
 \frac{1}{T}\sum\limits_{t=1}^{T-1} 4(1-\beta_{t+1}) C \Gamma \Omega^2 + \frac{6\gamma^2 \sigma^2}{eb}\nonumber\\ 
 &  +  \E\left[\max\limits_{z \in \mathcal{Z}} \left[ 2 \gamma \langle  \frac{1}{T}\sum\limits_{t=0}^{T-1} [F(z_{t+1/2}) - g_{t+1/2}],  - z \rangle \right]\right]\nonumber\\ 
 \leq& \frac{4\Gamma\Omega^2}{T} + 
\frac{1}{T}\sum\limits_{t=1}^{T-1}  4(1-\beta_{t+1}) C \Gamma \Omega^2+ \frac{6\gamma^2 \sigma^2}{eb}\nonumber\\ 
 &  +  \E\left[\max\limits_{z \in \mathcal{Z}} \left[  \frac{1}{T}\left\| \gamma\sum\limits_{t=0}^{T-1} F(z_{t+1/2}) - g_{t+1/2} \right\|^2_{\hat V^{-1}_{t}} + \frac{\| z \|^2_{\hat V_{t}}}{T} \right]\right]
 \nonumber\\ 
 \leq& \frac{4\Gamma\Omega^2}{T} + 
\frac{1}{T}\sum\limits_{t=1}^{T-1} 4(1-\beta_{t+1}) C \Gamma \Omega^2+ \frac{6\gamma^2 \sigma^2}{eb}\nonumber\\ 
 &  +  \E\left[\max\limits_{z \in \mathcal{Z}} \left[  \frac{1}{e T}\left\| \gamma\sum\limits_{t=0}^{T-1} F(z_{t+1/2}) - g_{t+1/2} \right\|^2 + \frac{\Gamma \| z \|^2}{T} \right]\right].
 \end{align}
With unbiasedness and independence of all $F(z_{t+1/2}) - g_{t+1/2}$, we get
 \begin{align*} 
 2\gamma \cdot \E\left[\text{gap}(z_{T}^{avg})\right] \leq& \frac{4\Gamma\Omega^2}{T} + 
\frac{1}{T}\sum\limits_{t=1}^{T-1} 4(1-\beta_{t+1}) C \Gamma \Omega^2 + \frac{6\gamma^2 \sigma^2}{eb}\nonumber\\ 
 &  +  \max\limits_{z \in \mathcal{Z}} \frac{\Gamma \| z \|^2}{T} + 
 \frac{\gamma^2}{e T}\sum\limits_{t=0}^{T-1} \E\left[  \left\|  F(z_{t+1/2}) - g_{t+1/2} \right\|^2 \right] \\
 \leq& \frac{5\Gamma\Omega^2}{T} + 
\frac{1}{T}\sum\limits_{t=1}^{T-1}  4(1-\beta_{t+1}) C \Gamma \Omega^2 + \frac{7\gamma^2 \sigma^2}{eb}.
 \end{align*}
Here we additionally use Assumption \ref{as:var}. Finally,
 \begin{align*} 
 \E\left[\text{gap}(z_{T}^{avg})\right] \leq& \frac{3\Gamma\Omega^2}{\gamma T} + 
\frac{2 C\Gamma \Omega^2}{\gamma T} \sum\limits_{t=2}^{T}  (1-\beta_{t}) + \frac{4\gamma \sigma^2}{eb}.
 \end{align*}
This completes the proof of the convex-concave case. $\square$

\textbf{Non-convex--non-concave case.} We start from \eqref{t1}, substitute $z = z^*$  and take the total expectation of both sides of the equation
\begin{align*}
\E\left[\|z_{t+1} - z^* \|^2_{\hat V_t} \right] \leq&  \E\left[\| z_t - z^* \|^2_{\hat V_t}\right] - \E\left[\| z_{t+1/2} -  z_t \|^2_{\hat V_t}\right]  - 2 \gamma \E\left[\langle  F(z_{t+1/2}),  z_{t+1/2} - z^* \rangle\right] \nonumber\\ 
 &  + \frac{3\gamma^2 L^2}{e} \E\left[\| z_{t+1/2} -  z_{t}\|^2\right] + \frac{3\gamma^2}{e} \E\left[\| g_{t+1/2} -   F (z_{t+1/2})\|^2\right] \nonumber\\ 
 & + \frac{3\gamma^2}{e} \E\left[\| F (z_{t}) -  g_{t}\|^2\right] + 2 \gamma \E\left[\langle  F(z_{t+1/2}) - g_{t+1/2},  z_{t+1/2} - z^* \rangle \right] \nonumber\\ 
 =&  \E\left[\| z_t - z^* \|^2_{\hat V_t}\right] - \E\left[\| z_{t+1/2} -  z_t \|^2_{\hat V_t}\right]  \nonumber\\ 
&- 2 \gamma \E\left[\langle  F(z_{t+1/2}),  z_{t+1/2} - z^* \rangle\right] + \frac{3\gamma^2 L^2}{e} \E\left[\| z_{t+1/2} -  z_{t}\|^2\right] \nonumber\\ 
 & + \frac{3\gamma^2}{e} \E\left[\E_{t+1/2}\left[\| g_{t+1/2} -   F (z_{t+1/2})\|^2\right]\right] + \frac{3\gamma^2}{e} \E\left[\E_{t}\left[\| F (z_{t}) -  g_{t}\|^2\right]\right] \nonumber\\ 
 &  + 2 \gamma \E\left[\langle  \E_{t+1/2}\left[F(z_{t+1/2}) - g_{t+1/2}\right],  z_{t+1/2} - z^* \rangle \right] .
\end{align*}
Assumptions \ref{as:conv} and \ref{as:var} on non-convexity--non-concavity and on stochastisity give
\begin{align*}
\E\left[\|z_{t+1} - z^* \|^2_{\hat V_t} \right] \leq&  \E\left[\| z_t - z^* \|^2_{\hat V_t}\right] - \E\left[\| z_{t+1/2} -  z_t \|^2_{\hat V_t}\right]  \nonumber\\ 
&+ \frac{3\gamma^2 L^2}{e} \E\left[\| z_{t+1/2} -  z_{t}\|^2\right] + \frac{6\gamma^2 \sigma^2}{eb}.
\end{align*}
With $e I \preccurlyeq \hat{V}_t$, we get
\begin{align*}
\E\left[\|z_{t+1} - z^* \|^2_{\hat V_t} \right] \leq&  \E\left[\| z_t - z^* \|^2_{\hat V_t}\right] - \left( 1 - \frac{3\gamma^2 L^2}{e^2} \right)\E\left[\| z_{t+1/2} -  z_t \|^2_{\hat V_t}\right] + \frac{6\gamma^2 \sigma^2}{eb} \nonumber\\ 
=&  \E\left[\| z_t - z^* \|^2_{\hat V_t}\right] - \left( 1 - \frac{3\gamma^2 L^2}{e^2} \right) \gamma^2 \E\left[\| g_t \|^2_{\hat V^{-1}_t}\right] + \frac{6\gamma^2 \sigma^2}{eb}.
\end{align*}
With $\gamma \leq \frac{e}{3 L}$ we get
\begin{align*}
\E\left[\|z_{t+1} - z^* \|^2_{\hat V_{t}} \right] \leq&  \E\left[\| z_t - z^* \|^2_{\hat V_t}\right] - \frac{2 \gamma^2}{3} \E\left[\| g_t \|^2_{\hat V^{-1}_t}\right] + \frac{6\gamma^2 \sigma^2}{eb}.
\end{align*}
Lemma \ref{lem:precond} with $C = \frac{\Gamma^2}{2e^2}$ for \eqref{eq:precond} and $C = \frac{2\Gamma}{e}$ for \eqref{eq:precond_add} gives
\begin{align*}
\left( 1 + (1-\beta_{t+1}) C \right)^{-1} \E\left[\|z_{t+1} - z^* \|^2_{\hat V_{t+1}} \right] \leq&   \E\left[\| z_t - z^* \|^2_{\hat V_t}\right] - \frac{2 \gamma^2}{3} \E\left[\| g_t \|^2_{\hat V^{-1}_t}\right] + \frac{6\gamma^2 \sigma^2}{eb}  \nonumber\\ 
\leq&   \E\left[\| z_t - z^* \|^2_{\hat V_t}\right] - \frac{2 \gamma^2}{3} \E\left[\| g_t \|^2_{\hat V^{-1}_t}\right] + \frac{6\gamma^2 \sigma^2}{eb} 
\nonumber\\
\leq&   \E\left[\| z_t - z^* \|^2_{\hat V_t}\right] - \frac{2 \gamma^2}{3 \Gamma} \E\left[\| g_t \|^2\right] + \frac{6\gamma^2 \sigma^2}{eb}.
\end{align*}
In the last step we use that $\frac{1}{\Gamma} I \preccurlyeq \hat{V}^{-1}_t$. Then, using Assumption \ref{as:var}, we have
\begin{align*}
\left( 1 + (1-\beta_{t+1}) C \right)^{-1} \E\left[\|z_{t+1} - z^* \|^2_{\hat V_{t+1}} \right] \leq&   \E\left[\| z_t - z^* \|^2_{\hat V_t}\right] - \frac{\gamma^2}{3 \Gamma} \E\left[\| F(z_t) \|^2\right]\nonumber\\ 
& + \frac{2\gamma^2}{3 \Gamma} \E\left[\| g_t - F(z_t) \|^2\right] + \frac{6\gamma^2 \sigma^2}{eb}  \nonumber\\
\leq&   \E\left[\| z_t - z^* \|^2_{\hat V_t}\right] - \frac{\gamma^2}{3 \Gamma} \E\left[\| F(z_t) \|^2\right]\nonumber\\ 
& + \frac{7\gamma^2 \sigma^2}{eb}.
\end{align*}
Small rearrangement gives
\begin{align*}
 \frac{\gamma^2}{3 \Gamma} \E\left[\| F(z_t) \|^2\right] \leq&  \E\left[\| z_t - z^* \|^2_{\hat V_t}\right]-  \left( 1 + (1-\beta_{t+1})C \right)^{-1}\E\left[\|z_{t+1} - z^* \|^2_{\hat V_{t+1}} \right]+ \frac{7\gamma^2 \sigma^2}{eb}.
\end{align*}
Summing over all $t$ from $0$ to $T-1$ and averaging, we get
\begin{align*}
 \frac{\gamma^2}{3 \Gamma} \cdot \E\left[ \frac{1}{T} \sum\limits_{t=0}^{T-1}\| F(z_t) \|^2\right] \leq&   \frac{\E\left[\| z_0 - z^* \|^2_{\hat V_0}\right]}{T} + \frac{7\gamma^2 \sigma^2}{eb}\nonumber\\ 
& +  \frac{1}{T} \sum\limits_{t=1}^{T-1} \left( 1 + (1-\beta_{t+1}) C \right)^{-1}(1-\beta_{t+1}) C  \E\left[\|z_{t} - z^* \|^2_{\hat V_{t}} \right]  \nonumber\\
\leq&   \frac{\Gamma \| z_0 - z^* \|^2}{T} + \frac{7\gamma^2 \sigma^2}{eb} \nonumber\\ 
& +  \frac{1}{T} \sum\limits_{t=1}^{T-1} \left( 1 + (1-\beta_{t+1})C \right)^{-1} (1-\beta_{t+1}) C \Gamma \E\left[\|z_{t} - z^* \|^2 \right] .
\end{align*}
Here we use $\hat{V}_t \preccurlyeq \Gamma I$. With the fact that iterations are bounded: $\|z_{t}\| \leq \Omega$, we obtain
\begin{align*}
 \frac{\gamma^2}{3 \Gamma} \cdot \E\left[ \frac{1}{T} \sum\limits_{t=0}^{T-1}\| F(z_t) \|^2\right]
\leq&  \frac{4\Gamma \Omega^2}{T} + \frac{7\gamma^2 \sigma^2}{eb} \nonumber\\ 
& +  \frac{1}{T} \sum\limits_{t=1}^{T-1} \left( 1 + (1-\beta_{t+1}) C \right)^{-1}\cdot4(1-\beta_{t+1}) C\Gamma \Omega^2  .
\end{align*}
One can note that $\left( 1 + (1-\beta_{t+1}) C \right) \geq 1$, and get 
\begin{align*}
 \frac{\gamma^2}{3 \Gamma} \cdot \E\left[ \frac{1}{T} \sum\limits_{t=0}^{T-1}\| F(z_t) \|^2\right]
\leq&  \frac{4\Gamma \Omega^2}{T} + \frac{7\gamma^2 \sigma^2}{eb}  +  4C\Gamma \Omega^2 \cdot \frac{1}{T} \sum\limits_{t=1}^{T-1} (1-\beta_{t+1}).
\end{align*}
Finally,
\begin{align*}
\E\left[ \frac{1}{T} \sum\limits_{t=0}^{T-1}\| F(z_t) \|^2\right]
\leq&  \frac{12\Gamma^2 \Omega^2}{\gamma^2 T} + \frac{21 \Gamma \sigma^2}{eb}  +  \frac{12 C\Gamma^2 \Omega^2}{ \gamma^2 } \cdot \frac{1}{T} \sum\limits_{t=1}^{T} (1-\beta_{t}).
\end{align*}
If we choose $\bar z_T$ randomly and uniformly, then $ \E\left[ \| F(\bar z_T) \|^2\right] \leq \E\left[ \frac{1}{T} \sum\limits_{t=0}^{T-1}\| F(z_t) \|^2\right]$ and 
\begin{align*}
\E\left[ \| F(\bar z_T) \|^2\right]
\leq&  \frac{12\Gamma^2 \Omega^2}{\gamma^2 T} + \frac{21 \Gamma \sigma^2}{eb}  +  \frac{12 C\Gamma^2 \Omega^2}{ \gamma^2 } \cdot \frac{1}{T} \sum\limits_{t=1}^{T} (1-\beta_{t}).
\end{align*}
This completes the proof of the non-convex-non-concave case. $\square$

\subsection{Proof of Corollary \ref{cor:main_oasis}}

\textbf{Strongly convex--strongly concave case.} With $\beta_t \equiv \beta \geq 1 - \frac{\gamma \mu}{ 2 \Gamma C}$, we have \eqref{eq:th_sc} in the following form
\begin{align*}
\E\left[ R^2_{t+1} \right] 
 \leq&  \left( 1 - \frac{\gamma \mu}{2\Gamma}\right)\E\left[R^2_{t}\right] + \frac{12\gamma^2 \sigma^2}{eb}.
\end{align*}
Running the recursion, we have
\begin{align*}
\E\left[ R^2_{T} \right] 
 \leq&  \left( 1 - \frac{\gamma \mu}{2\Gamma}\right)^T\E\left[R^2_{0}\right] + \frac{12\gamma^2 \sigma^2}{eb} \sum\limits_{t=0}^{T-1} \left( 1 - \frac{\gamma \mu}{2\Gamma}\right)^t \\
 \leq&  \left( 1 - \frac{\gamma \mu}{2\Gamma}\right)^T\E\left[R^2_{0}\right] + \frac{12\gamma^2 \sigma^2}{eb} \sum\limits_{t=0}^{\infty} \left( 1 - \frac{\gamma \mu}{2\Gamma}\right)^t \\
 \leq&  \left( 1 - \frac{\gamma \mu}{2\Gamma}\right)^T\E\left[R^2_{0}\right] + \frac{24\Gamma \gamma \sigma^2}{eb\mu} \\
 \leq&  \exp\left( - \frac{\gamma \mu T}{2\Gamma}\right)\E\left[R^2_{0}\right] + \frac{24\Gamma \gamma \sigma^2}{eb\mu}.
\end{align*}
Finally, we need tuning of $\gamma \leq \frac{e}{4 L}$:

$\bullet$ If $\frac{e}{4 L} \geq \frac{2\Gamma \ln\left( \max\{2, eb\mu^2 R^2_0 T/(48 \Gamma^2\sigma^2) \} \right)}{\mu T}$ then $\gamma = \frac{2\Gamma \ln\left( \max\{2, eb\mu^2 R^2_0 T/(48 \Gamma^2\sigma^2) \} \right)}{\mu T}$ gives
    \begin{align*}
    \mathcal{\tilde O} \left( \exp\left(- \ln\left( \max\{2, eb\mu^2 R^2_0 T/(48 \Gamma^2\sigma^2) \} \right) \right) R_0^2 + \frac{\Gamma^2 \sigma^2}{e \mu^2 T}\right) = \mathcal{\tilde O} \left( \frac{\Gamma^2 \sigma^2}{e b\mu^2 T} \right).  
    \end{align*}

$\bullet$ If $\frac{e}{4 L} \leq \frac{2\Gamma \ln\left( \max\{2, eb\mu^2 R^2_0 T/(48 \Gamma^2\sigma^2) \} \right)}{\mu T}$ then $\gamma = \frac{e}{4 L}$ gives
    \begin{align*}
    \mathcal{\tilde O} \left( \exp\left(- \frac{e\mu T}{8\Gamma L}\right) R_0^2 + \frac{\Gamma \gamma \sigma^2}{eb\mu}\right) \leq \mathcal{\tilde O} \left( \exp\left(- \frac{e\mu T}{8\Gamma L}\right) R_0^2 + \frac{\Gamma^2 \sigma^2}{eb \mu^2 T}\right).  
    \end{align*}
What in the end gives that
\begin{align*}
 \E\left[ R^2_{T} \right] = \mathcal{\tilde O} \left( \exp\left(- \frac{e \mu T}{8 \Gamma L}\right) R_0^2  + \frac{\Gamma^2 \sigma^2}{e b\mu^2 T}\right).  
\end{align*}
This completes the proof of the strongly convex--strongly concave case. $\square$

\textbf{Convex--concave case.}  With $\beta_t \equiv \beta \geq 1 - \frac{1}{CT}$, we have \eqref{eq:th_c} in the following form
 \begin{align*} 
 \E\left[\text{gap}(x_{T}^{avg}, y_{T}^{avg})\right] \leq& \frac{3\Gamma\Omega^2}{\gamma T} + 
\frac{2 \Gamma \Omega^2}{\gamma T} + \frac{4\gamma \sigma^2}{eb}.
 \end{align*}
If $\gamma = \min\left\{\frac{e}{2L}; \frac{\sqrt{\Gamma e b}\Omega}{\sigma \sqrt{T}} \right\}$, then
 \begin{align*} 
 \E\left[\text{gap}(x_{T}^{avg}, y_{T}^{avg})\right] =& \mathcal{O}\left(\frac{\Gamma L\Omega^2}{e T} + 
 \frac{\sqrt{\Gamma}\sigma \Omega}{\sqrt{e b T}} \right).
 \end{align*}
 This completes the proof of the convex--concave case. $\square$

\textbf{Non-convex--non-concave case.} With $\beta_t \equiv \beta \geq 1 - \frac{1}{CT}$, we have \eqref{eq:th_nc} in the following form
\begin{align*}
\E\left[ \| \nabla_x f(\bar x_T, \bar y_T) \|^2 + \| \nabla_y f(\bar x_T, \bar y_T) \|^2\right]
\leq&  \frac{24\Gamma^2 \Omega^2}{\gamma^2 T} + \frac{21 \Gamma \sigma^2}{e b}.
\end{align*}
The choice of $\gamma = \frac{e}{3L}$ gives
\begin{align*}
\E\left[ \| \nabla_x f(\bar x_T, \bar y_T) \|^2 + \| \nabla_y f(\bar x_T, \bar y_T) \|^2\right]
\leq&  \frac{240\Gamma^2 L^2 \Omega^2}{e^2 T} + \frac{21 \Gamma \sigma^2}{e b}.
\end{align*}

The batch size $b \sim \frac{\Gamma \sigma^2}{e \varepsilon^2}$ completes the proof of the non-convex--non-concave case. $\square$
 

\subsection{Proof of  Corollary \ref{cor:main_adam}}

\textbf{Strongly convex--strongly concave case.}
We start from \eqref{eq:th_sc} with the small rearrangement
\begin{align*}
\frac{\gamma \mu}{2\Gamma} \E\left[R^2_{t}\right] 
 \leq&  \left( 1 - \frac{\gamma \mu}{2\Gamma} + (1-\beta_{t+1})C\right)\E\left[R^2_{t}\right] - \E\left[ R^2_{t+1} \right]  + \frac{6\gamma^2 \sigma^2}{eb} \left( 1 + (1-\beta_{t+1})C \right).
\end{align*}
Summing over all $t$ from $0$ to $T-1$ and averaging, we get
\begin{align*}
\frac{\gamma \mu}{2\Gamma T} \E\left[ \sum_{t=0}^{T-1} \|z_t - z^* \|^2_{\hat V_t}\right] 
 \leq&  \frac{\E\left[\|z_0 - z^* \|^2_{\hat V_0}\right]}{T} + 
\frac{1}{T}\sum_{t=0}^{T-1} \left( (1-\beta_{t+1})C - \frac{\gamma \mu}{2\Gamma} \right)\E\left[\|z_t - z^* \|^2_{\hat V_t}\right]  \\
&+ \frac{6\gamma^2 \sigma^2}{eb} \cdot \frac{1}{T}\sum_{t=0}^{T-1} \left( 1 + (1-\beta_{t+1})C \right).
\end{align*}
Using $e I \preccurlyeq \hat{V}_t \preccurlyeq \Gamma I$ and $\|z_t\| \leq \Omega$, we have
\begin{align*}
\frac{e\gamma \mu}{2\Gamma T} \E\left[ \sum_{t=0}^{T-1} \|z_t - z^* \|^2\right] 
\leq&  \frac{4\Gamma \Omega^2}{T} + 
\frac{1}{T}\sum_{t=1}^{T} \left( (1-\beta_{t})C - \frac{\gamma \mu}{2\Gamma} \right)\E\left[\|z_{t-1} - z^* \|^2_{\hat V_{t-1}}\right]  \\
&+ \frac{6\gamma^2 \sigma^2}{eb} \cdot \frac{1}{T}\sum_{t=1}^{T} \left( 1 + (1-\beta_{t})C \right).
\end{align*}
Let us define $\alpha = \left(\frac{ 2 \Gamma C}{\gamma \mu}\right)^2$. Then, $1 - \beta_t = \frac{1 - \beta}{1 - \beta^{t+1}}$ with $\beta = 1 - \frac{1}{\alpha}$. \eqref{eq:tech_lem} gives 
$$
\beta^{\sqrt{\alpha}} \leq 1 - \frac{1}{2\sqrt{\alpha}}.
$$
And hence, for all $t \geq \sqrt{\alpha}$
$$
1 - \beta_t \leq  \frac{2\sqrt{\alpha}}{\alpha} =  \frac{2}{\sqrt{\alpha}} = \frac{\gamma \mu}{2 \Gamma C}.
$$
Then we get
\begin{align*}
\frac{e\gamma \mu}{2\Gamma T} \E\left[ \sum_{t=0}^{T-1} \|z_t - z^* \|^2\right] 
\leq&  \frac{4\Gamma \Omega^2}{T} + 
\frac{1}{T}\sum_{t=1}^{\sqrt{\alpha}} \left( (1-\beta_{t})C - \frac{\gamma \mu}{2\Gamma} \right)\E\left[\|z_{t-1} - z^* \|^2_{\hat V_{t-1}}\right]  \\
&+ \frac{6\gamma^2 \sigma^2}{eb} \cdot \frac{1}{T}\sum_{t=1}^{T} \left( 1 + (1-\beta_{t})C \right) \\
\leq&  \frac{4\Gamma \Omega^2}{T} + 
\frac{1}{T}\sum_{t=1}^{\sqrt{\alpha}} 4C\Gamma \Omega^2  + \frac{6\gamma^2 \sigma^2}{eb} \cdot \frac{1}{T}\sum_{t=1}^{T} \left( 1 + (1-\beta_{t})C \right) \\
\leq&  \frac{4\Gamma \Omega^2}{T} + 
\frac{4C\Gamma \Omega^2 }{T} \cdot \frac{ 2 \Gamma C}{\gamma \mu}  + \frac{6\gamma^2 \sigma^2}{eb} \cdot \frac{1}{T}\sum_{t=1}^{T} \left( 1 + (1-\beta_{t})C \right) \\
\leq&  \frac{12C^2\Gamma^2 \Omega^2 }{\gamma \mu T} + \frac{12 C\gamma^2 \sigma^2}{eb}.
\end{align*}
Then,
\begin{align*}
\E\left[ \frac{1}{T} \sum_{t=0}^{T-1} \|z_t - z^* \|^2\right] 
\leq&  \frac{24 C^2\Gamma^3 \Omega^2 }{e\gamma^2 \mu^2 T} + \frac{24 \Gamma C\gamma \sigma^2}{e^2 \mu b}.
\end{align*}

Finally, $\gamma = \min\left\{ \frac{e}{4L} ; \sqrt[3]{\frac{C \Gamma^2 \Omega^2 e b}{\mu \sigma^2 T}} \right\}$ gives
\begin{align*}
\E\left[ \left\|  \frac{1}{T}\sum_{t=0}^{T-1} z_t - z^* \right\|^2\right] \leq \E\left[ \frac{1}{T}\sum_{t=0}^{T-1} \|z_t - z^* \|^2\right] 
=& \mathcal{O} \left(\frac{C^2\Gamma^3 L^2 \Omega^2 }{e^3 \mu^2 T} + \left(\frac{C^4 \Gamma^5 \sigma^4 \Omega^2}{e^5 \mu^4 b^2 T}\right)^{1/3}\right).
\end{align*}

This completes the proof of the strongly convex--strongly concave case. $\square$

\textbf{Convex--concave case.}
With $\beta_t = \frac{\beta - \beta^{t+1}}{1 - \beta^{t+1}}$ we get \eqref{eq:th_c} in the following form
\begin{align*} 
 \E\left[\text{gap}(x_{T}^{avg}, y_{T}^{avg})\right] \leq& \frac{3\Gamma\Omega^2}{\gamma T} + \frac{4\gamma \sigma^2}{eb} + 
\frac{2 (1 - \beta) C\Gamma \Omega^2}{\gamma T} \sum\limits_{t=1}^{T}  \frac{1}{1 - \beta^{t+1}} \\
\leq& \frac{3\Gamma\Omega^2}{\gamma T} + \frac{4\gamma \sigma^2}{e b} \\
&+ 
\frac{2 (1 - \beta) C\Gamma \Omega^2}{\gamma T} \sum\limits_{t=1}^{\sqrt{T}}  \frac{1}{1 - \beta} + 
\frac{2 (1 - \beta) C\Gamma \Omega^2}{\gamma T} \sum\limits_{t=1}^{T}  \frac{1}{1 - \beta^{\sqrt{T}}} \\
\leq& \frac{3\Gamma\Omega^2}{\gamma T} + \frac{4\gamma \sigma^2}{e b} \\
&+ 
\frac{2 C\Gamma \Omega^2}{\gamma \sqrt{T}} + 
\frac{2 (1 - \beta) C\Gamma \Omega^2}{\gamma T} \sum\limits_{t=1}^{T}  \frac{1}{1 - \beta^{\sqrt{T}}}.
 \end{align*}
Next, we substitute $\beta = 1 - \frac{1}{T}$. From \eqref{eq:tech_lem} we get that $\beta^{\sqrt{T}} \leq 1 - \frac{1}{2\sqrt{T}}$. Then, we get
\begin{align*} 
 \E\left[\text{gap}(x_{T}^{avg}, y_{T}^{avg})\right] 
\leq& \frac{3\Gamma\Omega^2}{\gamma T} + \frac{4\gamma \sigma^2}{eb} \\
&+ 
\frac{2 C\Gamma \Omega^2}{\gamma \sqrt{T}} + 
\frac{4 (1 - \beta) C\Gamma \Omega^2 \sqrt{T}}{\gamma} \\
\leq& \frac{3\Gamma\Omega^2}{\gamma T} + \frac{4\gamma \sigma^2}{eb} + 
\frac{2 C\Gamma \Omega^2}{\gamma \sqrt{T}} + 
\frac{4 C\Gamma \Omega^2 }{\gamma \sqrt{T}}.
 \end{align*}
It means that 
\begin{align*} 
 \E\left[\text{gap}(x_{T}^{avg}, y_{T}^{avg})\right] 
=& \mathcal{O}\left(\frac{C\Gamma \Omega^2}{\gamma \sqrt{T}} + \frac{\gamma \sigma^2}{eb} \right).
 \end{align*}
 With $\gamma = \min\left\{\frac{e}{2L}; \frac{\sqrt{C\Gamma e b}\Omega}{\sigma {T^{1/4}}} \right\}$
\begin{align*} 
 \E\left[\text{gap}(x_{T}^{avg}, y_{T}^{avg})\right] 
=& \mathcal{O}\left(\frac{C\Gamma L\Omega^2}{e\sqrt{T}} + \frac{\sqrt{C\Gamma e b} \sigma \Omega}{e T^{1/4}} \right).
 \end{align*}
 This completes the proof of the convex--concave case. $\square$

\textbf{Non-convex--non-concave case.}
With $\beta_t = \frac{\beta - \beta^{t+1}}{1 - \beta^{t+1}}$ we get \eqref{eq:th_nc} in the following form
\begin{align*}
\E\left[ \| \nabla_x f(\bar x_T, \bar y_T) \|^2 + \| \nabla_y f(\bar x_T, \bar y_T) \|^2\right]
\leq&  \frac{12\Gamma^2 \Omega^2}{\gamma^2 T} + \frac{21 \Gamma \sigma^2}{e b}  +  \frac{12 (1 - \beta) C\Gamma^2 \Omega^2}{ \gamma^2 T } \sum\limits_{t=1}^{T} \frac{1}{1-\beta^{t+1}} \\
\leq& \frac{12\Gamma^2 \Omega^2}{\gamma^2 T} + \frac{21 \Gamma \sigma^2}{e b}  +  \frac{12 (1 - \beta) C\Gamma^2 \Omega^2}{ \gamma^2 T } \sum\limits_{t=1}^{\sqrt{T}} \frac{1}{1-\beta} \\
&+  \frac{12 (1 - \beta) C\Gamma^2 \Omega^2}{ \gamma^2 T } \sum\limits_{t=1}^{T} \frac{1}{1-\beta^{\sqrt{T}}} \\
\leq& \frac{12\Gamma^2 \Omega^2}{\gamma^2 T} + \frac{21 \Gamma \sigma^2}{e b}  +  \frac{12 C\Gamma^2 \Omega^2}{ \gamma^2 \sqrt{T} } \\
&+  \frac{12 (1 - \beta) C\Gamma^2 \Omega^2}{ \gamma^2} \cdot \frac{1}{1-\beta^{\sqrt{T}}}.
\end{align*}
Next, we substitute $\beta = 1 - \frac{1}{T}$. From \eqref{eq:tech_lem} we get that $\beta^{\sqrt{T}} \leq 1 - \frac{1}{2\sqrt{T}}$. Then, we get
\begin{align*}
\E\left[ \| \nabla_x f(\bar x_T, \bar y_T) \|^2 + \| \nabla_y f(\bar x_T, \bar y_T) \|^2\right]
\leq&  \frac{12\Gamma^2 \Omega^2}{\gamma^2 T} + \frac{21 \Gamma \sigma^2}{e b}  +  \frac{12 C\Gamma^2 \Omega^2}{ \gamma^2 \sqrt{T} } +  \frac{24 C\Gamma^2 \Omega^2 }{ \gamma^2 \sqrt{T}} \\
\leq&  \frac{48 C\Gamma^2 \Omega^2 }{ \gamma^2 \sqrt{T}} + \frac{21 \Gamma \sigma^2}{e b} .
\end{align*}
The batch size $b \sim \frac{\Gamma \sigma^2}{e \varepsilon^2}$ and $\gamma = \frac{e}{3L}$ complete the proof of the non-convex--non-concave case. $\square$
 
 \newpage

\subsection{Proof of Theorem \ref{th:main1}}

We start the same way as in the proof of Theorem \ref{th:main0}. Applying Lemma \ref{l1} with $z^+ = z_{t+1}$,  $z=z_{t}$, $u = z$, $y = - \eta (\hat V_{t-1/2})^{-1}(w_t - z_t)  +\gamma \hat V^{-1}_{t-1/2} g_{t+1/2}$ and $D = \hat V_{t-1/2}$, we get
\begin{align*}
 \|z_{t+1} - z \|^2_{\hat V_{t-1/2}} =& \| z_t - z \|^2_{\hat V_{t-1/2}} + 2\eta \langle w_t - z_t, z_{t+1/2} - z\rangle \\
 &- 2 \gamma \langle g_{t+1/2},  z_{t+1} - z \rangle - \| z_{t+1} -  z_t \|^2_{\hat V_{t-1/2}},
 \end{align*}
and with $z^+ =  z_{t+1/2}$,  $z= z_{t}$, $u = z_{t+1}$, $y = \gamma V^{-1}_{t-1/2} g_{t-1/2}$, $D = \hat V_{t-1/2}$:
\begin{align*}
 \| z_{t+1/2} -  z_{t+1} \|^2_{\hat V_{t-1/2}} &= \| z_t - z_{t+1} \|^2_{\hat V_{t-1/2}} - 2 \gamma \langle g_{t-1/2},  z_{t+1/2} -  z_{t+1} \rangle - \| z_{t+1/2} -  z_t \|^2_{\hat V_{t-1/2}}.
 \end{align*}
Next, we sum up the two previous equalities
 \begin{align*}
 \|z_{t+1} - z \|^2_{\hat V_{t-1/2}} + \| z_{t+1/2} -  z_{t+1} \|^2_{\hat V_{t-1/2}} =& \| z_t - z \|^2_{\hat V_{t-1/2}} - \| z_{t+1/2} -  z_t \|^2_{\hat V_{t-1/2}} \\
  &- 2 \gamma \langle g_{t+1/2},  z_{t+1} - z \rangle \\
  &- 2 \gamma \langle g_{t-1/2},  z_{t+1/2} -  z_{t+1} \rangle \\
  &+ 2\eta \langle w_t - z_t, z_{t+1/2} - z\rangle.
 \end{align*}
A small rearrangement gives
\begin{align*}
\|z_{t+1} - z \|^2_{\hat V_{t-1/2}} +& \| z_{t+1/2} -  z_{t+1} \|^2_{\hat V_{t-1/2}} \nonumber\\ 
=& \| z_t - z \|^2_{\hat V_{t-1/2}} - \| z_{t+1/2} -  z_t \|^2_{\hat V_{t-1/2}} \nonumber\\ 
 &- 2 \gamma \langle g_{t+1/2},  z_{t+1/2} - z \rangle - 2 \gamma \langle g_{t-1/2} - g_{t+1/2},  z_{t+1/2} -  z_{t+1} \rangle \\
  &+ 2\eta \langle w_t - z, z_{t+1/2} - z\rangle + 2\eta \langle z - z_t, z_{t+1/2} - z\rangle
 \nonumber\\
 \leq& \| z_t - z \|^2_{\hat V_{t-1/2}} - \| z_{t+1/2} -  z_t \|^2_{\hat V_{t-1/2}}  - 2 \gamma \langle  g^{t+1/2},  z_{t+1/2} - z \rangle \nonumber\\ 
 &  + \gamma^2 \| g_{t+1/2} -  g_{t-1/2}\|^2_{\hat V^{-1}_{t-1/2}} + \|  z_{t+1/2} -  z_{t+1}\|^2_{\hat V_{t-1/2}} \\
 & + \eta \|w_t - z \|^2 + \eta \| z_{t+1/2} - z \|^2 - \eta \| z_{t+1/2} - w_t\|^2 \\
 &-\eta \| z - z_t \|^2 - \eta \| z_{t+1/2} - z\|^2 + \eta \| z_{t+1/2} - z_t\|^2 \\
 \leq& \| z_t - z \|^2_{\hat V_{t-1/2}} - \| z_{t+1/2} -  z_t \|^2_{\hat V_{t-1/2}}  - 2 \gamma \langle  g^{t+1/2},  z_{t+1/2} - z \rangle \nonumber\\ 
 &  + \gamma^2 \| g_{t+1/2} -  g_{t-1/2}\|^2_{\hat V^{-1}_{t-1/2}} + \|  z_{t+1/2} -  z_{t+1}\|^2_{\hat V_{t-1/2}} \\
 & + \eta \|w_t - z \|^2 -\eta \| z - z_t \|^2 + \eta \| z_{t+1/2} - z_t\|^2 - \eta \| z_{t+1/2} - w_t\|^2.
 \end{align*}
And then, 
\begin{align}
\label{eq:temp1}
\|z_{t+1} - z \|^2_{\hat V_{t-1/2}} \leq& \| z_t - z \|^2_{\hat V_{t-1/2}} - \| z_{t+1/2} -  z_t \|^2_{\hat V_{t-1/2}}  - 2 \gamma \langle  g^{t+1/2},  z_{t+1/2} - z \rangle \nonumber\\ 
 &  + \gamma^2 \| g_{t+1/2} -  g_{t-1/2}\|^2_{\hat V^{-1}_{t-1/2}} \nonumber\\
 & + \eta \|w_t - z \|^2 -\eta \| z - z_t \|^2 + \eta \| z_{t+1/2} - z_t\|^2 - \eta \| z_{t+1/2} - w_t\|^2 \nonumber\\
 \leq& \| z_t - z \|^2_{\hat V_{t-1/2}} - \| z_{t+1/2} -  z_t \|^2_{\hat V_{t-1/2}}  - 2 \gamma \langle  g^{t+1/2},  z_{t+1/2} - z \rangle \nonumber\\ 
 &  + \frac{\gamma^2}{e} \| g_{t+1/2} -  g_{t-1/2}\|^2\nonumber\\
 & + \eta \|w_t - z \|^2 -\eta \| z - z_t \|^2 + \eta \| z_{t+1/2} - z_t\|^2 - \eta \| z_{t+1/2} - w_t\|^2
 \nonumber\\
 \leq& \| z_t - z \|^2_{\hat V_{t-1/2}} - \| z_{t+1/2} -  z_t \|^2_{\hat V_{t-1/2}}  - 2 \gamma \langle  g^{t+1/2},  z_{t+1/2} - z \rangle \nonumber\\ 
 &  + \frac{3\gamma^2}{e} \| F(z_{t+1/2}) -  F(z_{t-1/2})\|^2+ \frac{3\gamma^2}{e} \| g_{t+1/2} -  F(z_{t+1/2})\|^2\nonumber\\
 &+ \frac{3\gamma^2}{e} \| F(z_{t-1/2}) -  g_{t-1/2}\|^2 \nonumber\\
 & + \eta \|w_t - z \|^2 -\eta \| z - z_t \|^2 + \eta \| z_{t+1/2} - z_t\|^2 - \eta \| z_{t+1/2} - w_t\|^2.
 \end{align}
Using smoothness of $f$ and the update of Algorithm \ref{Scaled_ExtraGrad_mom}
\begin{align*}
\|& F(z_{t+1/2}) -  F(z_{t-1/2})\|^2 \nonumber\\
 \leq& L^2 \| z_{t+1/2} -  z_{t-1/2}\|^2 \nonumber\\
\leq& 2L^2 \| z_{t+1/2} -  z_{t}\|^2 + 2L^2 \| z_{t} -  z_{t-1/2}\|^2 \nonumber\\
\leq& 2L^2 \| z_{t+1/2} -  z_{t}\|^2 + \frac{2L^2}{e} \| z_{t} -  z_{t-1/2}\|^2_{\hat V_{t-3/2}} \nonumber\\
\leq& 2L^2 \| z_{t+1/2} -  z_{t}\|^2 \nonumber\\
&+ \frac{2L^2}{e} \| z_{t-1} + \eta (\hat V_{t-3/2})^{-1}(w_{t-1} - z_{t-1}) - \gamma (\hat V_{t-3/2})^{-1} g_{t-1/2} - z_{t-1} + \gamma (\hat V^x_{t-3/2})^{-1} g_{t-3/2}\|^2_{\hat V_{t-3/2}} \nonumber\\
\leq& 2L^2 \| z_{t+1/2} -  z_{t}\|^2 \nonumber\\
&+ \frac{2L^2}{e} \| \eta (w_{t-1} - z_{t-1}) - \gamma g_{t-1/2} + \gamma g_{t-3/2}\|^2_{\hat V^{-1}_{t-3/2}}.
 \end{align*}
Hence,
\begin{align}
\label{eq:temp2}
\| F(z_{t+1/2}) -  F(z_{t-1/2})\|^2 \leq& 2L^2 \| z_{t+1/2} -  z_{t}\|^2 \nonumber\\
&+ \frac{2L^2}{e^2} \| \eta (w_{t-1} - z_{t-1}) - \gamma g_{t-1/2} + \gamma g_{t-3/2}\|^2 \nonumber\\
\leq& 2L^2 \| z_{t+1/2} -  z_{t}\|^2 + \frac{4L^2 \eta^2}{e^2} \|  w_{t-1} - z_{t-1}\|^2\nonumber\\
&+ \frac{4L^2 \gamma^2}{e^2} \|  g_{t-1/2} -  g_{t-3/2}\|^2 \nonumber\\
\leq& 2L^2 \| z_{t+1/2} -  z_{t}\|^2 + \frac{4L^2 \eta^2}{e^2} \|  w_{t-1} - z_{t-1}\|^2\nonumber\\
&+ \frac{12L^2 \gamma^2}{e^2} \| F(z_{t-1/2}) -  F(z_{t-3/2})\|^2 \nonumber\\
&+ \frac{12L^2 \gamma^2}{e^2} \|  g_{t-1/2} -  F(z_{t-1/2})\|^2\nonumber\\
&+ \frac{12L^2 \gamma^2}{e^2} \| F(z_{t-3/2}) -  g_{t-3/2}\|^2.
 \end{align}
Combining \eqref{eq:temp1} and \eqref{eq:temp2}, we have
\begin{align*}
\|z_{t+1} - z \|^2_{\hat V_{t-1/2}} &+ \frac{3\gamma^2 M}{e} \| F(z_{t+1/2}) -  F(z_{t-1/2})\|^2
\\
\leq& \| z_t - z \|^2_{\hat V_{t-1/2}} - \| z_{t+1/2} -  z_t \|^2_{\hat V_{t-1/2}}  - 2 \gamma \langle  g^{t+1/2},  z_{t+1/2} - z \rangle \nonumber\\ 
 &  + \frac{3\gamma^2}{e} \| F(z_{t+1/2}) -  F(z_{t-1/2})\|^2+ \frac{3\gamma^2}{e} \| g_{t+1/2} -  F(z_{t+1/2})\|^2\nonumber\\
 &+ \frac{3\gamma^2}{e} \| F(z_{t-1/2}) -  g_{t-1/2}\|^2 \nonumber\\
 & + \eta \|w_t - z \|^2 -\eta \| z - z_t \|^2 + \eta \| z_{t+1/2} - z_t\|^2 - \eta \| z_{t+1/2} - w_t\|^2 \\
 &+ \frac{6\gamma^2 L^2 M}{e} \| z_{t+1/2} -  z_{t}\|^2 + \frac{12\gamma^2 L^2 \eta^2 M}{e^3}\|  w_{t-1} - z_{t-1}\|^2\nonumber\\
&+ \frac{36\gamma^4 L^2 M}{e^3} \| F(z_{t-1/2}) -  F(z_{t-3/2})\|^2 \nonumber\\
&+ \frac{36\gamma^4 L^2 M}{e^3} \|  g_{t-1/2} -  F(z_{t-1/2})\|^2+ \frac{36\gamma^4 L^2 M}{e^3} \| F(z_{t-3/2}) -  g_{t-3/2}\|^2.
 \end{align*}
 Here $M$ is some positive constant, which we will define later. Next, we get
\begin{align*}
\|z_{t+1} - z \|^2_{\hat V_{t-1/2}} &+ \frac{3\gamma^2 M}{e} \| F(z_{t+1/2}) -  F(z_{t-1/2})\|^2
\\
\leq& \| z_t - z \|^2_{\hat V_{t-1/2}} - \| z_{t+1/2} -  z_t \|^2_{\hat V_{t-1/2}}  - 2 \gamma \langle  g^{t+1/2},  z_{t+1/2} - z \rangle \nonumber\\ 
 &  + \frac{3\gamma^2}{e} \| F(z_{t+1/2}) -  F(z_{t-1/2})\|^2+ \frac{3\gamma^2}{e} \| g_{t+1/2} -  F(z_{t+1/2})\|^2\nonumber\\
 &+ \frac{3\gamma^2}{e} \| F(z_{t-1/2}) -  g_{t-1/2}\|^2 \nonumber\\
 & + \eta \|w_t - z \|^2 -\eta \| z - z_t \|^2 + \eta \| z_{t+1/2} - z_t\|^2 - \frac{\eta}{2} \| z_{t} - w_t\|^2 \\
 &+ \eta \| z_{t+1/2} - z_t\|^2 \\
 &+ \frac{6\gamma^2 L^2 M}{e} \| z_{t+1/2} -  z_{t}\|^2 + \frac{12\gamma^2 L^2 \eta^2 M}{e^3}\|  w_{t-1} - z_{t-1}\|^2\nonumber\\
&+ \frac{36\gamma^4 L^2 M}{e^3} \| F(z_{t-1/2}) -  F(z_{t-3/2})\|^2 \nonumber\\
&+ \frac{36\gamma^4 L^2 M}{e^3} \|  g_{t-1/2} -  F(z_{t-1/2})\|^2+ \frac{36\gamma^4 L^2 M}{e^3} \| F(z_{t-3/2}) -  g_{t-3/2}\|^2.
 \end{align*}
Small rearrangement gives
\begin{align*}
\|z_{t+1} - z \|^2_{\hat V_{t-1/2}} &+ \frac{3\gamma^2 (M - 1)}{e} \| F(z_{t+1/2}) -  F(z_{t-1/2})\|^2
+ \frac{\eta}{2} \| z_{t} - w_t\|^2 \\
\leq& \| z_t - z \|^2_{\hat V_{t-1/2}} -\eta \| z_t - z \|^2 + \eta \|w_t - z \|^2 - 2 \gamma \langle  g^{t+1/2},  z_{t+1/2} - z \rangle \\  
&+ \frac{12\gamma^2 L^2 M}{(M-1) e^2} \cdot \frac{3\gamma^2 (M-1)}{e}\| F(z_{t-1/2}) -  F(z_{t-3/2})\|^2 + \frac{12\gamma^2 L^2 \eta^2 M}{e^3}\|  z_{t-1} - w_{t-1} \|^2\nonumber\\ 
 &  + \frac{6\gamma^2 L^2 M}{e} \| z_{t+1/2} -  z_{t}\|^2 + 2\eta \| z_{t+1/2} - z_t\|^2 - \| z_{t+1/2} -  z_t \|^2_{\hat V_{t-1/2}} \\
&  + \frac{3\gamma^2}{e} \| g_{t+1/2} -  F(z_{t+1/2})\|^2 + \frac{3\gamma^2}{e} \| F(z_{t-1/2}) -  g_{t-1/2}\|^2 \nonumber\\
&+ \frac{36\gamma^4 L^2 M}{e^3} \|  g_{t-1/2} -  F(z_{t-1/2})\|^2+ \frac{36\gamma^4 L^2 M}{e^3} \| F(z_{t-3/2}) -  g_{t-3/2}\|^2.
 \end{align*}
With $M = 2$, $\gamma \leq \frac{e}{10 L}$ and $\eta \leq ep \leq \frac{e}{4}$, we have $\frac{12\gamma^2 L^2 M}{(M-1) e^2} \leq \frac{1}{2}$ and $\frac{12\gamma^2 L^2 \eta M}{e^3} \leq \frac{1}{4}$. Then,
\begin{align*}
\|z_{t+1} - z \|^2_{\hat V_{t-1/2}} &+ \frac{3\gamma^2}{e} \| F(z_{t+1/2}) -  F(z_{t-1/2})\|^2
+ \frac{\eta}{2} \| z_{t} - w_t\|^2 \\
\leq& \| z_t - z \|^2_{\hat V_{t-1/2}} -\eta \| z_t - z \|^2 + \eta \|w_t - z \|^2 - 2 \gamma \langle  g^{t+1/2},  z_{t+1/2} - z \rangle \\  
&+ \frac{1}{2} \cdot \frac{3\gamma^2}{e}\| F(z_{t-1/2}) -  F(z_{t-3/2})\|^2 + \frac{1}{2} \cdot \frac{\eta}{2}\|  z_{t-1} - w_{t-1} \|^2\nonumber\\ 
 &  + \frac{e}{4} \| z_{t+1/2} -  z_{t}\|^2 + \frac{e}{2} \| z_{t+1/2} - z_t\|^2 - \| z_{t+1/2} -  z_t \|^2_{\hat V_{t-1/2}} \\
&  + \frac{3\gamma^2}{e} \| g_{t+1/2} -  F(z_{t+1/2})\|^2 + \frac{3\gamma^2}{e} \| F(z_{t-1/2}) -  g_{t-1/2}\|^2 \nonumber\\
&+ \frac{3\gamma^2 L^2}{e} \|  g_{t-1/2} -  F(z_{t-1/2})\|^2+ \frac{3\gamma^2 L^2}{e} \| F(z_{t-3/2}) -  g_{t-3/2}\|^2.
 \end{align*}
Using $I \preccurlyeq \frac{1}{e}\hat{V}_{t-1/2}$, we obtain
\begin{align}
\label{eq:t5}
\|z_{t+1} - z \|^2_{\hat V_{t-1/2}} &+ \frac{3\gamma^2}{e} \| F(z_{t+1/2}) -  F(z_{t-1/2})\|^2
+ \frac{\eta}{2} \| z_{t} - w_t\|^2 \nonumber\\
\leq& \| z_t - z \|^2_{\hat V_{t-1/2}} -\eta \| z_t - z \|^2 + \eta \|w_t - z \|^2 - 2 \gamma \langle  g^{t+1/2},  z_{t+1/2} - z \rangle \nonumber\\  
&+ \frac{1}{2} \cdot \frac{3\gamma^2}{e}\| F(z_{t-1/2}) -  F(z_{t-3/2})\|^2 + \frac{1}{2} \cdot \frac{\eta}{2}\|  z_{t-1} - w_{t-1} \|^2\nonumber\\ 
 &  + \frac{1}{4} \| z_{t+1/2} -  z_{t}\|^2_{\hat V_{t-1/2}} + \frac{1}{2} \| z_{t+1/2} - z_t\|^2_{\hat V_{t-1/2}} - \| z_{t+1/2} -  z_t \|^2_{\hat V_{t-1/2}} \nonumber\\
&  + \frac{3\gamma^2}{e} \| g_{t+1/2} -  F(z_{t+1/2})\|^2 + \frac{3\gamma^2}{e} \| F(z_{t-1/2}) -  g_{t-1/2}\|^2 \nonumber\\
&+ \frac{3\gamma^2}{e} \|  g_{t-1/2} -  F(z_{t-1/2})\|^2+ \frac{3\gamma^2}{e} \| F(z_{t-3/2}) -  g_{t-3/2}\|^2 \nonumber\\
\leq& \| z_t - z \|^2_{\hat V_{t-1/2}} -\eta \| z_t - z \|^2 + \eta \|w_t - z \|^2 - 2 \gamma \langle  g^{t+1/2},  z_{t+1/2} - z \rangle \nonumber\\  
&+ \frac{1}{2} \cdot \frac{3\gamma^2}{e}\| F(z_{t-1/2}) -  F(z_{t-3/2})\|^2 + \frac{1}{2} \cdot \frac{\eta}{2}\|  z_{t-1} - w_{t-1} \|^2\nonumber\\ 
 &  - \frac{1}{4}\| z_{t+1/2} -  z_t \|^2_{\hat V_{t-1/2}} + \frac{3\gamma^2}{e} \| g_{t+1/2} -  F(z_{t+1/2})\|^2  \nonumber\\
&+ \frac{6\gamma^2}{e} \| F(z_{t-1/2}) -  g_{t-1/2}\|^2 + \frac{3\gamma^2}{e} \| F(z_{t-3/2}) -  g_{t-3/2}\|^2.
 \end{align}
\textbf{Strongly convex--strongly concave case.} We substitute $z = z^*$  and take the total expectation of both sides of the equation
\begin{align*}
\E\left[\|z_{t+1} - z^* \|^2_{\hat V_{t-1/2}}\right] &+ \frac{3\gamma^2}{e} \E\left[\| F(z_{t+1/2}) -  F(z_{t-1/2})\|^2\right]
+ \frac{\eta}{2} \E\left[\| z_{t} - w_t\|^2\right] \\
\leq& \E\left[\| z_t - z^* \|^2_{\hat V_{t-1/2}}\right] -\eta \E\left[\| z_t - z^* \|^2\right] + \eta \E\left[\|w_t - z^* \|^2\right] \\
&- 2 \gamma \E\left[\langle  \E_{t+1/2}\left[g^{t+1/2}\right],  z_{t+1/2} - z^* \rangle\right] \\  
&+ \frac{1}{2} \cdot \frac{3\gamma^2}{e} \E\left[\| F(z_{t-1/2}) -  F(z_{t-3/2})\|^2\right] + \frac{1}{2} \cdot \frac{\eta}{2} \E\left[\|  z_{t-1} - w_{t-1} \|^2\right]\nonumber\\ 
 &  - \frac{1}{4} \E\left[\| z_{t+1/2} -  z_t \|^2_{\hat V_{t-1/2}}\right] + \frac{3\gamma^2}{e} \E\left[\| g_{t+1/2} -  F(z_{t+1/2})\|^2\right]  \nonumber\\
&+ \frac{6\gamma^2}{e} \E\left[\| F(z_{t-1/2}) -  g_{t-1/2}\|^2\right] + \frac{3\gamma^2}{e} \E\left[\| F(z_{t-3/2}) -  g_{t-3/2}\|^2\right].
 \end{align*}
Next, using the property of the solution $z^*$: $\langle F(z^*),  z_{t+1/2} - z^*\rangle = 0$, and get
\begin{align*}
\E\left[\|z_{t+1} - z^* \|^2_{\hat V_{t-1/2}}\right] &+ \frac{3\gamma^2}{e} \E\left[\| F(z_{t+1/2}) -  F(z_{t-1/2})\|^2\right]
+ \frac{\eta}{2} \E\left[\| z_{t} - w_t\|^2\right] \\
\leq& \E\left[\| z_t - z^* \|^2_{\hat V_{t-1/2}}\right] -\eta \E\left[\| z_t - z^* \|^2\right] + \eta \E\left[\|w_t - z^* \|^2\right] \\
&- 2 \gamma \E\left[\langle  F(z_{t+1/2}) - z^*,  z_{t+1/2} - z^* \rangle\right] \\  
&+ \frac{1}{2} \cdot \frac{3\gamma^2}{e} \E\left[\| F(z_{t-1/2}) -  F(z_{t-3/2})\|^2\right] + \frac{1}{2} \cdot \frac{\eta}{2} \E\left[\|  z_{t-1} - w_{t-1} \|^2\right]\nonumber\\ 
 &  - \frac{1}{4} \E\left[\| z_{t+1/2} -  z_t \|^2_{\hat V_{t-1/2}}\right] + \frac{3\gamma^2}{e} \E\left[\| g_{t+1/2} -  F(z_{t+1/2})\|^2\right]  \nonumber\\
&+ \frac{6\gamma^2}{e} \E\left[\| F(z_{t-1/2}) -  g_{t-1/2}\|^2\right] + \frac{3\gamma^2}{e} \E\left[\| F(z_{t-3/2}) -  g_{t-3/2}\|^2\right].
 \end{align*}
Assumptions \ref{as:conv} and \ref{as:var} on strongly convexity - strongly concavity and on stochastisity give
\begin{align*}
\E\left[\|z_{t+1} - z^* \|^2_{\hat V_{t-1/2}}\right] &+ \frac{3\gamma^2}{e} \E\left[\| F(z_{t+1/2}) -  F(z_{t-1/2})\|^2\right]
+ \frac{\eta}{2} \E\left[\| z_{t} - w_t\|^2\right] \\
\leq& \E\left[\| z_t - z^* \|^2_{\hat V_{t-1/2}}\right] -\eta \E\left[\| z_t - z^* \|^2\right] + \eta \E\left[\|w_t - z^* \|^2\right] \\
&- 2 \gamma \mu \E\left[\| z_{t+1/2} - z^* \|^2\right] \\  
&+ \frac{1}{2} \cdot \frac{3\gamma^2}{e} \E\left[\| F(z_{t-1/2}) -  F(z_{t-3/2})\|^2\right] + \frac{1}{2} \cdot \frac{\eta}{2} \E\left[\|  z_{t-1} - w_{t-1} \|^2\right]\nonumber\\ 
 &  - \frac{1}{4} \E\left[\| z_{t+1/2} -  z_t \|^2_{\hat V_{t-1/2}}\right] + \frac{12\gamma^2 \sigma^2}{e b} \\
\leq& \E\left[\| z_t - z^* \|^2_{\hat V_{t-1/2}}\right] -\eta \E\left[\| z_t - z^* \|^2\right] + \eta \E\left[\|w_t - z^* \|^2\right] \\
&- \gamma \mu \E\left[\| z_{t} - z^* \|^2\right] + 2 \gamma \mu \E\left[\| z_{t+1/2} - z_t \|^2\right] \\  
&+ \frac{1}{2} \cdot \frac{3\gamma^2}{e} \E\left[\| F(z_{t-1/2}) -  F(z_{t-3/2})\|^2\right] + \frac{1}{2} \cdot \frac{\eta}{2} \E\left[\|  z_{t-1} - w_{t-1} \|^2\right]\nonumber\\ 
 &  - \frac{1}{4} \E\left[\| z_{t+1/2} -  z_t \|^2_{\hat V_{t-1/2}}\right] + \frac{12\gamma^2 \sigma^2}{e b}.
 \end{align*}
Using $I \preccurlyeq \frac{1}{e}\hat{V}_{t-1/2}$, we obtain
\begin{align*}
\E\left[\|z_{t+1} - z^* \|^2_{\hat V_{t-1/2}}\right] &+ \frac{3\gamma^2}{e} \E\left[\| F(z_{t+1/2}) -  F(z_{t-1/2})\|^2\right]
+ \frac{\eta}{2} \E\left[\| z_{t} - w_t\|^2\right] \\
\leq& \E\left[\| z_t - z^* \|^2_{\hat V_{t-1/2}}\right] -(\eta+ \gamma \mu) \E\left[\| z_t - z^* \|^2\right] + \eta \E\left[\|w_t - z^* \|^2\right] \\
&+ \frac{1}{2} \cdot \frac{3\gamma^2}{e} \E\left[\| F(z_{t-1/2}) -  F(z_{t-3/2})\|^2\right] + \frac{1}{2} \cdot \frac{\eta}{2} \E\left[\|  z_{t-1} - w_{t-1} \|^2\right]\nonumber\\ 
 &  - \left(\frac{1}{4} - \frac{2 \gamma \mu}{e} \right) \E\left[\| z_{t+1/2} -  z_t \|^2_{\hat V_{t-1/2}}\right] + \frac{12\gamma^2 \sigma^2}{e b}.
 \end{align*}
With $\gamma \leq \frac{e}{8 \mu}$,
\begin{align*}
\E\left[\|z_{t+1} - z^* \|^2_{\hat V_{t-1/2}}\right] &+ \frac{3\gamma^2}{e} \E\left[\| F(z_{t+1/2}) -  F(z_{t-1/2})\|^2\right]
+ \frac{\eta}{2} \E\left[\| z_{t} - w_t\|^2\right] \\
\leq& \E\left[\| z_t - z^* \|^2_{\hat V_{t-1/2}}\right] -(\eta+ \gamma \mu) \E\left[\| z_t - z^* \|^2\right] + \eta \E\left[\|w_t - z^* \|^2\right] \\
&+ \frac{1}{2} \cdot \frac{3\gamma^2}{e} \E\left[\| F(z_{t-1/2}) -  F(z_{t-3/2})\|^2\right] + \frac{1}{2} \cdot \frac{\eta}{2} \E\left[\|  z_{t-1} - w_{t-1} \|^2\right]\nonumber\\ 
 & + \frac{12\gamma^2 \sigma^2}{e b}.
 \end{align*}
The update of $w_{t+1}$ gives
\begin{align*}
\frac{\eta + \nicefrac{\gamma \mu}{2}}{p }\E\left[\|w_{t+1} - z^* \|^2\right] &=
\frac{\eta + \nicefrac{\gamma \mu}{2}}{p }\E\left[\E_{w_{t+1}}\left[\|w_{t+1} - z^* \|^2\right]\right] \\
&=
\left( \eta + \frac{\gamma \mu}{2}\right)\E\left[\| z_t - z^* \|^2\right] + \frac{(1-p)(\eta + \nicefrac{\gamma \mu}{2})}{p}\E\left[\|w_{t} - z^* \|^2\right].
 \end{align*}
Connecting with the previous equation, we get
\begin{align*}
\E\left[\|z_{t+1} - z^* \|^2_{\hat V_{t-1/2}}\right] &+ \frac{\eta + \nicefrac{\gamma \mu}{2}}{p}\E\left[\|w_{t+1} - z^* \|^2\right] \\
&+ \frac{3\gamma^2}{e} \E\left[\| F(z_{t+1/2}) -  F(z_{t-1/2})\|^2\right]
+ \frac{\eta}{2} \E\left[\| z_{t} - w_t\|^2\right] \\
\leq& \E\left[\| z_t - z^* \|^2_{\hat V_{t-1/2}}\right] -\frac{\gamma \mu}{2} \E\left[\| z_t - z^* \|^2\right]  \\
&+ (1-p)\cdot \frac{\eta + \nicefrac{\gamma \mu}{2}}{p}\E\left[\|w_{t} - z^* \|^2\right] + \eta \E\left[\|w_t - z^* \|^2\right]\\
&+ \frac{1}{2} \cdot \frac{3\gamma^2}{e} \E\left[\| F(z_{t-1/2}) -  F(z_{t-3/2})\|^2\right] + \frac{1}{2} \cdot \frac{\eta}{2} \E\left[\|  z_{t-1} - w_{t-1} \|^2\right]\nonumber\\ 
 & + \frac{12\gamma^2 \sigma^2}{e b}\\
\leq& \E\left[\| z_t - z^* \|^2_{\hat V_{t-1/2}}\right] -\frac{\gamma \mu}{2} \E\left[\| z_t - z^* \|^2\right]  \\
&+ \left(1- \frac{p \gamma \mu}{2\eta + \gamma \mu}\right)\cdot \frac{\eta + \nicefrac{\gamma \mu}{2}}{p}\E\left[\|w_{t} - z^* \|^2\right] \\
&+ \frac{1}{2} \cdot \frac{3\gamma^2}{e} \E\left[\| F(z_{t-1/2}) -  F(z_{t-3/2})\|^2\right] + \frac{1}{2} \cdot \frac{\eta}{2} \E\left[\|  z_{t-1} - w_{t-1} \|^2\right]\nonumber\\ 
 & + \frac{12\gamma^2 \sigma^2}{e b}.
 \end{align*}
Using $\frac{1}{\Gamma}\hat{V}_{t-1/2} \preccurlyeq  I$, we get
\begin{align*}
\E\left[\|z_{t+1} - z^* \|^2_{\hat V_{t-1/2}}\right] &+ \frac{\eta + \nicefrac{\gamma \mu}{2}}{p}\E\left[\|w_{t+1} - z^* \|^2\right] \\
&+ \frac{3\gamma^2}{e} \E\left[\| F(z_{t+1/2}) -  F(z_{t-1/2})\|^2\right]
+ \frac{\eta}{2} \E\left[\| z_{t} - w_t\|^2\right] \\
\leq& \left(1 - \frac{\gamma \mu}{2 \Gamma}\right)\E\left[\| z_t - z^* \|^2_{\hat V_{t-1/2}}\right]+ \left(1- \frac{p \gamma \mu}{2\eta + \gamma \mu}\right)\frac{\eta + \nicefrac{\gamma \mu}{2}}{p}\E\left[\|w_{t} - z^* \|^2\right] \\
&+ \frac{1}{2} \cdot \frac{3\gamma^2}{e} \E\left[\| F(z_{t-1/2}) -  F(z_{t-3/2})\|^2\right] + \frac{1}{2} \cdot \frac{\eta}{2} \E\left[\|  z_{t-1} - w_{t-1} \|^2\right]\nonumber\\ 
 & + \frac{12\gamma^2 \sigma^2}{e b}.
 \end{align*}
 Lemma \ref{lem:precond} with $C = \frac{\Gamma^2}{2e^2}$ for \eqref{eq:precond}, $C = \frac{2\Gamma}{e}$ for \eqref{eq:precond_add} and $B_{t+1} = \left( 1 + (1-\beta_{t+1}) C \right)$ gives
 \begin{align*}
 \frac{1}{B_{t+1}}\E\left[\|z_{t+1} - z^* \|^2_{\hat V_{t+1/2}}\right] &+ \frac{\eta + \nicefrac{\gamma \mu}{2}}{p}\E\left[\|w_{t+1} - z^* \|^2\right] \\
&+ \frac{3\gamma^2}{e} \E\left[\| F(z_{t+1/2}) -  F(z_{t-1/2})\|^2\right]
+ \frac{\eta}{2} \E\left[\| z_{t} - w_t\|^2\right] \\
\leq&
\E\left[\|z_{t+1} - z^* \|^2_{\hat V_{t-1/2}}\right] + \frac{\eta + \nicefrac{\gamma \mu}{2}}{p}\E\left[\|w_{t+1} - z^* \|^2\right] \\
&+ \frac{3\gamma^2}{e} \E\left[\| F(z_{t+1/2}) -  F(z_{t-1/2})\|^2\right]
+ \frac{\eta}{2} \E\left[\| z_{t} - w_t\|^2\right] \\
\leq& \left(1 - \frac{\gamma \mu}{2 \Gamma}\right)\E\left[\| z_t - z^* \|^2_{\hat V_{t-1/2}}\right]+ \left(1- \frac{p \gamma \mu}{2\eta + \gamma \mu}\right)\frac{\eta + \nicefrac{\gamma \mu}{2}}{p}\E\left[\|w_{t} - z^* \|^2\right] \\
&+ \frac{1}{2} \cdot \frac{3\gamma^2}{e} \E\left[\| F(z_{t-1/2}) -  F(z_{t-3/2})\|^2\right] + \frac{1}{2} \cdot \frac{\eta}{2} \E\left[\|  z_{t-1} - w_{t-1} \|^2\right]\nonumber\\ 
 & + \frac{12\gamma^2 \sigma^2}{e b}.
 \end{align*}
And then,
 \begin{align*}
\E\left[\|z_{t+1} - z^* \|^2_{\hat V_{t+1/2}}\right] &+ \frac{(\eta + \nicefrac{\gamma \mu}{2})B_{t+1}}{p}\E\left[\|w_{t+1} - z^* \|^2\right] \\
&+ \frac{3\gamma^2 B_{t+1}}{e} \E\left[\| F(z_{t+1/2}) -  F(z_{t-1/2})\|^2\right]
+ \frac{\eta B_{t+1}}{2} \E\left[\| z_{t} - w_t\|^2\right] \\
\leq& \left(1 - \frac{\gamma \mu}{2 \Gamma}\right)B_{t+1} \E\left[\| z_t - z^* \|^2_{\hat V_{t-1/2}}\right]+ \left(1- \frac{p \gamma \mu}{2\eta + \gamma \mu}\right)\frac{(\eta + \nicefrac{\gamma \mu}{2}) B_{t+1}}{p}\E\left[\|w_{t} - z^* \|^2\right] \\
&+ \frac{1}{2} \cdot \frac{3\gamma^2 B_{t+1}}{e} \E\left[\| F(z_{t-1/2}) -  F(z_{t-3/2})\|^2\right] + \frac{1}{2} \cdot \frac{\eta B_{t+1}}{2} \E\left[\|  z_{t-1} - w_{t-1} \|^2\right]\nonumber\\ 
 & + \frac{12\gamma^2 \sigma^2 B_{t+1}}{e b}.
 \end{align*}
One can note that in Section \ref{sec:scaled} we use $\beta_t \equiv \beta$ or
$\beta_t = \frac{\beta - \beta^t}{1 - \beta^t}$. For such $\beta_t$, it holds that $B_{t+1} \leq B_t$. Then,
 \begin{align*}
\E\left[\|z_{t+1} - z^* \|^2_{\hat V_{t+1/2}}\right] &+ \frac{(\eta + \nicefrac{\gamma \mu}{2})B_{t+1}}{p}\E\left[\|w_{t+1} - z^* \|^2\right] \\
&+ \frac{3\gamma^2 B_{t+1}}{e} \E\left[\| F(z_{t+1/2}) -  F(z_{t-1/2})\|^2\right]
+ \frac{\eta B_{t+1}}{2} \E\left[\| z_{t} - w_t\|^2\right] \\
\leq& \left(1 - \frac{\gamma \mu}{2 \Gamma}\right)B_{t} \E\left[\| z_t - z^* \|^2_{\hat V_{t-1/2}}\right]+ \left(1- \frac{p \gamma \mu}{2\eta + \gamma \mu}\right)\frac{(\eta + \nicefrac{\gamma \mu}{2}) B_{t}}{p}\E\left[\|w_{t} - z^* \|^2\right] \\
&+ \frac{1}{2} \cdot \frac{3\gamma^2 B_{t}}{e} \E\left[\| F(z_{t-1/2}) -  F(z_{t-3/2})\|^2\right] + \frac{1}{2} \cdot \frac{\eta B_{t}}{2} \E\left[\|  z_{t-1} - w_{t-1} \|^2\right]\nonumber\\ 
 & + \frac{12\gamma^2 \sigma^2 B_{t+1}}{e b}.
 \end{align*}
 The notation 
\begin{equation}
\label{eq:psi}
\begin{split}
    \Psi_{t+1} =& \|z_{t+1} - z^* \|^2_{\hat V_{t+1/2}} + \frac{(\eta + \nicefrac{\gamma \mu}{2})B_{t+1}}{p}\|w_{t+1} - z^* \|^2 \nonumber\\
&+ \frac{3\gamma^2 B_{t+1}}{e} \| F(z_{t+1/2}) -  F(z_{t-1/2})\|^2
+ \frac{\eta B_{t+1}}{2} \| z_{t} - w_t\|^2
\end{split}
\end{equation}
gives
 \begin{align*}
\E\left[\Psi_{t+1}\right] 
\leq& \max \left[ \left(1 - \frac{\gamma \mu}{2 \Gamma}\right)\left( 1 + (1-\beta_{t+1}) C \right); \left(1- \frac{1}{\frac{2\eta}{\gamma \mu p} + \frac{1}{p}}\right); \frac{1}{2}\right] \cdot  \E\left[\Psi_{t}\right] + \frac{12\gamma^2 \sigma^2 B_{t+1}}{e b}.
 \end{align*}

This completes the proof of the strongly convex--strongly concave case. $\square$

\textbf{Convex--concave case.} We start from \eqref{eq:t5} with small rearrangement
\begin{align*}
2 \gamma \langle  F(z_{t+1/2}),  z_{t+1/2} - z \rangle \leq& \| z_t - z \|^2_{\hat V_{t-1/2}} - \|z_{t+1} - z \|^2_{\hat V_{t-1/2}} \\
&+ \frac{\eta}{p} \|w_t - z \|^2 - \frac{\eta}{p} \|w_{t+1} - z \|^2\\
&+ \frac{\eta}{p} \|w_{t+1} - z \|^2 - \eta \| z_t - z \|^2 - (1-p) \frac{\eta}{p}  \|w_t - z \|^2 \\
&+ \frac{1}{2} \cdot \frac{3\gamma^2}{e}\| F(z_{t-1/2}) -  F(z_{t-3/2})\|^2 - \frac{3\gamma^2}{e} \| F(z_{t+1/2}) -  F(z_{t-1/2})\|^2 \\
&+\frac{1}{2} \cdot \frac{\eta}{2}\|  z_{t-1} - w_{t-1} \|^2 - \frac{\eta}{2} \| z_{t} - w_t\|^2 \\
&+ 2 \gamma \langle F(z_{t+1/2}) -g_{t+1/2},  z_{t+1/2} - z \rangle \\
&+ \frac{3\gamma^2}{e} \| g_{t+1/2} -  F(z_{t+1/2})\|^2  \nonumber\\
&+ \frac{6\gamma^2}{e} \| F(z_{t-1/2}) -  g_{t-1/2}\|^2 + \frac{3\gamma^2}{e} \| F(z_{t-3/2}) -  g_{t-3/2}\|^2.
\end{align*}
With $ I \preccurlyeq  \frac{1}{e}\hat{V}_t$ and Lemma \ref{lem:precond} with $C = \frac{\Gamma^2}{2e^2}$ for \eqref{eq:precond} and $C = \frac{2\Gamma}{e}$ for \eqref{eq:precond_add}, we get
\begin{align*}
2 \gamma \langle  F(z_{t+1/2}),  z_{t+1/2} - z \rangle \leq& \| z_t - z \|^2_{\hat V_{t-1/2}} - \left( 1 + (1-\beta_{t+1})C \right)^{-1}\|z_{t+1} - z \|^2_{\hat V_{t+1/2}} \\
&+ \frac{\eta}{p} \|w_t - z \|^2 - \frac{\eta}{p} \|w_{t+1} - z \|^2\\
&+ \frac{\eta}{p} \|w_{t+1} - z \|^2 - \eta \| z_t - z \|^2 - (1-p) \frac{\eta}{p}  \|w_t - z \|^2 \\
&+ \frac{1}{2} \cdot \frac{3\gamma^2}{e}\| F(z_{t-1/2}) -  F(z_{t-3/2})\|^2 - \frac{3\gamma^2}{e} \| F(z_{t+1/2}) -  F(z_{t-1/2})\|^2 \\
&+\frac{1}{2} \cdot \frac{\eta}{2}\|  z_{t-1} - w_{t-1} \|^2 - \frac{\eta}{2} \| z_{t} - w_t\|^2 \\
&+ 2 \gamma \langle F(z_{t+1/2}) -g_{t+1/2},  z_{t+1/2} - z \rangle \\
&+ \frac{3\gamma^2}{e} \| g_{t+1/2} -  F(z_{t+1/2})\|^2  \nonumber\\
&+ \frac{6\gamma^2}{e} \| F(z_{t-1/2}) -  g_{t-1/2}\|^2 + \frac{3\gamma^2}{e} \| F(z_{t-3/2}) -  g_{t-3/2}\|^2.
\end{align*}
Next, we sum over all $t$ from $0$ to $T-1$
\begin{align}
\label{eq:temp505}
2 \gamma \cdot \frac{1}{T}\sum\limits_{t=0}^{T-1} & \langle  F(z_{t+1/2}),  z_{t+1/2} - z \rangle \nonumber\\
\leq& \frac{\| z_0 - z \|^2_{\hat V_{-1/2}}}{T}+ 
\frac{1}{T}\sum\limits_{t=1}^{T-1} (1-\beta_{t+1}) C \|z_{t} - z \|^2_{\hat V_{t-1/2}} \nonumber\\
&+\frac{\eta\|w_0 - z \|^2}{pT} + \frac{3\gamma^2}{2e T}\| F(z_{-1/2}) -  F(z_{-3/2})\|^2
+\frac{\eta}{4T}\|  z_{-1} - w_{-1} \|^2 \nonumber\\
&- \frac{\eta}{4 T} \sum\limits_{t=1}^{T-1} \|  z_{t} - w_{t} \|^2 \nonumber\\
&+\frac{\eta}{p} \cdot \frac{1}{T}\sum\limits_{t=0}^{T-1} \left[\|w_{t+1} - z \|^2 -p \| z_t - z \|^2 - (1 - p) \|w_t - z \|^2\right] \nonumber\\
&+ 2 \gamma \cdot \frac{1}{T}\sum\limits_{t=0}^{T-1} \langle F(z_{t+1/2}) -g_{t+1/2},  z_{t+1/2} - z \rangle \nonumber\\
&+ \frac{3\gamma^2}{e} \frac{1}{T}\sum\limits_{t=0}^{T-1} \| g_{t+1/2} -  F(z_{t+1/2})\|^2  \nonumber\\
&+ \frac{6\gamma^2}{e} \frac{1}{T}\sum\limits_{t=0}^{T-1} \| F(z_{t-1/2}) -  g_{t-1/2}\|^2 + \frac{3\gamma^2}{e} \frac{1}{T}\sum\limits_{t=0}^{T-1} \| F(z_{t-3/2}) -  g_{t-3/2}\|^2.
\end{align}
The fact $\hat{V}_t \preccurlyeq \Gamma I$ gives
\begin{align*}
2 \gamma \cdot \frac{1}{T}\sum\limits_{t=0}^{T-1} & \langle  F(z_{t+1/2}),  z_{t+1/2} - z \rangle \\
\leq& \frac{\Gamma\| z_0 - z \|^2}{T}+ 
\frac{\Gamma}{T}\sum\limits_{t=1}^{T-1} (1-\beta_{t+1}) C \|z_{t} - z \|^2 \\
&+\frac{\eta\|w_0 - z \|^2}{pT} + \frac{3\gamma^2}{2e T}\| F(z_{-1/2}) -  F(z_{-3/2})\|^2
+\frac{\eta}{4T}\|  z_{-1} - w_{-1} \|^2\\
&+\frac{\eta}{p} \cdot \frac{1}{T}\sum\limits_{t=0}^{T-1} \left[\|w_{t+1} - z \|^2 -p \| z_t - z \|^2 - (1 - p) \|w_t - z \|^2\right]\\
&+ 2 \gamma \cdot \frac{1}{T}\sum\limits_{t=0}^{T-1} \langle F(z_{t+1/2}) -g_{t+1/2},  z_{t+1/2} - z \rangle \\
&+ \frac{3\gamma^2}{e} \frac{1}{T}\sum\limits_{t=0}^{T-1} \| g_{t+1/2} -  F(z_{t+1/2})\|^2  \nonumber\\
&+ \frac{6\gamma^2}{e} \frac{1}{T}\sum\limits_{t=0}^{T-1} \| F(z_{t-1/2}) -  g_{t-1/2}\|^2 + \frac{3\gamma^2}{e} \frac{1}{T}\sum\limits_{t=0}^{T-1} \| F(z_{t-3/2}) -  g_{t-3/2}\|^2.
\end{align*}
Taking maximum on $\mathcal{Z}$, using \eqref{temp8} and taking the full expectation, we obtain
\begin{align*}
2 \gamma \E\left[\text{gap}(z_{T}^{avg})\right]
\leq& \max\limits_{z \in \mathcal{Z}} \frac{\Gamma\| z_0 - z \|^2}{T}+ 
\E\left[\max\limits_{z \in \mathcal{Z}} \frac{\Gamma}{T}\sum\limits_{t=1}^{T-1} (1-\beta_{t+1}) C \|z_{t} - z \|^2\right] \\
&+\max\limits_{z \in \mathcal{Z}} \frac{\eta\|w_0 - z \|^2}{pT} + \frac{3\gamma^2}{2e T}\| F(z_{-1/2}) -  F(z_{-3/2})\|^2
+\frac{\eta}{4T}\|  z_{-1} - w_{-1} \|^2\\
&+\frac{\eta}{p} \cdot \E\left[\max\limits_{z \in \mathcal{Z}}\frac{1}{T}\sum\limits_{t=0}^{T-1} \left[\|w_{t+1} - z \|^2 -p \| z_t - z \|^2 - (1 - p) \|w_t - z \|^2\right]\right]\\
&+ 2 \gamma \cdot \E\left[\max\limits_{z \in \mathcal{Z}} \frac{1}{T}\sum\limits_{t=0}^{T-1} \langle F(z_{t+1/2}) -g_{t+1/2},  z_{t+1/2} - z \rangle \right] \\
&+ \frac{3\gamma^2}{e} \frac{1}{T}\sum\limits_{t=0}^{T-1} \E\left[\| g_{t+1/2} -  F(z_{t+1/2})\|^2\right]  \nonumber\\
&+ \frac{6\gamma^2}{e} \frac{1}{T}\sum\limits_{t=0}^{T-1} \E\left[\| F(z_{t-1/2}) -  g_{t-1/2}\|^2\right] + \frac{3\gamma^2}{e} \frac{1}{T}\sum\limits_{t=0}^{T-1} \E\left[\| F(z_{t-3/2}) -  g_{t-3/2}\|^2\right].
\end{align*}
Next, we use Assumption \ref{as:var} and the fact that iterations and the set $\mathcal{Z}$ are bounded: $\|z_{t}\|, \|z\| \leq \Omega$
\begin{align*}
2 \gamma \E\left[\text{gap}(z_{T}^{avg})\right]
\leq& \frac{4\Gamma \Omega^2}{T}+ 
\frac{4C \Gamma \Omega^2}{T}\sum\limits_{t=1}^{T-1} (1-\beta_{t+1}) +\frac{4 \eta \Omega^2}{pT} + \frac{6\gamma^2 L^2 \Omega^2}{e T}
+\frac{\eta \Omega^2}{T} + \frac{12\gamma^2 \sigma^2}{e b}\\
&+\frac{\eta}{p} \cdot \E\left[\max\limits_{z \in \mathcal{Z}}\frac{1}{T}\sum\limits_{t=0}^{T-1} \left[\|w_{t+1} - z \|^2 -p \| z_t - z \|^2 - (1 - p) \|w_t - z \|^2\right]\right]\\
&+ 2 \gamma \cdot \E\left[\max\limits_{z \in \mathcal{Z}} \frac{1}{T}\sum\limits_{t=0}^{T-1} \langle F(z_{t+1/2}) -g_{t+1/2},  z_{t+1/2} - z \rangle \right].
\end{align*}
One can estimate $ \E\left[\max\limits_{z \in \mathcal{Z}} \frac{1}{T}\sum\limits_{t=0}^{T-1} \langle F(z_{t+1/2}) -g_{t+1/2},  z_{t+1/2} - z \rangle \right]$ the same way as \eqref{eq:temp404}
\begin{align*}
2 \gamma \E&\left[\text{gap}(z_{T}^{avg})\right] \\
\leq& \frac{4\Gamma \Omega^2}{T}+ 
\frac{4C \Gamma \Omega^2}{T}\sum\limits_{t=1}^{T-1} (1-\beta_{t+1}) +\frac{4 \eta \Omega^2}{pT} + \frac{6\gamma^2 L^2 \Omega^2}{e T}
+\frac{\eta \Omega^2}{T} + \frac{12\gamma^2 \sigma^2}{e b}\\
&+\frac{\eta}{p} \cdot \E\left[\max\limits_{z \in \mathcal{Z}}\frac{1}{T}\sum\limits_{t=0}^{T-1} \left[\|w_{t+1} - z \|^2 -p \| z_t - z \|^2 - (1 - p) \|w_t - z \|^2\right]\right]\\
&+ \frac{\Gamma\Omega^2}{T} +\frac{\gamma^2 \sigma^2}{eb} \\
\leq& \frac{5\Gamma \Omega^2}{T}+ 
\frac{4C \Gamma \Omega^2}{T}\sum\limits_{t=1}^{T-1} (1-\beta_{t+1}) +\frac{4 \eta \Omega^2}{pT} + \frac{6\gamma^2 L^2 \Omega^2}{e T}
+\frac{\eta \Omega^2}{T} + \frac{13\gamma^2 \sigma^2}{e b}\\
&+\frac{\eta}{p} \cdot \E\left[\max\limits_{z \in \mathcal{Z}}\frac{1}{T}\sum\limits_{t=0}^{T-1} \left[\|w_{t+1} - z \|^2 -p \| z_t - z \|^2 - (1 - p) \|w_t - z \|^2\right]\right] \\
=&
\frac{5\Gamma \Omega^2}{T}+ \frac{4C \Gamma \Omega^2}{T}\sum\limits_{t=1}^{T-1} (1-\beta_{t+1}) +\frac{4 \eta \Omega^2}{pT} + \frac{6\gamma^2 L^2 \Omega^2}{e T}
+\frac{\eta \Omega^2}{T} + \frac{13\gamma^2 \sigma^2}{e b}\\
&+\frac{\eta}{p} \cdot \E\left[\max\limits_{z \in \mathcal{Z}}\frac{1}{T}\sum\limits_{t=0}^{T-1} \left[ -2 \langle p z_{t} + (1-p) w_t - w_{t+1}, z\rangle - p\|z_t \|^2 - (1-p) \| w_t\|^2 + \|w_{t+1} \|^2\right]\right] \\
=&
\frac{5\Gamma \Omega^2}{T}+ \frac{4C \Gamma \Omega^2}{T}\sum\limits_{t=1}^{T-1} (1-\beta_{t+1}) +\frac{4 \eta \Omega^2}{pT} + \frac{6\gamma^2 L^2 \Omega^2}{e T}
+\frac{\eta \Omega^2}{T} + \frac{13\gamma^2 \sigma^2}{e b}\\
&+\frac{\eta}{p} \cdot \E\left[\max\limits_{z \in \mathcal{Z}}\frac{1}{T}\sum\limits_{t=0}^{T-1} \left[ -2 \langle p z_{t} + (1-p) w_t - w_{t+1}, z\rangle\right]\right] \\
&+\frac{\eta}{p} \cdot \E\left[\frac{1}{T}\sum\limits_{t=0}^{T-1} \left[ - p\|z_t \|^2 - (1-p) \| w_t\|^2 + \|w_{t+1} \|^2\right]\right].
\end{align*}
One can note that by definition $w_{t+1}$: $\EE\left[ p\|z_t \|^2 + (1-p)\| w_t\|^2 - \|w_{t+1} \|^2\right] = 0$
\begin{align*}
2 \gamma \E\left[\text{gap}(z_{T}^{avg})\right] 
\leq& 
\frac{5\Gamma \Omega^2}{T}+ \frac{4C \Gamma \Omega^2}{T}\sum\limits_{t=1}^{T-1} (1-\beta_{t+1}) +\frac{4 \eta \Omega^2}{pT} + \frac{6\gamma^2 L^2 \Omega^2}{e T}
+\frac{\eta \Omega^2}{T} + \frac{13\gamma^2 \sigma^2}{e b}\\
&+\frac{\eta}{p} \cdot \E\left[\max\limits_{z \in \mathcal{Z}}\frac{1}{T}\sum\limits_{t=0}^{T-1} \left[ -2 \langle p z_{t} + (1-p) w_t - w_{t+1}, z\rangle\right]\right].
\end{align*}
With $\gamma \leq \frac{e}{10L}$ and $\eta \leq ep \leq e \leq \Gamma$, we get
\begin{align*}
2 \gamma \E\left[\text{gap}(z_{T}^{avg})\right] 
\leq& 
\frac{11\Gamma \Omega^2}{T}+ \frac{4C \Gamma \Omega^2}{T}\sum\limits_{t=1}^{T-1} (1-\beta_{t+1}) + \frac{13\gamma^2 \sigma^2}{e b}\\
&+\frac{2\eta}{p T} \cdot \E\left[\max\limits_{z \in \mathcal{Z}}\left[  \langle \sum\limits_{t=0}^{T-1}  (w_{t+1} - p z_{t} - (1-p) w_t), z\rangle\right]\right]\\
\leq& \frac{11\Gamma \Omega^2}{T}+ \frac{4C \Gamma \Omega^2}{T}\sum\limits_{t=1}^{T-1} (1-\beta_{t+1}) + \frac{13\gamma^2 \sigma^2}{e b}\\
&+\frac{2\eta}{p T} \cdot \E\left[\max\limits_{z \in \mathcal{Z}}\left[  \left\| \sum\limits_{t=0}^{T-1}  (w_{t+1} - p z_{t} - (1-p) w_t) \right\|^2 + \| z \|^2\right]\right].
\end{align*}
Again with the fact that $\EE\left[ pz_t  + (1-p) w_t - w_{t+1}\right] = 0$.
\begin{align}
\label{eq:temp606}
2 \gamma \E\left[\text{gap}(z_{T}^{avg})\right] 
\leq& \frac{11\Gamma \Omega^2}{T}+ \frac{4C \Gamma \Omega^2}{T}\sum\limits_{t=1}^{T-1} (1-\beta_{t+1}) + \frac{13\gamma^2 \sigma^2}{e b} + \frac{2\eta}{p T} \max\limits_{z \in \mathcal{Z}} \| z \|^2 \nonumber\\
&+\frac{2\eta}{p T} \cdot \E\left[ \sum\limits_{t=0}^{T-1}  \left\| w_{t+1} - p z_{t} - (1-p) w_t\right\|^2\right] \nonumber\\
\leq& \frac{13\Gamma \Omega^2}{T}+ \frac{4C \Gamma \Omega^2}{T}\sum\limits_{t=1}^{T-1} (1-\beta_{t+1}) + \frac{13\gamma^2 \sigma^2}{e b}\nonumber\\
&+\frac{2\eta}{p T} \cdot \E\left[ \sum\limits_{t=0}^{T-1}  \left\| w_{t+1} - p z_{t} - (1-p) w_t\right\|^2\right] \nonumber\\
=& \frac{13\Gamma \Omega^2}{T}+ \frac{4C \Gamma \Omega^2}{T}\sum\limits_{t=1}^{T-1} (1-\beta_{t+1}) + \frac{13\gamma^2 \sigma^2}{e b}\nonumber\\
&+\frac{2\eta}{p T} \cdot \E\left[ \sum\limits_{t=0}^{T-1} \E_{w_{t+1}} \left[\left\| w_{t+1} - \E_{w_{t+1}}[w_{t+1}]\right\|^2\right]\right] \nonumber\\
=& \frac{13\Gamma \Omega^2}{T}+ \frac{4C \Gamma \Omega^2}{T}\sum\limits_{t=1}^{T-1} (1-\beta_{t+1}) + \frac{13\gamma^2 \sigma^2}{e b} \nonumber\\
&+\frac{2\eta}{p T} \cdot \E\left[ \sum\limits_{t=0}^{T-1} \E_{w_{t+1}} \left[\left\| w_{t+1}\right\|^2\right] - \left\|\E_{w_{t+1}}[w_{t+1}]\right\|^2 \right] \nonumber\\
=& \frac{13\Gamma \Omega^2}{T}+ \frac{4C \Gamma \Omega^2}{T}\sum\limits_{t=1}^{T-1} (1-\beta_{t+1}) + \frac{13\gamma^2 \sigma^2}{e b} \nonumber\\
&+\frac{2\eta}{p T} \cdot \E\left[ \sum\limits_{t=0}^{T-1} \left[(1-p)\left\| w_{t}\right\|^2 + p\left\| z_{t}\right\|^2\right] - \left\|pz_t +(1-p)w_t\right\|^2 \right] \nonumber\\
=& \frac{13\Gamma \Omega^2}{T}+ \frac{4C \Gamma \Omega^2}{T}\sum\limits_{t=1}^{T-1} (1-\beta_{t+1}) + \frac{13\gamma^2 \sigma^2}{e b}\nonumber\\
&+\frac{2\eta}{p T} \cdot \E\left[\sum\limits_{t=0}^{T-1} (1-p) p \left\| z_t - w_{t}\right\|^2 \right] \nonumber\\
\leq& \frac{13\Gamma \Omega^2}{T}+ \frac{4C \Gamma \Omega^2}{T}\sum\limits_{t=1}^{T-1} (1-\beta_{t+1}) + \frac{13\gamma^2 \sigma^2}{e b}\nonumber\\
&+8 \cdot \E\left[ \frac{\eta}{4T}\sum\limits_{t=0}^{T-1}  \left\| z_t - w_{t}\right\|^2 \right].
\end{align}
It remains to estimate $\E\left[ \frac{\eta}{4T}\sum\limits_{t=0}^{T-1} \left\| z_t - w_{t}\right\|^2 \right]$. For this, we substitute $z = z^*$ and take the expectation
\begin{align*}
2 \gamma \cdot \frac{1}{T}\sum\limits_{t=0}^{T-1} & \E\left[\langle  F(z_{t+1/2}),  z_{t+1/2} - z^* \rangle \right] \nonumber\\
\leq& \frac{\E\left[\| z_0 - z^* \|^2_{\hat V_{-1/2}}\right]}{T}+ 
\frac{1}{T}\sum\limits_{t=1}^{T-1} (1-\beta_{t+1}) C \E\left[\|z_{t} - z^* \|^2_{\hat V_{t-1/2}}\right] \nonumber\\
&+\frac{\eta \E\left[\|w_0 - z^* \|^2\right]}{pT} + \frac{3\gamma^2}{2e T}\| F(z_{-1/2}) -  F(z_{-3/2})\|^2
+\frac{\eta}{4T}\|  z_{-1} - w_{-1} \|^2 \nonumber\\
&- \E\left[\frac{\eta}{4 T} \sum\limits_{t=1}^{T-1} \|  z_{t} - w_{t} \|^2 \right] \nonumber\\
&+\frac{\eta}{p} \cdot \frac{1}{T}\sum\limits_{t=0}^{T-1} \E\left[\|w_{t+1} - z^* \|^2 -p \| z_t - z^* \|^2 - (1 - p) \|w_t - z^* \|^2\right] \nonumber\\
&+ 2 \gamma \cdot \frac{1}{T}\sum\limits_{t=0}^{T-1} \E\left[\langle F(z_{t+1/2}) -g_{t+1/2},  z_{t+1/2} - z^* \rangle \right]\nonumber\\
&+ \frac{3\gamma^2}{e} \frac{1}{T}\sum\limits_{t=0}^{T-1} \E\left[\| g_{t+1/2} -  F(z_{t+1/2})\|^2\right]  \nonumber\\
&+ \frac{6\gamma^2}{e} \frac{1}{T}\sum\limits_{t=0}^{T-1} \E\left[\| F(z_{t-1/2}) -  g_{t-1/2}\|^2\right] + \frac{3\gamma^2}{e} \frac{1}{T}\sum\limits_{t=0}^{T-1} \E\left[\| F(z_{t-3/2}) -  g_{t-3/2}\|^2\right].
\end{align*}
Assumptions \ref{as:conv} (convexity), \ref{as:var} and the fact: $\EE\left[ p\|z_t \|^2 + (1-p)\| w_t\|^2 - \|w_{t+1} \|^2\right] = 0$, give
\begin{align}
\label{eq:temp707}
\E\left[\frac{\eta}{4 T} \sum\limits_{t=1}^{T-1} \|  z_{t} - w_{t} \|^2 \right] \leq& \frac{\E\left[\| z_0 - z^* \|^2_{\hat V_{-1/2}}\right]}{T}+ 
\frac{1}{T}\sum\limits_{t=1}^{T-1} (1-\beta_{t+1}) C \E\left[\|z_{t} - z^* \|^2_{\hat V_{t-1/2}}\right] \nonumber\\
&+\frac{\eta \E\left[\|w_0 - z^* \|^2\right]}{pT} + \frac{3\gamma^2}{2e T}\| F(z_{-1/2}) -  F(z_{-3/2})\|^2
+\frac{\eta}{4T}\|  z_{-1} - w_{-1} \|^2 \nonumber\\
&+ \frac{12\gamma^2 \sigma^2}{e b} \nonumber\\
\leq& \frac{4\Gamma \Omega^2}{T}+ 
\frac{4C \Gamma \Omega^2}{T}\sum\limits_{t=1}^{T-1} (1-\beta_{t+1}) +\frac{4\eta \Omega^2}{pT} + \frac{3\gamma^2 L^2 \Omega^2}{2e T}
+\frac{\eta \Omega^2}{T} + \frac{12\gamma^2 \sigma^2}{e b}.
\end{align}
Combining \eqref{eq:temp606} and \eqref{eq:temp707}, we get
\begin{align*}
2 \gamma \E\left[\text{gap}(z_{T}^{avg})\right] 
\leq& \frac{13\Gamma \Omega^2}{T}+ \frac{4C \Gamma \Omega^2}{T}\sum\limits_{t=1}^{T-1} (1-\beta_{t+1}) + \frac{13\gamma^2 \sigma^2}{e b}\nonumber\\
&+8 \cdot \left[ \frac{4\Gamma \Omega^2}{T}+ 
\frac{4C \Gamma \Omega^2}{T}\sum\limits_{t=1}^{T-1} (1-\beta_{t+1}) +\frac{4\eta \Omega^2}{pT} + \frac{3\gamma^2 L^2 \Omega^2}{2e T}
+\frac{\eta \Omega^2}{T} + \frac{12\gamma^2 \sigma^2}{e b} \right] \\
\leq& \frac{100\Gamma \Omega^2}{T}+ \frac{36C \Gamma \Omega^2}{T}\sum\limits_{t=1}^{T-1} (1-\beta_{t+1}) + \frac{110\gamma^2 \sigma^2}{e b}.
\end{align*}
Finally, we have
\begin{align*}
\E\left[\text{gap}(z_{T}^{avg})\right] 
\leq& \frac{50\Gamma \Omega^2}{\gamma T}+ \frac{18C \Gamma \Omega^2}{\gamma T}\sum\limits_{t=1}^{T} (1-\beta_{t}) + \frac{55\gamma \sigma^2}{e b}.
\end{align*}

This completes the proof of the convex--concave case. $\square$
 
\textbf{Non-convex--non-concave case.} We start from \eqref{eq:t5}, substitute $z = z^*$  and take the total expectation of both sides of the equation
\begin{align*}
\E\left[\|z_{t+1} - z^* \|^2_{\hat V_{t-1/2}}\right] &+ \frac{3\gamma^2}{e} \E\left[\| F(z_{t+1/2}) -  F(z_{t-1/2})\|^2\right]
+ \frac{\eta}{2} \E\left[\| z_{t} - w_t\|^2\right] \nonumber\\
\leq& \E\left[\| z_t - z \|^2_{\hat V_{t-1/2}}\right] -\eta \E\left[\| z_t - z^* \|^2\right] + \eta \E\left[\|w_t - z^* \|^2\right] \nonumber\\  
&- 2 \gamma \E\left[\langle F(z^{t+1/2}),  z_{t+1/2} - z^* \rangle\right] \nonumber\\  
&+ \frac{1}{2} \cdot \frac{3\gamma^2}{e} \E\left[\| F(z_{t-1/2}) -  F(z_{t-3/2})\|^2\right] + \frac{1}{2} \cdot \frac{\eta}{2} \E\left[\|  z_{t-1} - w_{t-1} \|^2 \right] \nonumber\\ 
 &  - \frac{1}{4} \E\left[\| z_{t+1/2} -  z_t \|^2_{\hat V_{t-1/2}} \right] + \frac{3\gamma^2}{e} \E\left[\| g_{t+1/2} -  F(z_{t+1/2})\|^2\right]  \nonumber\\
&+ \frac{6\gamma^2}{e} \E\left[\| F(z_{t-1/2}) -  g_{t-1/2}\|^2\right] + \frac{3\gamma^2}{e} \E\left[\| F(z_{t-3/2}) -  g_{t-3/2}\|^2\right].
 \end{align*}
 Assumptions \ref{as:conv} and \ref{as:var} on non-convexity--non-concavity and on stochastisity give
\begin{align*}
\E\left[\|z_{t+1} - z^* \|^2_{\hat V_{t-1/2}}\right] &+ \frac{3\gamma^2}{e} \E\left[\| F(z_{t+1/2}) -  F(z_{t-1/2})\|^2\right]
+ \frac{\eta}{2} \E\left[\| z_{t} - w_t\|^2\right] \nonumber\\
\leq& \E\left[\| z_t - z^* \|^2_{\hat V_{t-1/2}}\right] -\eta \E\left[\| z_t - z^* \|^2\right] + \eta \E\left[\|w_t - z^* \|^2\right] \nonumber\\  
&+ \frac{1}{2} \cdot \frac{3\gamma^2}{e} \E\left[\| F(z_{t-1/2}) -  F(z_{t-3/2})\|^2\right] + \frac{1}{2} \cdot \frac{\eta}{2} \E\left[\|  z_{t-1} - w_{t-1} \|^2 \right] \nonumber\\ 
 &  - \frac{\gamma^2}{4} \E\left[\| g_{t-1/2} \|^2_{\hat V^{-1}_{t-1/2}} \right] + \frac{12\gamma^2 \sigma^2}{e b}\nonumber\\
\leq& \E\left[\| z_t - z^* \|^2_{\hat V_{t-1/2}}\right] -\eta \E\left[\| z_t - z^* \|^2\right] + \eta \E\left[\|w_t - z^* \|^2\right] \nonumber\\  
&+ \frac{1}{2} \cdot \frac{3\gamma^2}{e} \E\left[\| F(z_{t-1/2}) -  F(z_{t-3/2})\|^2\right] + \frac{1}{2} \cdot \frac{\eta}{2} \E\left[\|  z_{t-1} - w_{t-1} \|^2 \right] \nonumber\\ 
 &  - \frac{\gamma^2}{4 \Gamma} \E\left[\| g_{t-1/2} \|^2 \right] + \frac{12\gamma^2 \sigma^2 }{e b} \nonumber\\
\leq& \E\left[\| z_t - z^* \|^2_{\hat V_{t-1/2}}\right] -\eta \E\left[\| z_t - z^* \|^2\right] + \eta \E\left[\|w_t - z^* \|^2\right] \nonumber\\  
&+ \frac{1}{2} \cdot \frac{3\gamma^2}{e} \E\left[\| F(z_{t-1/2}) -  F(z_{t-3/2})\|^2\right] + \frac{1}{2} \cdot \frac{\eta}{2} \E\left[\|  z_{t-1} - w_{t-1} \|^2 \right] \nonumber\\ 
 &  - \frac{\gamma^2}{8 \Gamma} \E\left[\| F(z_{t-1/2}) \|^2 \right] + \frac{\gamma^2}{4 \Gamma} \E\left[\| g_{t-1/2} - F(z_{t-1/2}) \|^2 \right] + \frac{12\gamma^2 \sigma^2}{e b}
 \nonumber\\
\leq& \E\left[\| z_t - z^* \|^2_{\hat V_{t-1/2}}\right] -\eta \E\left[\| z_t - z^* \|^2\right] + \eta \E\left[\|w_t - z^* \|^2\right] \nonumber\\  
&+ \frac{1}{2} \cdot \frac{3\gamma^2}{e} \E\left[\| F(z_{t-1/2}) -  F(z_{t-3/2})\|^2\right] + \frac{1}{2} \cdot \frac{\eta}{2} \E\left[\|  z_{t-1} - w_{t-1} \|^2 \right] \nonumber\\ 
 &  - \frac{\gamma^2}{8 \Gamma} \E\left[\| F(z_{t-1/2}) \|^2 \right] +\frac{13\gamma^2 \sigma^2}{e b}.
 \end{align*}
Here we additionally use $\frac{1}{\Gamma} I \preccurlyeq \hat{V}^{-1}_{t-1/2}$. Lemma \ref{lem:precond} with $C = \frac{\Gamma^2}{2e^2}$ for \eqref{eq:precond}, $C = \frac{2\Gamma}{e}$ for \eqref{eq:precond_add} and $B_{t+1} = \left( 1 + (1-\beta_{t+1}) C \right)$ gives
\begin{align*}
\frac{1}{B_{t+1}}\E\left[\|z_{t+1} - z^* \|^2_{\hat V_{t+1/2}}\right] &+ \frac{3\gamma^2}{e} \E\left[\| F(z_{t+1/2}) -  F(z_{t-1/2})\|^2\right]
+ \frac{\eta}{2} \E\left[\| z_{t} - w_t\|^2\right] \nonumber\\
\leq& \E\left[\| z_t - z^* \|^2_{\hat V_{t-1/2}}\right] -\eta \E\left[\| z_t - z^* \|^2\right] + \eta \E\left[\|w_t - z^* \|^2\right] \nonumber\\  
&+ \frac{1}{2} \cdot \frac{3\gamma^2}{e} \E\left[\| F(z_{t-1/2}) -  F(z_{t-3/2})\|^2\right] + \frac{1}{2} \cdot \frac{\eta}{2} \E\left[\|  z_{t-1} - w_{t-1} \|^2 \right] \nonumber\\ 
 &  - \frac{\gamma^2}{8 \Gamma} \E\left[\| F(z_{t-1/2}) \|^2 \right] +\frac{13\gamma^2 \sigma^2}{e b}.
 \end{align*}
Next, after small rearrangement, we obtain
\begin{align*}
\frac{\gamma^2}{8 \Gamma} \E\left[\| F(z_{t-1/2}) \|^2 \right] \leq& \E\left[\| z_t - z^* \|^2_{\hat V_{t-1/2}}\right] - \frac{1}{B_{t+1}}\E\left[\|z_{t+1} - z^* \|^2_{\hat V_{t+1/2}}\right] \\
& -\eta \E\left[\| z_t - z^* \|^2\right] + \eta \E\left[\|w_t - z^* \|^2\right]\\
&+\frac{1}{2} \cdot \frac{3\gamma^2}{e} \E\left[\| F(z_{t-1/2}) -  F(z_{t-3/2})\|^2\right] - \frac{3\gamma^2}{e} \E\left[\| F(z_{t+1/2}) -  F(z_{t-1/2})\|^2\right]  \\
&+\frac{1}{2} \cdot \frac{\eta}{2} \E\left[\|  z_{t-1} - w_{t-1} \|^2 \right] - \frac{\eta}{2} \E\left[\| z_{t} - w_t\|^2\right] \\
&+\frac{13\gamma^2 \sigma^2}{e b}.
 \end{align*}
The update of $w_{t+1}$ gives
\begin{align*}
\frac{\eta}{p }\E\left[\|w_{t+1} - z^* \|^2\right] &=
\frac{\eta}{p }\E\left[\E_{w_{t+1}}\left[\|w_{t+1} - z^* \|^2\right]\right] \\
&=
\eta \E\left[\| z_t - z^* \|^2\right] + \frac{(1-p)\eta }{p}\E\left[\|w_{t} - z^* \|^2\right].
 \end{align*}
Connecting with the previous equation, we get
\begin{align*}
\frac{\gamma^2}{8 \Gamma} \E\left[\| F(z_{t-1/2}) \|^2 \right] \leq& \E\left[\| z_t - z^* \|^2_{\hat V_{t-1/2}}\right] - \frac{1}{B_{t+1}}\E\left[\|z_{t+1} - z^* \|^2_{\hat V_{t+1/2}}\right] \\
& + \frac{\eta}{p} \E\left[\|w_t - z^* \|^2\right] - \frac{\eta}{p} \E\left[\|w_{t+1} - z^* \|^2\right]\\
&+\frac{1}{2} \cdot \frac{3\gamma^2}{e} \E\left[\| F(z_{t-1/2}) -  F(z_{t-3/2})\|^2\right] - \frac{3\gamma^2}{e} \E\left[\| F(z_{t+1/2}) -  F(z_{t-1/2})\|^2\right]  \\
&+\frac{1}{2} \cdot \frac{\eta}{2} \E\left[\|  z_{t-1} - w_{t-1} \|^2 \right] - \frac{\eta}{2} \E\left[\| z_{t} - w_t\|^2\right] \\
&+\frac{13\gamma^2 \sigma^2}{e b}.
 \end{align*}
Summing over all $t$ from $0$ to $T-1$ and averaging, we get
\begin{align*}
\frac{\gamma^2}{8 \Gamma} \cdot \E\left[\frac{1}{T}\sum\limits_{t=0}^{T-1} \| F(z_{t-1/2}) \|^2 \right] \leq& \frac{\E\left[\| z_0 - z^* \|^2_{\hat V_{-1/2}}\right]}{T} + \frac{1}{T} \sum\limits_{t=1}^{T-1} \frac{B_{t} - 1}{B_{t}} \E\left[\|z_{t} - z^* \|^2_{\hat V_{t-1/2}}\right]\\
& + \frac{\eta}{p} \E\left[\|w_0 - z^* \|^2\right]+\frac{3\gamma^2}{2e} \E\left[\| F(z_{-1/2}) -  F(z_{-3/2})\|^2\right] \\
&+\frac{\eta}{4} \E\left[\|  z_{-1} - w_{-1} \|^2 \right] +\frac{13\gamma^2 \sigma^2}{e b}.
 \end{align*}
With the fact that iterations are bounded: $\|z_{t}\| \leq \Omega$, we obtain
\begin{align*}
\frac{\gamma^2}{8 \Gamma} \cdot \E\left[\frac{1}{T}\sum\limits_{t=0}^{T-1} \| F(z_{t-1/2}) \|^2 \right] \leq& \frac{4\Gamma \Omega^2}{T}  + \frac{8\eta \Omega^2}{p T} + \frac{12\gamma^2 L^2 \Omega^2}{2e T} +\frac{13\gamma^2 \sigma^2}{e b}\\
& + \frac{4\Gamma \Omega^2}{T} \sum\limits_{t=1}^{T-1} \frac{(1-\beta_{t}) C}{1 + (1-\beta_{t}) C} \\
\leq& \frac{4\Gamma \Omega^2}{T}  + \frac{8\eta \Omega^2}{p T} + \frac{12\gamma^2 L^2 \Omega^2}{2e T} +\frac{13\gamma^2 \sigma^2}{e b}\\
& + \frac{4\Gamma C \Omega^2}{T} \sum\limits_{t=1}^{T-1} (1-\beta_{t}).
 \end{align*}
Here we use $\left( 1 + (1-\beta_{t}) C \right) \geq 1$. With $\eta \leq ep$, we get
\begin{align*}
\E\left[\frac{1}{T}\sum\limits_{t=0}^{T-1} \| F(z_{t-1/2}) \|^2 \right] 
\leq& \frac{96\Gamma^2 \Omega^2}{\gamma^2 T}  + \frac{48 \Gamma L^2 \Omega^2}{e T} + \frac{32 \Gamma^2 C \Omega^2}{\gamma^2 T} \sum\limits_{t=1}^{T-1} (1-\beta_{t}) +\frac{105 \Gamma \sigma^2}{e b}.
 \end{align*}
If we choose $\bar z_{T-1/2}$ randomly and uniformly, then $ \E\left[ \| F(\bar z_{T-1/2}) \|^2\right] \leq \E\left[ \frac{1}{T} \sum\limits_{t=0}^{T-1}\| F(z_{t-1/2}) \|^2\right]$ and 
\begin{align*}
\E\left[\| F(\bar z_{T-1/2}) \|^2 \right] 
\leq& \frac{96\Gamma^2 \Omega^2}{\gamma^2 T}  + \frac{48 \Gamma L^2 \Omega^2}{e T} + \frac{32 \Gamma^2 C \Omega^2}{\gamma^2 T} \sum\limits_{t=1}^{T-1} (1-\beta_{t}) +\frac{105 \Gamma \sigma^2}{e b}.
 \end{align*}
This completes the proof of the non-convex--non-concave case. $\square$

\subsection{Proof of Corollary \ref{cor:main1_oasis}}

\textbf{Strongly convex--strongly concave case.} With $\beta_t \equiv \beta \geq 1 - \frac{\gamma \mu}{ 4 \Gamma C}$, we have \eqref{eq:th1_sc} in the following form
\begin{align*}
\E\left[\Psi_{t+1}\right] \leq&  \max \left[ \left(1 - \frac{\gamma \mu}{4 \Gamma}\right); \left(1- \frac{1}{\frac{2\eta}{\gamma \mu p} + \frac{1}{p}}\right)\right] \cdot  \E\left[\Psi_{t}\right] +  \frac{24\gamma^2 \sigma^2 }{e b}.
\end{align*}
Running the recursion, we have
\begin{align*}
\E\left[ \Psi_{T} \right] 
 \leq&  \max \left[ \left(1 - \frac{\gamma \mu}{4 \Gamma}\right)^T; \left(1- \frac{1}{\frac{2\eta}{\gamma \mu p} + \frac{1}{p}}\right)^T\right]\cdot \Psi_0 \\
 &+ \frac{12\gamma^2 \sigma^2}{eb} \sum\limits_{t=0}^{T-1} \max \left[ \left(1 - \frac{\gamma \mu}{4 \Gamma}\right)^t; \left(1- \frac{1}{\frac{2\eta}{\gamma \mu p} + \frac{1}{p}}\right)^t\right] \\
 \leq&  \max \left[\exp\left( - \frac{\gamma \mu T}{4\Gamma}\right); \exp\left( - \frac{1}{\frac{2\eta}{\gamma \mu p} + \frac{1}{p}} \right)\right]\cdot \Psi_0 + \frac{48\Gamma \gamma \sigma^2}{eb\mu} + \frac{48 \eta \gamma \sigma^2 }{e p \mu b} + \frac{24\gamma^2 \sigma^2 }{e p b}.
\end{align*}
Finally, we need tuning of $\gamma \leq \frac{e}{10 L}$:

$\bullet$ If $\frac{e}{10 L} \geq \frac{4\Gamma \ln\left( \max\{2, eb\mu^2 \Psi_0 T/(48 \Gamma^2\sigma^2) \} \right)}{\mu T}$ then $\gamma = \frac{2\Gamma \ln\left( \max\{2, eb\mu^2 R^2_0 T/(48 \Gamma^2\sigma^2) \} \right)}{\mu T}$ gives
    \begin{align*}
    \mathcal{\tilde O} \left( \exp\left(- \ln\left( \max\{2, eb\mu^2 R^2_0 T/(48 \Gamma^2\sigma^2) \} \right) \right) R_0^2 + \frac{\Gamma^2 \sigma^2}{e \mu^2 T}\right) = \mathcal{\tilde O} \left( \frac{\Gamma^2 \sigma^2}{e b\mu^2 T} \right).  
    \end{align*}

$\bullet$ If $\frac{e}{10 L} \leq \frac{4\Gamma \ln\left( \max\{2, eb\mu^2 R^2_0 T/(48 \Gamma^2\sigma^2) \} \right)}{\mu T}$ then $\gamma = \frac{e}{10 L}$ gives
    \begin{align*}
    \mathcal{\tilde O} \left( \exp\left(- \frac{e\mu T}{40\Gamma L}\right) \Psi_0 + \exp\left(- \frac{ep\mu T}{40 \eta L}\right) \Psi_0 + \exp\left(- p\right) \Psi_0 + \frac{\Gamma \gamma \sigma^2}{eb\mu}\right) \\
    \leq \mathcal{\tilde O} \left(  \exp\left(- \frac{e\mu T}{40\Gamma L}\right) \Psi_0 + \exp\left(- \frac{ep\mu T}{40\eta L}\right) \Psi_0 + \exp\left(- p\right) \Psi_0 + \frac{\Gamma^2 \sigma^2}{eb \mu^2 T}\right).  
    \end{align*}
What in the end with $\eta \leq ep \leq e$ gives that
\begin{align*}
 \E\left[ \Psi_{T} \right] = \mathcal{\tilde O} \left( \exp\left(- \frac{e \mu T}{40 \Gamma L}\right) \Psi_0 +  \exp\left(- \frac{\mu T}{40 L}\right) \Psi_0 +  \exp\left(- p\right) \Psi_0  + \frac{\Gamma^2 \sigma^2}{e b\mu^2 T}\right).  
\end{align*}
This completes the proof of the strongly convex--strongly concave case. $\square$

\textbf{Convex--concave case.}  With $\beta_t \equiv \beta \geq 1 - \frac{1}{CT}$, we have \eqref{eq:th1_c} in the following form
 \begin{align*} 
 \E\left[\text{gap}(x_{T}^{avg}, y_{T}^{avg})\right] \leq& \frac{68\Gamma \Omega^2}{\gamma T}+ \frac{55\gamma \sigma^2}{e}.
 \end{align*}
If $\gamma = \min\left\{\frac{e}{10L}; \frac{\sqrt{\Gamma e b}\Omega}{\sigma \sqrt{T}} \right\}$, then
 \begin{align*} 
 \E\left[\text{gap}(x_{T}^{avg}, y_{T}^{avg})\right] =& \mathcal{O}\left(\frac{\Gamma L\Omega^2}{e T} + 
 \frac{\sqrt{\Gamma}\sigma \Omega}{\sqrt{e b T}} \right).
 \end{align*}
 This completes the proof of the convex--concave case. $\square$

\textbf{Non-convex--non-concave case.} With $\beta_t \equiv \beta \geq 1 - \frac{1}{CT}$, we have \eqref{eq:th1_nc} in the following form
 \begin{equation*}
 \begin{split}
\E[ \| \nabla_x f(\bar x_{T-1/2}, \bar y_{T-1/2}) \|^2 &+ \| \nabla_y f(\bar x_{T-1/2}, \bar y_{T-1/2}) \|^2] \\
\leq& \frac{128\Gamma^2 \Omega^2}{\gamma^2 T}  + \frac{48 \Gamma L^2 \Omega^2}{e T} +\frac{105 \Gamma \sigma^2}{e b}.
 \end{split}
 \end{equation*}
The choice of $\gamma = \frac{e}{10L}$ gives
 \begin{equation*}
 \begin{split}
\E[ \| \nabla_x f(\bar x_{T-1/2}, \bar y_{T-1/2}) \|^2 + \| \nabla_y f(\bar x_{T-1/2}, \bar y_{T-1/2}) \|^2]
\leq& \frac{120^2 \Gamma^2 L^2 \Omega^2}{e^2 T}  +\frac{105 \Gamma \sigma^2}{e b}.
 \end{split}
 \end{equation*}

The batch size $b \sim \frac{\Gamma \sigma^2}{e \varepsilon^2}$ completes the proof of the non-convex--non-concave case. $\square$
 

\subsection{Proof of  Corollary \ref{cor:main1_adam}}

\textbf{Strongly convex--strongly concave case.}
We start from \eqref{eq:th1_sc} with small rearrangement
 \begin{equation*}
 \begin{split}
     \min\left[ \frac{\gamma \mu}{4 \Gamma} ;  \frac{1}{\frac{4\eta}{\gamma \mu p} + \frac{2}{p}} \right]\E\left[\Psi_{t}\right] \leq&  \max \left[ \left(1 - \frac{\gamma \mu}{4 \Gamma} + (1-\beta_{t+1}) C \right); \left(1- \frac{1}{\frac{4\eta}{\gamma \mu p} + \frac{2}{p}}\right)\right] \cdot  \E\left[\Psi_{t}\right] \\
&- \E\left[\Psi_{t+1}\right]  +  \frac{12\gamma^2 \sigma^2 \left( 1 + (1-\beta_{t+1}) C \right)}{e b};
 \end{split}
 \end{equation*}
Summing over all $t$ from $0$ to $T-1$ and averaging, we get
\begin{align*}
\min\left[ \frac{\gamma \mu}{4 \Gamma } ;  \frac{1}{\frac{4\eta }{\gamma \mu p} + \frac{2 }{p}} \right] &\E\left[ \frac{1}{T}\sum_{t=0}^{T-1} \Psi_t\right]  \\
 \leq&  \frac{\E\left[\Psi_0\right]}{T} + 
\frac{1}{T}\sum_{t=0}^{T-1} \left( (1-\beta_{t+1})C - \frac{\gamma \mu}{4\Gamma} \right)\E\left[\Psi_t\right]  \\
&+ \frac{12\gamma^2 \sigma^2}{eb} \cdot \frac{1}{T}\sum_{t=0}^{T-1} \left( 1 + (1-\beta_{t+1})C \right).
\end{align*}
Using $e I \preccurlyeq \hat{V}_t \preccurlyeq \Gamma I$ and $\|z_t\| \leq \Omega$, we have
\begin{align*}
\min\left[ \frac{e\gamma \mu}{4 \Gamma } ;  \frac{e}{\frac{4\eta }{\gamma \mu p} + \frac{2 }{p}} \right] \E\left[ \frac{1}{T}\sum_{t=0}^{T-1} \| z_t - z^*\|^2\right] 
\leq&  \frac{20\Gamma \Omega^2}{T} + 
\frac{1}{T}\sum_{t=1}^{T} \left( (1-\beta_{t})C - \frac{\gamma \mu}{4\Gamma} \right)\E\left[\Psi_t\right]  \\
&+ \frac{12\gamma^2 \sigma^2}{eb} \cdot \frac{1}{T}\sum_{t=1}^{T} \left( 1 + (1-\beta_{t})C \right).
\end{align*}
Let us define $\alpha = \left(\frac{ 4 \Gamma C}{\gamma \mu}\right)^2$. Then, $1 - \beta_t = \frac{1 - \beta}{1 - \beta^{t+1}}$ with $\beta = 1 - \frac{1}{\alpha}$. \eqref{eq:tech_lem} gives 
$$
\beta^{\sqrt{\alpha}} \leq 1 - \frac{1}{2\sqrt{\alpha}}.
$$
And hence, for all $t \geq \sqrt{\alpha}$
$$
1 - \beta_t \leq  \frac{2\sqrt{\alpha}}{\alpha} =  \frac{2}{\sqrt{\alpha}} = \frac{\gamma \mu}{4 \Gamma C}.
$$
Then we get
\begin{align*}
\min\left[ \frac{e\gamma \mu}{4 \Gamma } ;  \frac{e}{\frac{4\eta }{\gamma \mu p} + \frac{2 }{p}} \right] &\E\left[ \frac{1}{T}\sum_{t=0}^{T-1} \| z_t - z^*\|^2\right]  \\
\leq&  \frac{20\Gamma \Omega^2}{T} + 
\frac{1}{T}\sum_{t=1}^{\sqrt{\alpha}} \left( (1-\beta_{t})C - \frac{\gamma \mu}{4\Gamma} \right)\E\left[\Psi_t\right]  \\
&+ \frac{12\gamma^2 \sigma^2}{eb} \cdot \frac{1}{T}\sum_{t=1}^{T} \left( 1 + (1-\beta_{t})C \right) \\
\leq&  \frac{20\Gamma \Omega^2}{T} + 
\frac{1}{T}\sum_{t=1}^{\sqrt{\alpha}} 20 C\Gamma \Omega^2  + \frac{12\gamma^2 \sigma^2}{eb} \cdot \frac{1}{T}\sum_{t=1}^{T} \left( 1 + (1-\beta_{t})C \right) \\
\leq&  \frac{20\Gamma \Omega^2}{T} + 
\frac{20 C\Gamma \Omega^2 }{T} \cdot \frac{ 4 \Gamma C}{\gamma \mu}  + \frac{12\gamma^2 \sigma^2}{eb} \cdot \frac{1}{T}\sum_{t=1}^{T} \left( 1 + (1-\beta_{t})C \right) \\
\leq&  \frac{100 C^2\Gamma^2 \Omega^2 }{\gamma \mu T} + \frac{24 C\gamma^2 \sigma^2}{eb}.
\end{align*}
Then, 
\begin{align*}
\E\left[ \frac{1}{T}\sum_{t=0}^{T-1} \| z_t - z^*\|^2\right] 
=&  \mathcal{O}\left(\frac{C^2\Gamma^3 \Omega^2 }{e\gamma^2 \mu^2 T} + \frac{C \Gamma \gamma \sigma^2}{e^2 \mu  b} + \frac{ C^2\Gamma^2 \Omega^2 }{\gamma^2 \mu^2 T} + \frac{ C\gamma \sigma^2}{e \mu b} + \frac{C^2\Gamma^2 \Omega^2 }{e p\gamma \mu T} + \frac{C\gamma^2 \sigma^2}{e^2 pb} \right).
\end{align*}
Finally, $\gamma = \min\left\{ \frac{e}{10L} ; \sqrt[3]{\frac{C \Gamma^2 \Omega^2 e b}{\mu \sigma^2 T}} \right\}$
\begin{align*}
\E&\left[ \left\|  \frac{1}{T}\sum_{t=0}^{T-1} z_t - z^* \right\|^2\right] \leq \E\left[ \frac{1}{T}\sum_{t=0}^{T-1} \|z_t - z^* \|^2\right] \\
=& \mathcal{O}\Bigg(\frac{C^2\Gamma^3 L^2 \Omega^2 }{e^3 \mu^2 T} + \frac{ C^2\Gamma^2 L^2 \Omega^2 }{e^2 \mu^2 T} + \frac{C^2\Gamma^2  L \Omega^2 }{e^2 p \mu T} + \left(\frac{C^4 \Gamma^5 \sigma^4 \Omega^2}{e^5 \mu^4  b^2 T}\right)^{1/3}  \\
&+ \left(\frac{ C^4 \Gamma^2 \sigma^4 \Omega^2}{e^2 \mu^4 b^2 T}\right)^{1/3}  + \left(\frac{C^5 \Gamma^4 \sigma^2 \Omega^4}{e^4 \mu^2 p^3 b T^2}\right)^{1/3} \Bigg).
\end{align*}

This completes the proof of the strongly convex--strongly concave case. $\square$

\textbf{Convex--concave case.}
With $\beta_t = \frac{\beta - \beta^{t+1}}{1 - \beta^{t+1}}$ we get \eqref{eq:th1_c} in the following form
\begin{align*} 
 \E\left[\text{gap}(x_{T}^{avg}, y_{T}^{avg})\right] \leq& \frac{50\Gamma \Omega^2}{\gamma T}  + \frac{55\gamma \sigma^2}{e b} \\
 &+ \frac{18(1-\beta)C \Gamma \Omega^2}{\gamma T}\sum\limits_{t=1}^{\sqrt{T}} \frac{1}{1-\beta} + \frac{18C \Gamma \Omega^2}{\gamma T}\sum\limits_{t=1}^{T} \frac{1}{1-\beta^{\sqrt{T}}} \\
 \leq& \frac{50\Gamma \Omega^2}{\gamma T}  + \frac{55\gamma \sigma^2}{e b} \\
 &+ \frac{18C \Gamma \Omega^2}{\gamma \sqrt{T}} + \frac{18 (1 - \beta)C \Gamma \Omega^2}{\gamma T}\sum\limits_{t=1}^{T} \frac{1}{1-\beta^{\sqrt{T}}}.
 \end{align*}
Next, we substitute $\beta = 1 - \frac{1}{T}$. From \eqref{eq:tech_lem} we get that $\beta^{\sqrt{T}} \leq 1 - \frac{1}{2\sqrt{T}}$. Then, we get
\begin{align*} 
 \E\left[\text{gap}(x_{T}^{avg}, y_{T}^{avg})\right] 
\leq& \frac{50\Gamma \Omega^2}{\gamma T}  + \frac{55\gamma \sigma^2}{e b} 
 + \frac{54C \Gamma \Omega^2}{\gamma \sqrt{T}} .
 \end{align*}
It means that 
\begin{align*} 
 \E\left[\text{gap}(x_{T}^{avg}, y_{T}^{avg})\right] 
=& \mathcal{O}\left(\frac{C\Gamma \Omega^2}{\gamma \sqrt{T}} + \frac{\gamma \sigma^2}{eb} \right).
 \end{align*}
 With $\gamma = \min\left\{\frac{e}{10L}; \frac{\sqrt{C\Gamma e b}\Omega}{\sigma {T^{1/4}}} \right\}$
\begin{align*} 
 \E\left[\text{gap}(x_{T}^{avg}, y_{T}^{avg})\right] 
=& \mathcal{O}\left(\frac{C\Gamma L\Omega^2}{e\sqrt{T}} + \frac{\sqrt{C\Gamma e b} \sigma \Omega}{e T^{1/4}} \right).
 \end{align*}
 This completes the proof of the convex--concave case. $\square$

\textbf{Non-convex--non-concave case.}
With $\beta_t = \frac{\beta - \beta^{t+1}}{1 - \beta^{t+1}}$ we get \eqref{eq:th1_nc} in the following form
\begin{align*}
\E[ \| \nabla_x f(\bar x_{T-1/2}, \bar y_{T-1/2}) \|^2 &+ \| \nabla_y f(\bar x_{T-1/2}, \bar y_{T-1/2}) \|^2] \\
\leq& \frac{96\Gamma^2 \Omega^2}{\gamma^2 T}  + \frac{48 \Gamma L^2 \Omega^2}{e T} + \frac{32 (1 - \beta) \Gamma^2 C \Omega^2}{\gamma^2 T} \sum\limits_{t=1}^{T} \frac{1}{1-\beta^{t}} +\frac{105 \Gamma \sigma^2}{e b} \\ 
\leq& \frac{96\Gamma^2 \Omega^2}{\gamma^2 T}  + \frac{48 \Gamma L^2 \Omega^2}{e T} +\frac{105 \Gamma \sigma^2}{e b} \\
&+ \frac{32 (1 - \beta) \Gamma^2 C \Omega^2}{\gamma^2 T} \sum\limits_{t=1}^{\sqrt{T}} \frac{1}{1-\beta} +  \frac{32 (1 - \beta) \Gamma^2 C \Omega^2}{\gamma^2 T} \sum\limits_{t=1}^{T} \frac{1}{1-\beta^{\sqrt{T}}}.
\end{align*}
Next, we substitute $\beta = 1 - \frac{1}{T}$. From \eqref{eq:tech_lem} we get that $\beta^{\sqrt{T}} \leq 1 - \frac{1}{2\sqrt{T}}$. Then, we get
\begin{align*}
\E[ \| \nabla_x f(\bar x_{T-1/2}, \bar y_{T-1/2}) \|^2 &+ \| \nabla_y f(\bar x_{T-1/2}, \bar y_{T-1/2}) \|^2] \\
\leq& \frac{96\Gamma^2 \Omega^2}{\gamma^2 T}  + \frac{48 \Gamma L^2 \Omega^2}{e T} +\frac{105 \Gamma \sigma^2}{e b} + \frac{96  \Gamma^2 C \Omega^2}{\gamma^2 \sqrt{T}}.
\end{align*}
The batch size $b \sim \frac{\Gamma \sigma^2}{e \varepsilon^2}$ and $\gamma = \frac{e}{10L}$ complete the proof of the non-convex--non-concave case. $\square$

\end{document}